\documentclass{article}

\PassOptionsToPackage{numbers, compress}{natbib}

\usepackage[preprint]{neurips_2026}

 \usepackage{booktabs}
 \usepackage{tabularx}
 \usepackage{makecell}
 \usepackage{array}
 \usepackage{adjustbox}

\usepackage{blkarray}
\usepackage{multirow}
\usepackage{tikz-cd}
\usepackage{yfonts}

\usepackage[utf8]{inputenc} %
\usepackage{amsmath,amssymb,amsthm,amsfonts,mathtools}
\usepackage[T1]{fontenc}    %
\usepackage{hyperref}       %
\usepackage{url}            %
\usepackage{booktabs}       %
\usepackage{amsfonts}       %
\usepackage{nicefrac}       %
\usepackage{microtype}      %
\usepackage{xcolor}         %
\usepackage{xspace}
\usepackage{wrapfig}
\usepackage{graphicx}
\usepackage{tikz}
\usepackage{capt-of} %
\usepackage{subcaption}            %
\usepackage{bm}                    %
\usepackage{algorithm2e}           %
\usepackage{enumitem}              %
\usepackage{cleveref}
\usepackage{cancel}
\usepackage{titletoc}
\usepackage{comment}
\usepackage{ragged2e}
\usepackage{wrapstuff}

\usetikzlibrary{arrows.meta, positioning, calc, backgrounds, decorations.pathreplacing}

\newtheorem{definition}{Definition}[section]
\newtheorem{proposition}[definition]{Proposition}

\theoremstyle{definition} %
\newtheorem{example}{Example}

\newcommand{\name}{TNO\xspace}

\definecolor{sc}{rgb}{0.6, 0.30, 0.6}
\definecolor{seagreen}{rgb}{0.18, 0.55, 0.34}

\renewcommand{\paragraph}[1]{{\vspace{0.3mm}\noindent \bf #1}.}

\crefname{figure}{Fig.}{Figs.}
\Crefname{figure}{Fig.}{Figs.}
\crefname{table}{Tab.}{Tabs.}
\Crefname{table}{Tab.}{Tabs.}
\crefname{section}{Sec.}{Secs.}
\Crefname{section}{Sec.}{Secs.}
\crefname{equation}{Eq.}{Eqs.}
\Crefname{equation}{Eq.}{Eqs.}
\crefname{theorem}{Thm.}{Thms.}
\Crefname{theorem}{Thm.}{Thms.}
\crefname{remark}{Rem.}{Rems.}
\Crefname{remark}{Rem.}{Rems.}
\crefname{definition}{Dfn.}{Dfns.}
\Crefname{definition}{Dfn.}{Dfns.}
\crefname{algorithm}{Alg.}{Algs.}
\Crefname{algorithm}{Alg.}{Algs.}
\crefname{proposition}{Prop.}{Props.}
\Crefname{proposition}{Prop.}{Props.}

\title{Topological Neural Operators}

\author{
\large
\textbf{Lennart Bastian}\textsuperscript{1}\quad
\textbf{Samuel Leventhal}\textsuperscript{2}\quad
\textbf{Mustafa Hajij}\textsuperscript{$\dagger$,2}\quad
\textbf{Tolga Birdal}\textsuperscript{$\dagger$,1}
\\[0.75em]
\large
\textsuperscript{1}Imperial College London, UK
\quad
\textsuperscript{2}University of San Francisco, USA
}

\begin{document}

\maketitle
\renewcommand\thefootnote{$\dagger$}\footnotetext{Equal senior authorship.}
\renewcommand\thefootnote{\arabic{footnote}}

\begin{abstract}
We introduce \textbf{Topological Neural Operators} (TNOs), a principled framework for operator learning on cell complexes that lifts neural operators (NOs) from functions on points and/or edges to topological domains. 
TNOs represent data as features defined on cells of varying dimension and model their interactions through \emph{Discrete Exterior Calculus}, enabling explicit cross-dimensional coupling via gradient\mbox{-,} curl\mbox{-,} and divergence-type operators. 
The key design principle is to decouple \emph{where} information flows, as governed by fixed topological operators, from \emph{how} it is transformed (which is learned), yielding models that respect the geometric support of physical quantities and expose conservation and compatibility structure. 
We further propose \textbf{Hierarchical TNOs} (HTNOs), which incorporate learned coarse complexes to propagate long-range and topology-dependent information. 
Our framework recovers standard point- and graph-based NOs as restricted cases, providing a unified perspective on operator learning across discretizations. 
Across a range of PDE benchmarks, including irregular-geometry flow problems, TNOs and HTNOs improve accuracy; controlled studies further isolate the benefits of native higher-rank and topological structure.
\end{abstract}

\vspace{-3mm}
\section{Introduction}
\vspace{-2mm}
Modeling and predicting complex physical systems is a central challenge in science and engineering. Many systems, from fluid dynamics to electromagnetism and elasticity, are governed by \emph{partial differential equations} (PDEs), whose solutions under varying parameters, geometries, and boundary conditions underpin applications such as digital twins, design optimization, and real-time simulation~\citep{azizzadenesheli2024neural}. Classical numerical methods (e.g., finite elements, spectral solvers) are accurate but computationally costly, often requiring fine discretizations and multiple solves/iterations.

\textbf{Operator learning} addresses this by learning the solution itself: a map between infinite-dimensional function spaces that captures how PDE solutions depend on inputs such as coefficients, forcing terms, and geometry~\citep{boulle2024mathematical,kovachki2024operator,jiao2025one,lu2021learning}. \emph{Neural operators} (NOs)~\citep{kovachki2023neural} have emerged as a powerful instantiation of this idea, achieving strong performance and generalization across discretizations and resolutions via variants such as Fourier Neural Operators (FNOs)~\citep{li2021fourier,liu2023domain} and their geometric or transformer-based extensions~\citep{hao2023gnot,alkin2024upt}. These methods enable orders-of-magnitude speedups over classical solvers while maintaining high fidelity, establishing a scalable paradigm for scientific machine learning.

Despite these successes, existing NOs are fundamentally \emph{point-centric}: they represent all physical quantities as functions defined on nodes, with interactions modeled via learned kernels or graph-based message passing. This abstraction neglects a key structural aspect of continuum physics: \emph{physical quantities have geometric type} \cite{hirani2003discrete}. Potentials live on vertices, circulations on edges, fluxes on faces, and densities in volumes; their interactions are governed by differential operators like gradient, curl, and divergence, that couple quantities across dimensions. In compatible numerical discretizations, these relationships are encoded directly through incidence structures, ensuring conservation laws and identities (e.g., $\mathrm{div}\,\mathrm{curl}=0$) hold by construction\footnote{The boundary-of-boundary is zero: a boundary shape (like a compact/closed surface) has no boundary of its own.}. 
By collapsing all quantities to nodes, current neural operators obscure this structure, limiting their ability to model multi-physics systems, enforce conservation and compatibility, and capture topological effects such as cycles and harmonic modes.

\begin{figure}[t!]
  \centering
  \includegraphics[width=0.99\textwidth]{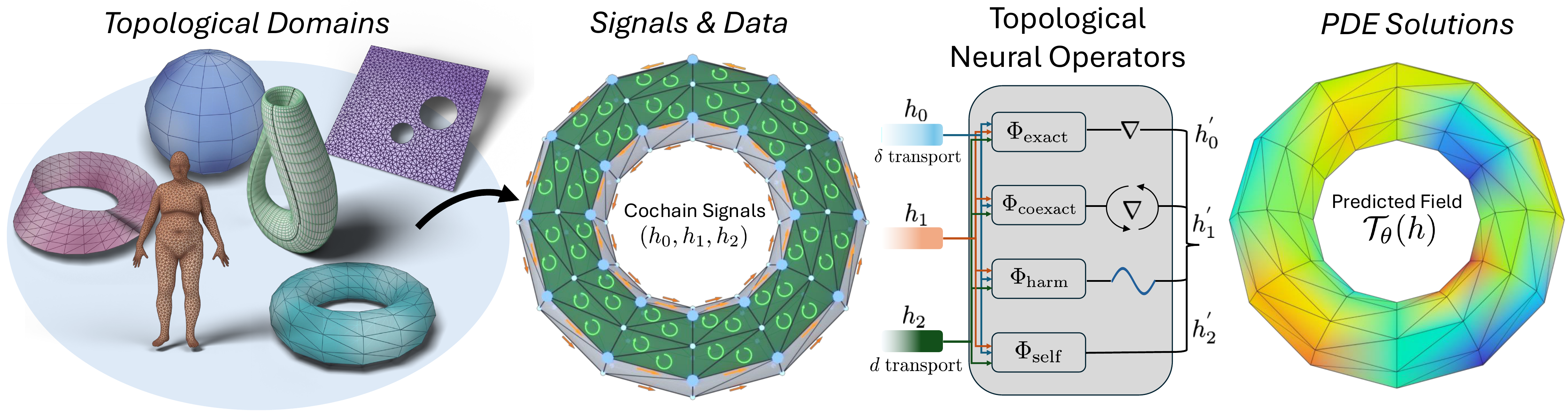}
  \caption{\textbf{Topological Neural Operators} operate on cell complexes (i) whose physical signals are cochains at multiple ranks (ii). A TNO layer (iii) couples ranks through fixed DEC operators ($d^{k}$, $\delta^{k}$) and learns the rank-wise channel mixing in four blocks, signifying gradient, curl, harmonic, and self maps, producing a predicted PDE field on the complex (iv).}
  \vspace{-5mm}
  \label{fig:TNO-teaser}
\end{figure}

As a remedy, we introduce \textbf{Topological Neural Operators (TNOs)}, a principled framework for operator learning on \emph{cell complexes}, the natural discrete domain for multi-dimensional physical systems (see \cref{fig:TNO-teaser}). Our key design principle is to separate where information flows, governed by fixed topological operators, from how it is transformed, which is learned. Rather than representing data solely on points (or edges), TNOs operate on \emph{cochains}, assigning features to cells of different dimensions (vertices, edges, faces, and higher-dimensional elements), and model their interactions using \emph{Discrete Exterior Calculus} (DEC)~\citep{hirani2003discrete}. This design explicitly encodes cross-dimensional structure: information flows between cells through incidence-based operators, while learnable components act on feature transformations. As a result, TNOs (i) respect the geometric support of physical quantities, (ii) incorporate conservation and compatibility structure into the architecture, and (iii) enable explicit modeling of multi-degree PDE systems.

This motivates a minimal extension of the NO paradigm: keep operator
learning, but replace rank-$0$ point fields by cochain fields supported on
cells of different dimensions. TNOs recover standard point-based NOs when
restricted to $0$-cochains, while allowing vertex-, edge-, face-, and
volume-supported fields to interact through DEC operators. To capture
long-range and topology-dependent effects, we further introduce
\textbf{Hierarchical TNOs} (HTNOs), which propagate information across
learned coarse complexes while preserving the cochain structure. In brief,
\textbf{our contributions are:}
\begin{itemize}[noitemsep,leftmargin=*,topsep=0em]
    \item A \textbf{framework for operator learning on cell complexes}, extending neural operators to cochain-valued fields and enabling explicit cross-dimensional modeling via DEC.
    \item A \textbf{unifying perspective} encapsulating existing NOs as special cases of the proposed framework.
    \item \textbf{Hierarchical HTNOs} for efficient multi-scale propagation of global and topological information.
    \item \textbf{Comprehensive ablation studies} validating the design choices, including PDEs with native multi-rank structure, which show that TNOs can model physical signals across cochain degrees.
    
\end{itemize}
We validate our approach across synthetic and real PDE benchmarks. (H)TNOs achieve improved accuracy and generalization across meshes, while exhibiting stronger physical consistency in structured settings. We will make our implementation publicly available upon publication at \href{https://circle-group.github.io/research/TNO}{the project page}.

\vspace{-2mm}
\section{Background and the Language of Discrete Topology}
\label{sec:background}
\vspace{-2mm}
Physical quantities are dimensional: scalars live on vertices, 
circulations on edges, fluxes on faces, densities in volumes. 
The laws relating them are statements about boundaries, orientations, and dimensional couplings; this is the language of discrete exterior calculus used throughout. 
We provide the necessary background and refer the reader to~\citep{crane2013digital,crane2018discrete,hirani2003discrete,desbrun2005discrete,meyer2003discrete,desbrun2006discrete} for a thorough treatment.

\vspace{-1mm}
\subsection{Topological Structures}
\vspace{-1mm}
\begin{definition}[Regular Cell Complex]
A \emph{regular cell complex} is a finite set $K$ together with a
rank function $\mathrm{rk}: K \to \mathbb{Z}_{\geq 0}$ and a
\emph{face relation} $\sigma \prec \tau$, defined only when
$\mathrm{rk}(\tau) - \mathrm{rk}(\sigma) = 1$, such that for any
$\tau, \rho \in K$ with $\mathrm{rk}(\tau) - \mathrm{rk}(\rho) = 2$,
exactly two elements $\sigma \in K$ satisfy both $\rho \prec \sigma$
and $\sigma \prec \tau$.
\end{definition}

Elements of $K$ are called \emph{cells}; those of rank $k$ form the
set $K_k$, with $n_k = |K_k|$ and $N = \max_{\tau} \mathrm{rk}(\tau)$.
Rank-$0$ cells are vertices, rank-$1$ edges, rank-$2$ faces, and rank-$3$ volumes. 
This structure is the correct domain language for
physics: quantities such as pressure, circulation, magnetic flux, and charge density live on cells of rank $0,1,2,3$ respectively (see \cref{tab:cochains}). 
Every $k$-cell is a first-class computational entity able to exchange information with any incident cell.
As we show in~\cref{sec:pdes}, the governing
equations of Maxwell, Navier--Stokes, and elasticity are precisely statements about how quantities at different ranks interact, a
structure graphs cannot represent and cell complexes encode natively. 
In the following, we abbreviate regular cell complexes as RCC. 

The face relation $\sigma \prec \tau$ records immediate incidence
and determines differential operators. For every $k$-cell $\tau$ and
$(k-1)$-cell $\sigma$ one assigns an orientation sign
$[\tau:\sigma] \in \{-1,0,+1\}$, with $[\tau:\sigma]\neq 0$ iff
$\sigma \prec \tau$. Assembling these into the \emph{signed incidence
matrix} $B_k \in \mathbb{R}^{n_{k-1}\times n_k}$ with
$(B_k)_{\sigma\tau} = [\tau:\sigma]$, \textit{ the diamond condition} \cite{basak2010combinatorial,aschbacher1996combinatorial,savoy2022combinatorial} forces
\begin{equation}
    B_k B_{k+1} = 0 \quad \forall\,k, \label{eq:bdbd}
\end{equation}
the discrete statement that \textbf{the boundary of a boundary is
empty}, which underlies $\mathrm{curl}(\mathrm{grad})=0$ and
$\mathrm{div}(\mathrm{curl})=0$, and is the structural foundation
of everything that follows.

A \emph{$k$-cochain} $u^k:K_k\to\mathbb{R}^{d_k}$ assigns a feature 
vector to each $k$-cell; the cochain space is $C^k(K;\mathbb{R}^{d_k})
\cong\mathbb{R}^{n_k\times d_k}$. The degree $k$ is geometrically 
meaningful: it records the type of object on which the quantity lives, and assigning a field to the wrong degree 
breaks the structural laws governing it (see \cref{app:derham-cochains}). The full multirank signal space 
is $C^\bullet=\bigoplus_{k=0}^N C^k$.

In practice, we also use more flexible Combinatorial Complexes~\citep{hajij2023topological,hajij2023combinatorial}, which generalize graphs, RCCs, and hypergraphs by allowing cells at different ranks to be specified independently rather than constrained to arise as boundaries of higher-rank cells. We introduce these in~\cref{sec:tno_main}.

\vspace{-1mm}
\subsection{Discrete Exterior Calculus}
\label{sec:DEC}
\vspace{-1mm}
From the boundary matrices, one derives three families of operators that serve as the computational primitives of the Discrete Exterior Calculus (DEC) framework \cite{crane2013digital,crane2018discrete,desbrun2006discrete}.

\paragraph{Discrete exterior derivative}
The \emph{discrete exterior derivative} $d^k = B_{k+1}^\top : C^k \to C^{k+1}$ lifts a $k$-cochain to a $(k+1)$-cochain by accumulating signed values over incidence relations. Concretely, $d^0$ is the discrete gradient (oriented differences across edges), $d^1$ is the discrete curl (oriented sums around faces), and $d^2$ is the discrete divergence-of-dual (oriented sums over bounding faces of a volume).  The identity $d^{k+1} \circ d^k = 0$ follows directly from~\cref{eq:bdbd} and is the discrete counterpart of $\mathrm{curl}(\mathrm{grad})=0$ and $\mathrm{div}(\mathrm{curl})=0$. See Section \ref{app:derham-cochains} for details.

\paragraph{Hodge operators and codifferential} Equipping each $C^k$ with a positive-definite \emph{Hodge star}\footnote{See Section \ref{app:derham-cochains} for the precise definition.} $M_k \in \mathbb{R}^{n_k \times n_k}$ encoding geometric data such as edge lengths, face areas, and cell volumes, gives the formal $M_k$-adjoint of $d^{k-1}$, the \emph{codifferential}, mapping downward in degree:
\begin{equation}
\delta^k = M_{k-1}^{-1} B_k M_k : C^k \to C^{k-1}.    
\end{equation}
While $d^k$ is purely topological (determined by $B_{k+1}$ alone), $\delta^k$ depends on geometry encoded in $M_k$.

\paragraph{Hodge Laplacian} Together, $d^k$ and $\delta^k$ yield the \emph{$k$-Hodge Laplacian}
\begin{equation}
    \Delta_k = \underbrace{\delta^{k+1} \circ d^k}_{\Delta_k^{\uparrow}} + \underbrace{d^{k-1} \circ \delta^k}_{\Delta_k^{\downarrow}} : C^k \to C^k, \label{eq:hodge-lap}
\end{equation}
which is self-adjoint and positive semidefinite with respect to $M_k$. The decomposition into $\Delta_k^\uparrow$ and $\Delta_k^\downarrow$ is physically meaningful: $\Delta_k^\uparrow$ couples $u^k$ upward to $(k+1)$-cells via $d^k$ (curl-like coupling), while $\Delta_k^\downarrow$ couples $u^k$ downward to $(k-1)$-cells via $\delta^k$ (divergence-like coupling). 
At $k=0$, $\Delta_0 = \delta^1 d^0$ recovers the weighted graph Laplacian; at $k=1$, $\Delta_1$ simultaneously involves edges, vertices, and faces, and thus already an operator with no graph-level analog \cite{crane2018discrete}.

\paragraph{Hodge Decomposition (HD)}
The three operator families organize every cochain space into an orthogonal decomposition with respect to the $M_k$-inner product:
\begin{equation}
    C^k(K;\mathbb{R}) = \underbrace{\mathrm{im}(d^{k-1})}_{\text{exact}} 
    \oplus \underbrace{\ker(\Delta_k)}_{\text{harmonic}} \oplus 
    \underbrace{\mathrm{im}(\delta^{k+1})}_{\text{coexact}}, 
    \label{eq:hodge-decomp}
\end{equation}
Exact cochains are 
\textit{potential-driven}; coexact cochains are \textit{divergence-free}; harmonic cochains are \textit{topologically constrained}, with
\begin{equation}
\dim\ker(\Delta_k)=\beta_k
\end{equation}
where $\beta_k$ is the $k$-th Betti number, counting independent $k$-dimensional holes \cite{hirani2003discrete,arnold2006finite}.

Hodge Decomposition is central to our TNO for two reasons: it characterizes what physical
solutions look like (incompressible flows are coexact; conservative
fields are exact; topological modes are harmonic), and it provides the
DEC-compatible channels through which conservation and compatibility
constraints can be preserved, as we show in~\cref{sec:tnn}.

The complex $K$ thus encodes in a single structure everything a learning framework needs: the domain $\{K_k\}_{k=0}^N$ stratified by dimension, the signal spaces $\{C^k(K;\mathbb{R}^{d_k})\}$ where physical fields live, the differential operators $d^k = B_{k+1}^\top$, $\delta^k = M_{k-1}^{-1}B_k M_k$, and $\Delta_k = \delta^{k+1}d^k + d^{k-1}\delta^k$ that move information between dimensions, and the identity $B_k B_{k+1} = 0$ that encodes the algebraic backbone of discrete physical laws.

\vspace{-1.5mm}
\section{Discrete PDEs on Cell Complexes}
\label{sec:pdes}
\vspace{-1.5mm}
The operators $d^k$, $\delta^k$, $\Delta_k$ appear directly in the
governing equations of physical systems, and most of these systems
couple fields across more than one degree.

As in~\cref{sec:background}, let $K$ be a finite cell complex of dimension
$N$ equipped with Hodge stars $\{M_k\}_{k=0}^{N}$. The de~Rham complex gives
a common discrete language for physical PDEs on bounded domains: a field is
stored on the cells where it naturally lives, while $d^k$, $\delta^k$, and
$\Delta_k$ describe how that field changes across neighboring dimensions
\cite{desbrun2006discrete,arnold2006finite}. Thus scalar potentials live on
vertices, circulations on edges, fluxes on faces, and densities on volumes.

A \emph{discrete PDE on $(K,\{M_k\})$ at degree $k$} is an equation
\begin{equation}
    \mathcal{F}\!\left(u^k,\; d^k u^k,\; \delta^k u^k,\; \Delta_k u^k,\;
    a,\; f\right) = 0 \quad \text{in } C^k(K;\mathbb{R}^d),
    \tag{PDE}
    \label{eq:pde}
\end{equation}
where $u^k\in C^k(K;\mathbb{R}^d)$ is the unknown field and
$f\in C^k(K;\mathbb{R}^d)$ is the source. The coefficient field
$a\in C^\bullet(K;\mathbb{R}^p)$ contains the material or constitutive data
that define the particular PDE instance, such as diffusion coefficients,
conductivity, permeability, permittivity, or density. It may live on one
degree or across several degrees, depending on the physics. The map
$\mathcal{F}$ is the PDE residual; solving the PDE means finding $u^k$ so
that this residual vanishes.

Boundary conditions are imposed on a subcomplex $\partial K\subseteq K$. 
Essential conditions prescribe the boundary trace $u^k|_{\partial K}$, while natural conditions prescribe flux-type data and enter through the Hodge-adjoint codifferential $\delta^k$, hence through restricted incidence and Hodge-star operators.

Many physical systems involve more than one degree at once: Maxwell equations couple a 1-form with a 2-form, Darcy couples a 0-form with a 1-form, Navier--Stokes in vorticity form couples 1- and 2-forms. 
A \emph{multi-degree discrete PDE} is a system
\begin{equation}
    \mathcal{F}_k\!\left(\{u^j\}_{j=0}^{N},\;
    \{d^j u^j\}_j,\; \{\delta^j u^j\}_j,\; a,\; f\right) = 0
    \quad \text{in } C^k(K;\mathbb{R}^{d_k}),\quad k=0,\ldots,N,
    \tag{mPDE}
    \label{eq:mpde}
\end{equation}
with unknown $u\in C^\bullet(K;\mathbb{R}^{d_\bullet})$. The equation
at degree $k$ can depend on $d^{k-1}u^{k-1}$ (information from below,
via $\Delta_k^\downarrow$-type coupling), on $\delta^{k+1}u^{k+1}$
(from above, via $\Delta_k^\uparrow$-type coupling), and on
$\Delta_k u^k$. The coupling pattern is fixed by the cellular structure
of $K$; the weights, by $\{M_k\}$. A learnable map that processes
degrees independently, without $d$ and $\delta$ connecting them, cannot
represent this.

For time-dependent problems, the unknown is $u:[0,T]\to
C^\bullet(K;\mathbb{R}^{d_\bullet})$ satisfying
\begin{equation}
    \partial_t u^k(t) = \mathcal{A}_k\!\left(\{u^j(t)\}_{j=0}^{N},\;
    \{d^j u^j(t)\}_j,\; \{\delta^j u^j(t)\}_j,\; a,\; f(t)\right),
    \qquad u(0)=u_0,
    \tag{Evol}
    \label{eq:evol}
\end{equation}

encompassing parabolic/hyperbolic problems, and fully coupled multi-physics
systems. 
Discretizing in time yields one-step maps $u(t)\mapsto
u(t+\Delta t)$, the natural targets for time-stepping operators.

\paragraph{Examples}
In the~\cref{app:pde-examples}, we illustrate \eqref{eq:evol} with two standard systems whose fields live on different cochain degrees and couple through $d$ and $\delta$: discrete Maxwell- and wave-type couplings. 
In both cases, the dynamics are incidence-driven: fields on different cochain degrees are coupled by $d$ and $\delta$, and the associated constraints are preserved by the algebraic identities of the complex.

\vspace{-1.5mm}
\section{Topological Neural Operators (TNOs)}
\label{sec:tno_main}
\vspace{-1.5mm}
Each PDE admits a
\emph{solution operator} sending input data (initial conditions,
sources, parameters) to the output field: for the wave system, a map
$C^0\times C^1 \to C^0\times C^1$ advancing $(u, p)$ in time; for
Maxwell, a map $C^1\times C^2 \to C^1\times C^2$ advancing $(E, B)$.
Approximating such an operator $\mathcal{G}$ from data is the subject
of operator learning~\citep{kovachki2023neural}.
Here $\mathcal{G}$ acts on tuples of cochains organized by degree, with its action mediated by $d$, $\delta$, $\Delta$ on $K$.
The following defines our class of learnable maps:

\begin{definition}[Topological Neural Operators]
\label{def:tno}
Let $K$ be a finite cell complex of dimension $N$ with boundary
matrices $\{B_k\}$ and Hodge stars $\{M_k\}_{k=0}^{N}$, inducing
discrete operators $\{d^k\}$, $\{\delta^k\}$, $\{\Delta_k\}$. For
$i=1,\ldots,m$ and $j=1,\ldots,n$, let $k_i,\ell_j\in\{0,\ldots,N\}$
and $d_i, r_j\geq 1$, and define
\begin{equation}
\mathcal{X}(K) = \prod_{i=1}^{m} C^{k_i}(K;\mathbb{R}^{d_i}),
\qquad
\mathcal{Y}(K) = \prod_{j=1}^{n} C^{\ell_j}(K;\mathbb{R}^{r_j}).
\end{equation}
A \emph{Topological Neural Operator (TNO)} is a parameterized family of neural networks
$\{\mathcal{T}_\theta^{K}:\mathcal{X}(K)\to\mathcal{Y}(K)\}_{\theta\in\Theta}$
approximating a target operator
$\mathcal{G}:\mathcal{X}(K)\to\mathcal{Y}(K)$ and satisfying:

\begin{enumerate}[label=\textbf{(P\arabic*)}, leftmargin=*, nosep]
\item \textbf{Cellular intrinsicness.}
$\mathcal T_\theta^K$ is built from the cellular operators of $K$,
including boundary / coboundary maps, Hodge stars, codifferentials, Hodge Laplacians, trace / restriction maps, pointwise nonlinearities,
and learned channel-mixing maps shared across cells of the same degree.
Thus it depends on incidence and metric data, not on a fixed grid or cell ordering.

\item \textbf{Multi-degree coupling.}
Information may move between cochain degrees through boundary,
coboundary, codifferential, Laplacian, and learned composed operators.
Thus an output at one degree may depend on inputs at any other degree.

\item \textbf{Discretization transferability.}
The weights $\theta$ are shared across discretizations. If $K'$ refines
$K$, then lifting to $K'$ (prolonging cochains via $P_{K\to K'}$), applying the operator, and restricting back (via $R_{K'\to K}$)
should agree with applying the operator on $K$:
\begin{equation}
R_{K'\to K}\mathcal T_\theta^{K'}P_{K\to K'}x
\approx
\mathcal T_\theta^K x.
\end{equation}
\end{enumerate}
\end{definition}

\paragraph{Remarks}
(\textbf{P1}) makes the TNO intrinsic to the cellular discretization: incidence enters through the boundary maps, while geometric information enters through the Hodge stars, not through an extrinsic grid or an ordering of cells. (\textbf{P2}) is the key higher-order feature:
signals may live on vertices, edges, faces, and higher cells, with $\mathcal T_\theta$ coupling these degrees. 
Such cross-degree coupling is not merely an architectural extension, but is imposed by the physics (e.g., Maxwell/wave PDEs in~\cref{app:pde-examples}).
(\textbf{P3}) is the analogue of discretization consistency in NOs~\citep{kovachki2023neural}: shared weights act across refinements. %

The choice of degrees $\{k_i\}$ and $\{\ell_j\}$ in~\cref{def:tno} determines the operator type:
degree-preserving ($C^k\to C^k$) for single-field PDEs;
gradient-type ($C^0\to C^1$) for potential-to-flux maps like
Darcy or Fourier; divergence-type ($C^1\to C^0$) for flux-to-source maps; and multi-degree
($C^0\times C^1\times C^2\to C^0\times C^1\times C^2$) for coupled systems like Maxwell. We next detail how $\mathcal{T}_\theta$ is computed (\emph{learned}).

\vspace{-0.5mm}
\subsection{Realization as a Topological Neural Network}
\label{sec:tnn}
\vspace{-0.5mm}
\cref{def:tno} specifies what a TNO must be (P1--P3); it does not say how to compute one. 
We now present one possible architecture that realizes these axioms as a topological neural network (TNN)~\citep{hajij2023topological}. 

\paragraph{Architecture}
The TNO of~\cref{def:tno} is posed on an RCC $K$.
For computation, it can be convenient to lift $K$ to a combinatorial complex (CC) rather than an RCC (see \cref{app:background}).
A TNN realizing $\mathcal{T}_\theta^K : \mathcal{X}(K) \to \mathcal{Y}(K)$ has the form
\begin{equation}
\mathcal{T}_\theta^K
=
\mathrm{Dec}_\theta
\circ
\mathcal{L}_\theta^{(L-1)}
\circ \cdots \circ
\mathcal{L}_\theta^{(0)}
\circ
\mathrm{Enc}_\theta.
\end{equation}
The encoder lifts the input cochains on $K$ to hidden cochains $h_0^k \in C^k(\widetilde{K}; \mathbb{R}^{d_h})$ at every rank $k = 0, \ldots, \dim \widetilde{K}$. 
The incidence structure of $\widetilde{K}$ supplies the discrete operators used in the network.
The topological layers $\mathcal{L}_\theta^{(\ell)}$ update these cochains on $\widetilde{K}$ while preserving their cochain-valued structure.
The decoder reads the target degrees from the final hidden cochains and restricts to $K$. 

The encoder and decoder bridge between $K$ and $\widetilde{K}$.
Material parameters and other coefficient fields enter as input cochains, which keep the architecture intrinsic to $(\widetilde{K}, \{M_k\})$ in the sense of (P1). Next, we explain how the individual topological layers $\mathcal{L}_\theta^{(\ell)}$ are realized.

\vspace{-0.5mm}
\subsection{Topological layers}
\label{sec:tno-layers}
\vspace{-0.5mm}

\begin{wrapfigure}[6]{r}{.5\textwidth}
\vspace{-8mm}
\begin{equation}
\mathbf{H}_{\mathrm{out}}
=
\phi_\theta\!\left(
\begin{bmatrix}
\Delta_0 & \delta^1 & 0 & \cdots \\
d^0 & \Delta_1 & \delta^2 & \cdots \\
0 & d^1 & \Delta_2 & \ddots \\
\vdots & \vdots & \ddots & \ddots
\end{bmatrix}
\mathbf{H}W
\right)
\nonumber
\end{equation}
\end{wrapfigure}
Each topological layer updates the hidden cochains on the lifted complex $\widetilde{K}$ while preserving their geometric type. 
Let $\mathbf{H} = (H^0, \dots, H^N)$ with $H^k \in \mathbb{R}^{n_k \times d_h}$. 
The layer applies a learned nonlinear map $\phi_\theta$ to a fixed block-structured operator that encodes the discrete exterior calculus of $\widetilde{K}$. As shown on the right, $\mathbf{H}_{\mathrm{out}}$ is
defined through incidence matrices and Hodge stars of $\widetilde{K}$. Its tridiagonal sparsity is exact: equations at degree $k$ couple only ranks $k\pm 1$. 
The trainable objects are the channel-mixing matrices $W$ and transformations $\phi_\theta$. 
In the rigid DEC version, the transport maps themselves are fixed; in the copresheaf variant \cite{hajij2025copresheaf} below, the incidence support remains fixed, but the fiber maps carried along incidences may be learned. See Appendix \ref{app:cellwise-tno}.

For a given output rank $k$, the rank-$k$ block update is (out-of-range terms omitted):
\begin{equation}
H^k_{\mathrm{out}}
=
\phi_\theta\!\Bigl(
d^{k-1}H^{k-1}W^\downarrow_k,\;
\Delta_k^\uparrow H^k W^{\Delta,\uparrow}_k,\;
\Delta_k^\downarrow H^k W^{\Delta,\downarrow}_k,\;
H^k,\;
\delta^{k+1}H^{k+1}W^\uparrow_k
\Bigr).
\label{eq:tl-single}
\end{equation}
\begin{wrapfigure}[16]{r}{0.5\textwidth}
\vspace{-0.5em}
\centering
\includegraphics[width=0.50\textwidth]{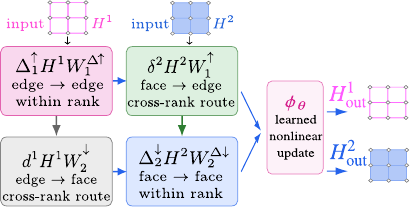}
\caption{
Visual TNO layer induced by Maxwell-type routing. Cross-rank transport
is carried by $d^1$ and $\delta^2$, while same-rank propagation is
carried by the Hodge-Laplacian channels
$\Delta^\uparrow$ and $\Delta^\downarrow$. The learned weights
$W^\bullet$ and nonlinear update $\phi_\theta$ determine how strongly the routed features are mixed.
}
\label{fig:maxwell-tno-layer}
\vspace{-1.2em}
\end{wrapfigure}
\cref{fig:maxwell-tno-layer} illustrates the template nature of TNOs: Maxwell fixes the active degrees and DEC routes, while the network learns how to mix the routed edge and face cochains (see App. \ref{sec:pde_templates}). The construction admits a continuous \textit{rigidity spectrum}. At the rigid extreme, the transport operators are exactly the DEC operators of $\widetilde{K}$. 
At the flexible extreme, each incidence $y \prec x$ carries a learnable fiber map $\rho_{y \to x}$, yielding twisted operators $d_\rho$, $\delta_\rho$, $\Delta_{\rho,k}$ (copresheaf realization). 

\vspace{1mm}

\noindent\textbf{Operator-learning interpretation}
The distinction from a generic CCNN layer \citep{hajij2023topological} is that the support of spatial transport is prescribed by the DEC operators. 
Cross-rank coupling is carried by $d$ and $\delta$, while same-rank propagation is carried by $\Delta^\uparrow$ and $\Delta^\downarrow$. 
Thus, the architecture does not learn where information can flow; it learns how transported features are mixed through $W^\bullet$ and $\phi_\theta$.
This is the architectural mechanism behind discretization transfer.
Under compatible refinements and stable Hodge discretizations, the DEC operators $d,\delta,\Delta$ approximate their continuum counterparts in the FEEC sense~\citep{arnold2018feec}. 
Since the same weights $\theta$ are shared across discretizations with support maps recomputed from each cell complex, the resulting family $\{\mathcal{T}_\theta^K\}_K$ is biased toward a common continuum rather than a discretization-specific message passing rule.

\paragraph{Linear realization of TNO} The simplest special case takes $\phi_\theta$ to be linear and uses the Hodge decomposition channels directly. The rank-$k$ linear TNO layer is
\begin{equation}
\label{eq:tno-linear}
\mathcal{T}_{\theta,k}(h)
=
\delta^{k+1}h^{k+1}\,W_k^{\uparrow}
+
d^{k-1}h^{k-1}\,W_k^{\downarrow}
+
P_k^{\mathrm{harm}}h^k\,W_k^{\mathrm{harm}}
+
h^k\,W_k^{\mathrm{self}} .
\end{equation}
The four terms correspond to coexact, exact, harmonic, and local channels: $\mathrm{im}(\delta^{k+1})$, $\mathrm{im}(d^{k-1})$, $\ker(\Delta_k)$, and a residual self-channel, respectively. Here $P_k^{\mathrm{harm}}$ denotes the $M_k$-orthogonal projection onto the harmonic subspace $\ker(\Delta_k)$. 
Thus, the layer separates the three topological components of a rank-$k$ signal while keeping transport fixed by the DEC operators. 
The following propositions show how FNO and GNO arise as rank-$0$ specializations of the TNO framework.

\begin{proposition}[FNO as a rank-0 spectral TNO ~\citep{li2021fourier}]
\label[proposition]{prop:fno-recovery}
Let \(K_N\) be a uniform periodic cubical complex with \(N_1\times\cdots\times N_d\)
vertices, and let \(u\in C^0(K_N;\mathbb R^{c_{\rm in}})\). Let
\(\mathcal F_N\) denote the discrete Fourier transform on \(0\)-cochains.  
A linear FNO layer
\begin{equation}
u \longmapsto Wu+\mathcal F_N^{-1}R_\theta \mathcal F_N u
\end{equation}
is realized by a rank-\(0\) spectral TNO layer whose same-rank spectral channel
is diagonal in the Fourier basis of \(K_N\). 
\emph{Proof is provided in~\cref{app:proof-fno}.}
\end{proposition}

\begin{proposition}[GNO as a rank-$0$ graph-quadrature TNO~\citep{li2020gno}]
\label[proposition]{prop:gno-recovery}
On a directed graph complex whose vertices are sample points with
rank-$1$ cells the quadrature neighborhoods, the GNO update :
\begin{equation}
u_i \mapsto Wu_i+\sum_{j\in\mathcal N(i)}
\kappa_\theta(x_i,x_j)u_j\omega_j,
\end{equation}
is recovered as a rank-$0$ TNO with a learned incidence-supported kernel
message $h^1(e_{ij})=\kappa_\theta(x_i,x_j)u_j\omega_j$ on each directed rank-$1$ cell $e_{ij}:v_j\to v_i$, followed by
target-incidence aggregation at each vertex.
\emph{Proof is provided in~\cref{app:proof-gno}.}
\end{proposition}

\paragraph{Hierarchical Realization of TNO (HTNO)}
The TNO construction extends naturally to a hierarchy of cell complexes. Let
$K_0 \leftarrow K_1 \leftarrow \cdots \leftarrow K_L$, where
$K_{\ell+1}$ is a coarsening of $K_\ell$ and $K_0=K$. Each level carries
cochain spaces $C^k(K_\ell)$ and discrete exterior derivatives
$d^k_\ell:C^k(K_\ell)\to C^{k+1}(K_\ell)$. Levels are connected by
degree-preserving transfer maps
\begin{equation}
    R^k_\ell:C^k(K_\ell)\to C^k(K_{\ell+1}),
    \qquad
    \Pi^k_\ell:C^k(K_{\ell+1})\to C^k(K_\ell).
\end{equation}
Ideally, these transfers commute with the coboundary:
\begin{equation}
    d^k_{\ell+1}R^k_\ell=R^{k+1}_\ell d^k_\ell,
\qquad
    d^k_\ell\Pi^k_\ell=\Pi^{k+1}_\ell d^k_{\ell+1}.
\label{eq:htno-commute}
\end{equation}
This is the analogue of the commuting-diagram condition in FEEC multigrid
~\citep{arnold2006finite,arnold2018feec}. 
When \eqref{eq:htno-commute} holds, exact components are preserved across levels, so transfer respects the de~Rham structure rather than mixing cochain types arbitrarily. Coexact and harmonic components are controlled through the corresponding Hodge-adjoint and Hodge-Laplacian structure induced by the level-wise Hodge stars. For learned coarsenings, such as soft-clustering or soft Voronoi partitions, exact commutation is generally not guaranteed; in that case, \eqref{eq:htno-commute} acts rather as an architectural bias or regularization.

An HTNO arranges TNO blocks as a learned $V$-cycle for de Rham complexes~\citep{hiptmair1998multigrid,trottenberg2000multigrid}: TNO layers on $K_\ell$ are Hodge-compatible smoothers~\citep{hiptmair2002finite}, with an additive coarse-grid correction transferred via $R^k_\ell,\Pi^k_\ell$ that come from a partition of $K_\ell$ — either fixed ($k$-means) or learned end-to-end as a soft Voronoi with trainable centroids. 
The operator class is unchanged: it is still a TNO at every level. (P1)--(P3) hold level-wise — with fine blocks capturing local cross-degree physics and coarse blocks propagating long-range information.
\cref{app:multigrid} gives the $V$-cycle equation and implementation.

\vspace{-1.5mm}
\section{Experiments}
\label{sec:experiments}

\vspace{-1.5mm}
We evaluate \name{} and HTNO on a broad span of steady-state PDEs over irregular 2D and 3D meshes:
Poisson's equation with Gaussian and multiscale-sinusoidal sources, hyper-elastic deformations of variable domains, compressible flow past 2D airfoils spanning subsonic, transonic, and supersonic regimes, and large-scale 3D wing-surface aerodynamics under compressible RANS in the transonic regime.
These come from three established public suites ~\cite{mousavi2025rigno,hao2024gaot,paischer2025going}, with mesh sizes ranging from ${\sim}1$K to 65K nodes.
We complement these with two controlled studies of our own: an anisotropic-Darcy experiment with per-face random tensor orientation that isolates how \name{} natively ingests higher rank signals ~(\cref{sec:rankin-aniso}), and a component ablation on synthetic topologies (Darcy with reaction; conservative advection--diffusion) that disentangles the harmonic-basis input and sheaf transport~(\cref{sec:ablations}).
Datasets descriptions and full training protocol are deferred to~\cref{app:experimental-details}.
Implementations are in JAX and run on GH200 chips (96GB VRAM).

\vspace{-1.5mm}
\subsection{Steady-State Benchmarks on Irregular Geometry}
\vspace{-1.5mm}
\label{sec:steady-state}

\begin{table}[t]
  \centering
  \vspace{-2.5mm}
  \caption{Established steady-state benchmarks~\cite{li2023fourier,herde2024poseidon} as used in~\cite{mousavi2025rigno}. We depict test relative $L^1$~(\%) error.
  \textbf{Bold}: best; \underline{underline}: second.
  $^\dagger$ our runs with official implementations (see \cref{app:gaot-splits}).}
  \label{tab:rigno-steady-state}
  \small
  \setlength{\tabcolsep}{4pt}
  \resizebox{\linewidth}{!}{
  \begin{tabular}{@{}l cc c cccccccc@{}}
    \toprule
    & \multicolumn{2}{c}{\textit{Ours}} & & \multicolumn{8}{c}{\textit{Baselines}} \\
    \cmidrule(lr){2-3} \cmidrule(l){5-12}
    Dataset & HTNO & \name{} & & RIGNO-18 & RIGNO-12 & GAOT$^\dagger$ & MGN & Geo-FNO & FNO~DSE & GINO & UPT \\
            &      &         & & {\scriptsize\cite{mousavi2025rigno}} & {\scriptsize\cite{mousavi2025rigno}} & {\scriptsize\cite{hao2024gaot}} & {\scriptsize\cite{pfaff2021learning}} & {\scriptsize\cite{li2023fourier}} & {\scriptsize\cite{lingsch2024beyond}} & {\scriptsize\cite{li2023geometry}} & {\scriptsize\cite{alkin2024upt}} \\
    \midrule
    Poisson-Gauss & \underline{1.30} & \textbf{1.03}    & & 2.26          & 2.52             & 1.39 & 30.9 & 8.16 & 2.27 & 7.57 & 48.4 \\
    Airfoil       & 1.21             & 1.11             & & \textbf{1.00} & \underline{1.09} & 3.12 & 10.1 & 4.48 & 1.99 & 2.00 & 45.7 \\
    Elasticity    & \textbf{1.70}    & \underline{2.73} & & 4.31          & 4.63             & 4.07 & 11.9 & 5.53 & 4.81 & 4.38 & 12.6 \\
    \bottomrule
  \end{tabular}
  }
  \vspace{-5.0mm}
\end{table}

\paragraph{Poisson-Gauss, Airfoil, Elasticity (\cref{tab:rigno-steady-state})}
We evaluate on three established steady-state benchmarks assembled by~\cite{mousavi2025rigno}: Poisson-Gauss~(PG, fixed grid; from~\cite{herde2024poseidon}), Airfoil Flow~(AF, variable geometry), and Elasticity~(variable; both from~\cite{li2023fourier}).
We compare against the seven baselines reported in~\cite{mousavi2025rigno}: RIGNO-18, RIGNO-12, MeshGraphNet~(MGN)~\cite{pfaff2021learning}, Geo-FNO~\cite{li2023fourier}, FNO~DSE~\cite{lingsch2024beyond}, GINO~\cite{li2023geometry}, and UPT~\cite{alkin2024upt}. (see \cref{tab:rigno-steady-state}).
Both \name{} and HTNO outperform all seven on PG and Elasticity; 
TNO is competitive with RIGNO-18 on AF, which fixes the far-field at $M_\infty{=}0.8$, $\alpha{=}0^\circ$ and varies only the airfoil geometry~\cite{li2023fourier}, so methods saturate near the discretization noise floor; the GAOT airfoils (\cref{tab:gaot-suite}) genuinely sweep $M\in[0.5,1.4]$ and $\alpha\in[0.5^\circ,5.0^\circ]$ per sample~\cite{hao2024gaot}, crossing the shock-formation boundary (see \cref{app:af-vs-gaot-regime}).

\begin{figure}[t]
  \vspace{-4.5mm}
  \centering
  \includegraphics[width=\linewidth]{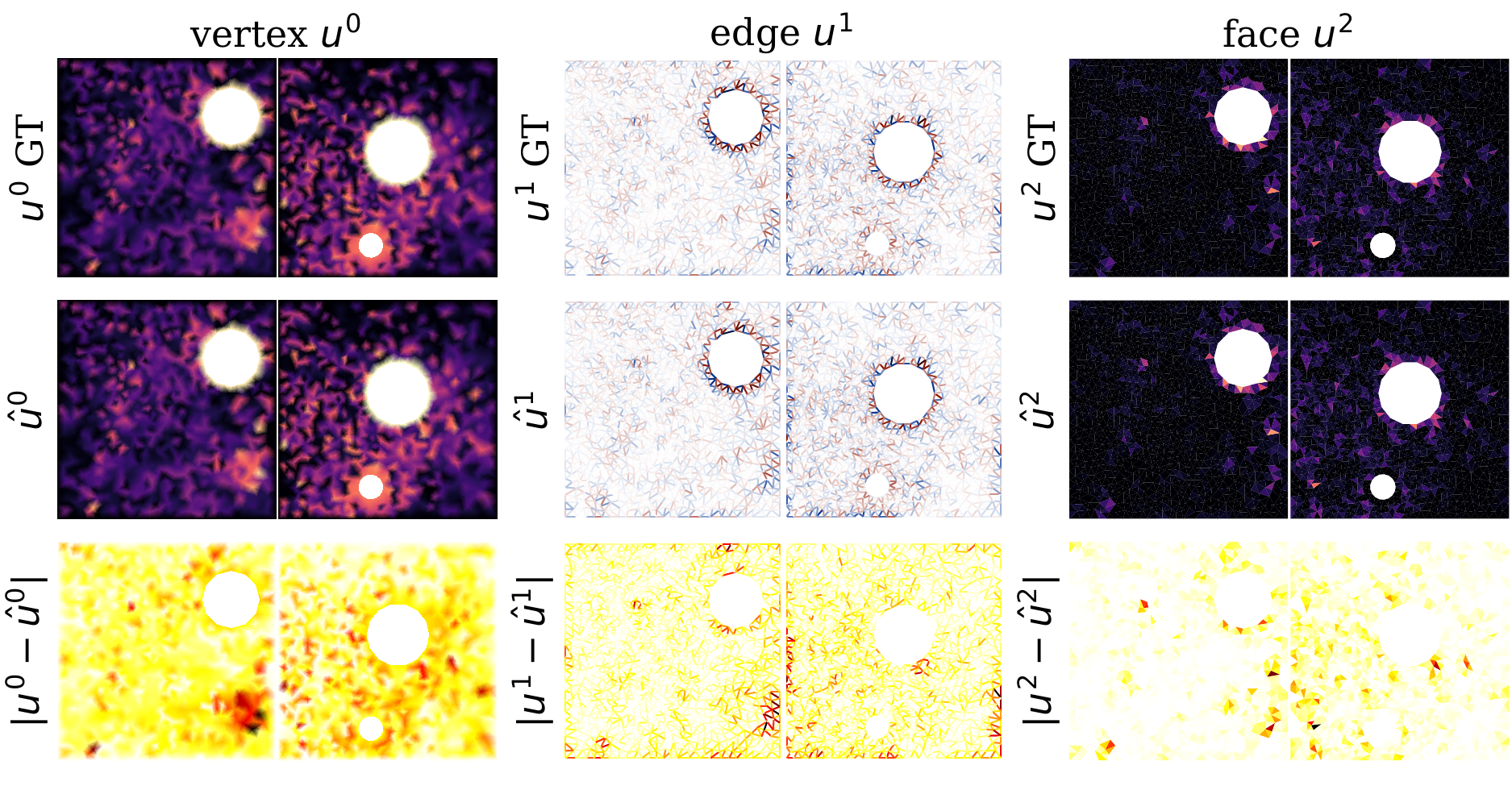}
  \vspace{-6.0mm}
  \caption{\textbf{Qualitative results for Anisotropic Darcy} with per-face random tensor orientations. Coupled signals require disentangling physical quantities at multiple topological ranks simultaneously.}
  \label{fig:darcy-random-face}
  \vspace{-5.5mm}
\end{figure}

\begin{table}[t]
  \centering
  \caption{Additional steady-state benchmarks: test relative $L^1$~(\%).
  $^\dagger$Runs with official implementation; \textbf{bold}: best, \underline{underline}: second. EmmiWing is given per-variable ($p_s$, $\tau_{x,y,z}$) and aggregate (Agg).}
  \label{tab:steady-state-extra}
  \begin{subtable}[t]{0.49\linewidth}
    \centering
    \caption{Compressible airfoils and PCS~\cite{hao2024gaot}}
    \label{tab:gaot-suite}
    \small
    \setlength{\tabcolsep}{3pt}
    \begin{tabular}{@{}l cc cc@{}}
      \toprule
      & \multicolumn{2}{c}{{Ours}} & \multicolumn{2}{c}{{Baselines}} \\
      \midrule
      Dataset & HTNO & \name{} & RIGNO$^\dagger$ & GAOT$^\dagger$ \\
      \midrule
      NACA0012 & \underline{4.53} & \textbf{3.77} & 5.09 & 12.13         \\
      NACA2412 & \underline{4.73} & \textbf{3.92} & 5.73 & 12.25         \\
      RAE2822  & \underline{4.55} & \textbf{3.92} & 4.81 & 12.98         \\
      PCS      & \textbf{0.52}    & 3.09          & 7.11 & \underline{0.72} \\
      \bottomrule
    \end{tabular}
  \end{subtable}
  \hfill
  \begin{subtable}[t]{0.49\linewidth}
    \centering
    \caption{EmmiWing~\cite{paischer2025going}.}
    \label{tab:emmiwing}
    \small
    \setlength{\tabcolsep}{3pt}
    \begin{tabular}{@{}l ccccc@{}}
      \toprule
      Model & $p_s$ & $\tau_x$ & $\tau_y$ & $\tau_z$ & Agg. \\
      \midrule
      PointNet$^\dagger$    & 0.29             & 4.24             & 5.75             & 5.90             & 2.62 \\
      Transformer$^\dagger$ & 0.30             & 4.21             & 5.57             & 6.13             & 2.60 \\
      RIGNO-18$^\dagger$    & 0.29             & 4.46             & 6.02             & 6.30             & 2.78 \\
      Transolver$^\dagger$  & \underline{0.28} & \underline{4.14} & \underline{5.52} & \underline{5.57} & \underline{2.55} \\
      HTNO              & \textbf{0.24}    & \textbf{3.94}    & \textbf{5.32}    & \textbf{5.37}    & \textbf{2.41} \\
      \bottomrule
    \end{tabular}
  \end{subtable}
  \vspace{-5.0mm}
\end{table}

\paragraph{Compressible airfoils and Poisson-with-sines (\cref{tab:gaot-suite})}
We additionally evaluate on the four single-file steady-state datasets~\cite{hao2024gaot}: three compressible-airfoil sets (NACA0012/2412, RAE2822) on $8$K-node meshes, spanning subsonic to supersonic regimes, and Poisson-with-sines on a fixed $16$K-node mesh~(PCS).
On all three airfoils, both \name{} and HTNO clearly outperform RIGNO and GAOT.
This can be attributed to the elliptic characteristics that dominate steady compressible flow at subsonic and transonic conditions (smooth pressure/velocity outside shocks), which align naturally with the Hodge / harmonic-basis inductive biases that characterize \name{}s.
PCS is the converse setting: a fixed mesh with a smooth, multiscale-sinusoidal Poisson source, for which GAOT's structured-latent encoder is well matched, beating RIGNO by an order of magnitude. 
Nevertheless, the HTNO, which combines local and global interactions, yields strong performance (\textbf{0.52} vs.\ \underline{0.72}).

\subsection{Large-Scale Surface PDEs: EmmiWing}
\label{sec:emmiwing}
EmmiWing~\cite{paischer2025going} comprises 3D wing surfaces (raw meshes $244$K--$426$K nodes; variable per sample) with four output channels (surface pressure $p_s$ plus three wall-shear-stress components $\tau_{x,y,z}$; official $25{,}674/999/2{,}992$ split). Following the AB-UPT training regime~\cite{alkin2025abupt}, we train on $65$K shared FPS query points per sample with $16{,}384$-node random subsampling per step, and evaluate uniformly on the $65$K FPS subset rather than the full raw mesh; see~\cref{app:emmiwing-details} for the full protocol.
HTNO leads aggregate $L^1$ at $2.41\%$, beating Transolver~\cite{wu2024transolver} ($2.55\%$), our matched-budget self-attention Transformer ($2.60\%$) and PointNet ($2.62\%$), and RIGNO-18 ($2.78\%$); HTNO additionally converges in ${\sim}6{\times}$ fewer epochs than RIGNO-18.
\name{} is subsumed by the hierarchical HTNO, which models long-range interactions without per-sample Hodge decompositions on large resampled geometries.

\subsection{Anisotropic Darcy with Per-Face Random Tensor Orientation}
\label{sec:rankin-aniso}

\paragraph{Setup}
\label{par:synth-mesh-family}
\name{}s naturally accommodate higher-rank signals. To validate this topology-aware inductive bias, we simulate the following anisotropic Darcy PDE on $100$ triangulated planar square domains $[-1,1]^2$ with one or two randomly placed circular holes:
\begin{equation}
  -\nabla\!\cdot\!\big(\kappa(x)\,\nabla u\big) = f, \qquad u|_{\partial\Omega_{\mathrm{outer}}}=0,\quad u|_{\partial\Omega_{\mathrm{hole},k}}=g_k\!\sim\!\mathcal{U}[0.5,1.5],
\end{equation}
The diffusivity $\kappa(x)$ is a piecewise-constant face-valued symmetric tensor with eigenvalues $(\kappa_\parallel,\kappa_\perp)=(4,1)$ and a per-face principal axis $\phi_t \!\sim\! \mathcal{U}[0,\pi)$ \emph{drawn iid per triangle and fixed across all samples on a given mesh}, so that the rank-$2$ orientation field $\cos(2\phi_t)$ has unit-order variance on every face, but \textit{vanishing population mean}.
\textbf{Projection onto vertices is therefore lossy by construction:} averaging $\cos(2\phi_t)$ over incident faces concentrates to $\approx\!0$ at every vertex regardless of the underlying $\kappa$.
This isolates the effect of \emph{how} the orientation is utilised.
An architecture that ingests it at its native rank $2$ (as a face-supported $2$-cochain) retains this orientation signal that architectures isolated to vertices discard.
We compare \name{} (rank-$2$, harmonic basis, sheaf transport) against an MPNN baseline at matched and parameter-matched width ($w$), each consuming the orientation field in three ways: not at all (vertex inputs only), as a vertex-projected channel, or natively at rank $2$.
PDE simulation, training recipe, mesh construction, and face-supervision cells are in~\cref{app:rankin-aniso-details}.

\vspace{-2.0mm}
\begin{figure}[t]
  \centering
  \begin{minipage}[t]{0.49\linewidth}
    \centering
    \includegraphics[width=\linewidth]{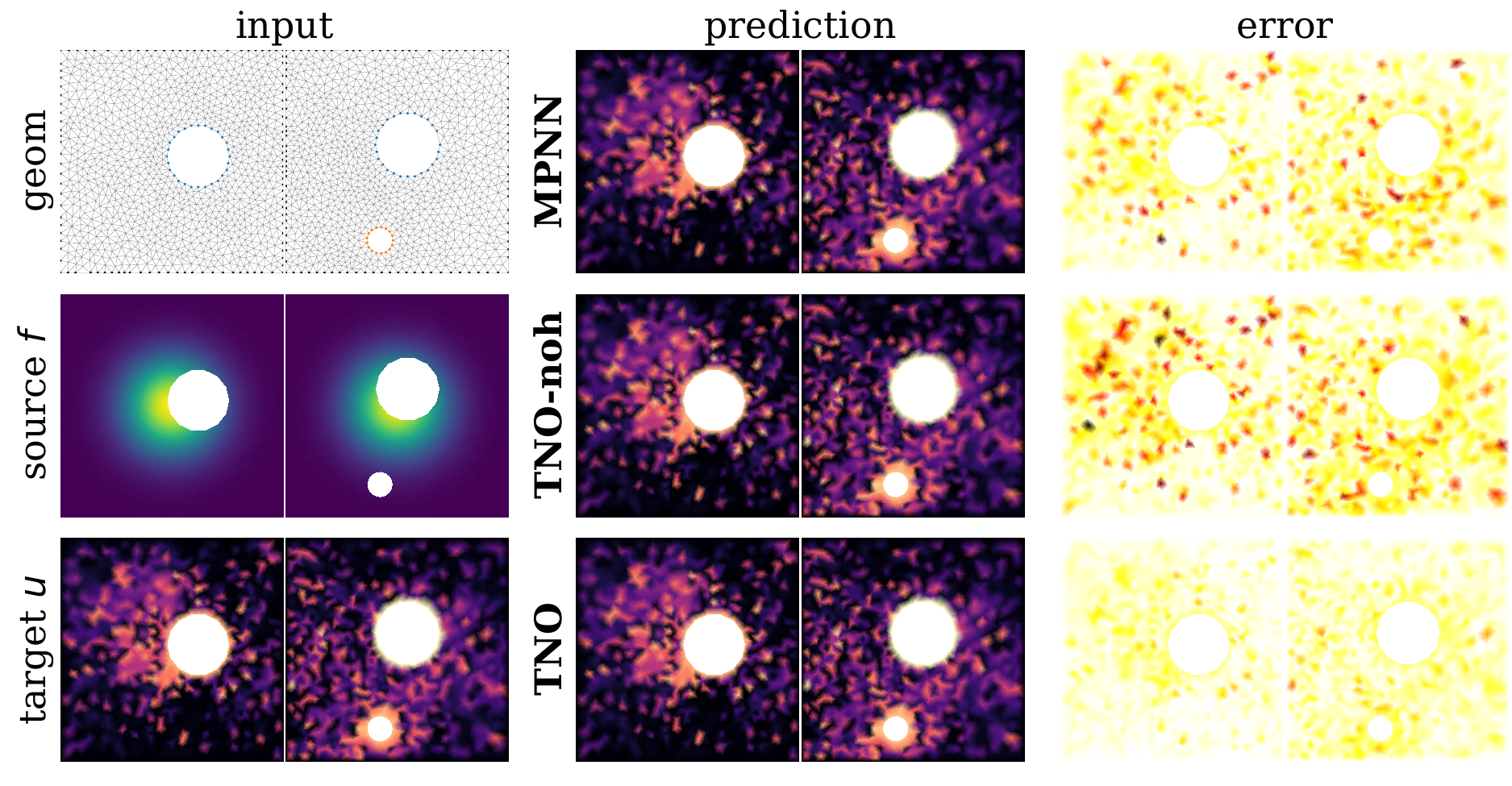}
  \end{minipage}
  \hfill
  \begin{minipage}[t]{0.49\linewidth}
    \centering
    \includegraphics[width=\linewidth]{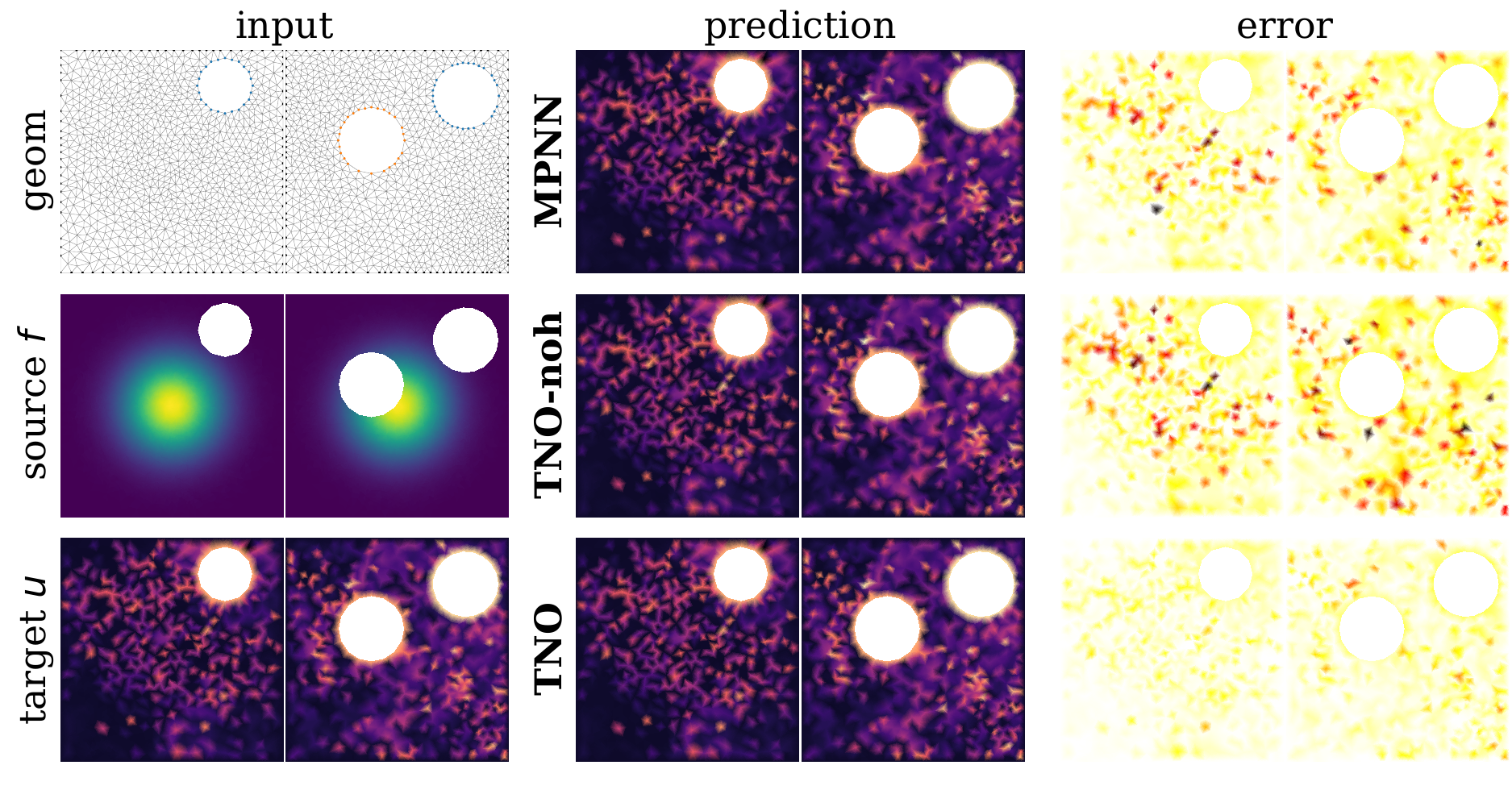}
  \end{minipage}
  \vspace{-2.0mm}
  \caption{Qualitative ablation across three models: Vanilla MPNN vs.\ TNO (no harmonics) vs.\ TNO with harmonics / copresheaves on Anisotropic Darcy (left) and Advection Diffusion (right), for which we depict model inputs and targets (col. 1), model predictions (col. 2), and model errors (col. 3).}
  \label{fig:ablation-qualitative}
  \vspace{-4.5mm}
\end{figure}

\begin{table}[t]
  \centering
  \caption{Rank-input and component studies: test relative error~(\%); lower is better, \textbf{bold}: best, \underline{underline}: second; $^\ast$param-matched MPNN ($w{=}272$, ${\sim}5.35$M\,$\approx$\,\name{} $5.0$M).}
  \label{tab:controlled-studies}
  \begin{subtable}[t]{0.49\linewidth}
    \centering
    \caption{Anisotropic Darcy with per-face random tensor orientation~(\cref{sec:rankin-aniso}); $L^1$\,/\,$L^2$.}
    \label{tab:rankin-aniso}
    \small
    \setlength{\tabcolsep}{4pt}
    \begin{tabular}{@{}l cc@{}}
      \toprule
      Variant & $L^1$ & $L^2$ \\
      \midrule
      MPNN ($w{=}192$), vertex            & 9.97             & 7.03 \\
      MPNN ($w{=}192$), projected         & 9.10             & 6.52 \\
      MPNN ($w{=}272$)$^\ast$, projected  & 7.59             & 5.28 \\
      \midrule
      \name{}, vertex                     & \underline{5.39} & 3.85 \\
      \name{}, projected                  & 5.42             & \underline{3.66} \\
      \name{}, native rank-$2$            & \textbf{4.88}    & \textbf{3.32} \\
      \bottomrule
    \end{tabular}
  \end{subtable}\hfill
  \begin{subtable}[t]{0.49\linewidth}
    \centering
    \caption{\name{} component ablation on synthetic topologies~(\cref{sec:ablations}); $L^1$ per PDE.}
    \label{tab:ablation-v19}
    \small
    \setlength{\tabcolsep}{4pt}
    \begin{tabular}{@{}l cc@{}}
      \toprule
      Variant & Darcy & Adv.-Diff.\ \\
      \midrule
      MPNN ($w{=}192$)                & 8.87             & 9.78  \\
      MPNN ($w{=}272$)$^\ast$         & 7.29             & 7.66  \\
      \midrule
      \name{}                         & 9.44             & 10.68 \\
      \name{} $+\,$harm.\ basis       & \underline{5.16} & \underline{5.77} \\
      \name{} $+\,$copresheaf              & 11.64            & 13.05 \\
      \name{} $+\,$copresheaf$\,+\,$harm.\ & \textbf{4.87}    & \textbf{5.20} \\
      \bottomrule
    \end{tabular}
  \end{subtable}
  \vspace{-5.0mm}
\end{table}

\paragraph{Results}
Two distinct effects emerge (see \cref{tab:rankin-aniso}).
\textbf{(i) Architecture, ${\sim}4$--$5$~pp, independent of how orientation is consumed.} \name{} outperforms same-width MPNN by $4.58$~pp on vertex inputs and by $3.68$~pp on projected inputs; against the parameter-matched projected MPNN, the margin remains $2.17$~pp.
The harmonic-basis channel resolves the spectral signature of the anisotropic operator, which vanilla message-passing cannot.
\textbf{(ii) Native rank-$2$ ingestion, an additional ${\approx}0.5$~pp.}
On the domain where vertex projection is lossy by design, \name{} with native rank-$2$ orientation beats \name{} with projected orientation by $0.54$~pp ($\sim\!10\%$ relative), while the projected and vertex-only variants are within $0.03$~pp of each other, confirming that the projected channel carries no operator information, as demanded by the iid-$\phi$ construction.
The two effects are additive: the architecture earns ${\sim}5$~pp; native rank-$2$ ingestion yields another ${\sim}0.5$~pp improvement.

\vspace{-2.mm}
\subsection{Ablations}

\label{sec:ablations}

\paragraph{Setup} We isolate the contribution of \name{}'s harmonic-basis input and sheaf transport on two scalar PDEs whose flux carries non-trivial curl on the same mesh family as~\cref{sec:rankin-aniso}: Darcy with reaction ($-\nabla\!\cdot\!(\kappa\nabla u) + \sigma u = f$) and conservative advection--diffusion.
Per-sample variability is restricted to the per-hole Dirichlet BC values, so the topology is the only axis that varies across samples on a fixed mesh. 
All variants share a common backbone (hidden dim $192$; run $300$ epochs); MPNN baselines at matched width and parameter count bracket the comparison.

\paragraph{Results}
Disabling the harmonic basis yields a loss of $6.8$~pp on Darcy and $7.9$~pp on Adv.-Diff (see \cref{tab:ablation-v19}). 
Sheaf transport and the harmonic basis are synergistic, not additive: with harmonic on, sheaf helps; with harmonic off, sheaf actively hurts (Darcy $9.44 \to 11.64$). 
The complete \name{} wins on both PDEs, beating the param-matched MPNN by ${\sim}33\%$ (see \cref{fig:ablation-qualitative}).
This implies harmonic and sheaf components are an effective inductive bias, particularly for PDEs with non-trivial curl.

\vspace{-1mm}
\section{Conclusion}
\vspace{-1mm}
We introduced \textbf{TNOs}, a topological framework for operator
learning on cell complexes. Fields are represented as cochains on
vertices, edges, faces, and volumes, and their interactions are routed
by DEC operators. Thus topology determines where information flows,
while learning determines how transported features are transformed.
This yields architectures suited to multi-physics coupling, irregular
geometries, and discretization transfer, with existing neural operators arising as special cases. 
Empirically, TNOs and HTNOs perform strongly across irregular geometries and higher-rank PDE systems, improving physical consistency. 
This points toward scientific foundation models that can learn operators on structured spaces, unifying geometry, physics, and computation.

\paragraph{Limitations and future work}
Our work leaves ample room for future work: scaling to dynamic, adaptive, and very large domains; developing approximation, stability,
and convergence theory, clarifying the role of harmonic and
sheaf-based channels; and extending TNOs to multiscale systems, evolving manifolds, inverse problems, and gauge- or symmetry-equivariant operator families. 

\paragraph{Acknowledgements}
The authors are grateful for support from the UK AI Research Resource (AIRR) through grant 0251-4584-0945-1.
T. B. acknowledges support from the UKRI Engineering and Physical Sciences Research Council (EPSRC) through the Future Leaders Fellowship [grant number MR/Y018818/1]. 
L.B. was supported by the UK Royal Society through grant NIF/R1/254128.

\paragraph{Broader impact}
Topological Neural Operators (TNOs) aim to bring geometric and topological structure into operator learning by modeling physical quantities on their natural supports, such as vertices, edges, faces, and volumes. We hope this perspective enables more faithful and efficient surrogate models for scientific computing, with applications in physics, engineering, and simulation-driven design.
At the same time, TNOs remain learned approximations and do not guarantee certified physical correctness or numerical stability outside the training distribution. Careful validation against trusted numerical solvers is therefore essential, particularly in safety-critical applications. More broadly, this work explores the initial steps of an exciting idea: that data-driven simulation systems may benefit from respecting the shape and structure of the worlds they model.

\vspace{5mm}

\bibliographystyle{plainnat}
\bibliography{refs}

\clearpage

\appendix
\section*{\centering \Large Topological Neural Operators\\[0.3em]\large\normalfont Appendix}
\vspace{-1mm}
\startcontents[appendices]
\setcounter{tocdepth}{1}
\printcontents[appendices]{0}{1}{\section*{Table of Contents}}

\section{Background}
\label{app:background}

\subsection{Combinatorial Complexes}
\begin{figure}[h]
\centering
\includegraphics[width=0.72\textwidth]{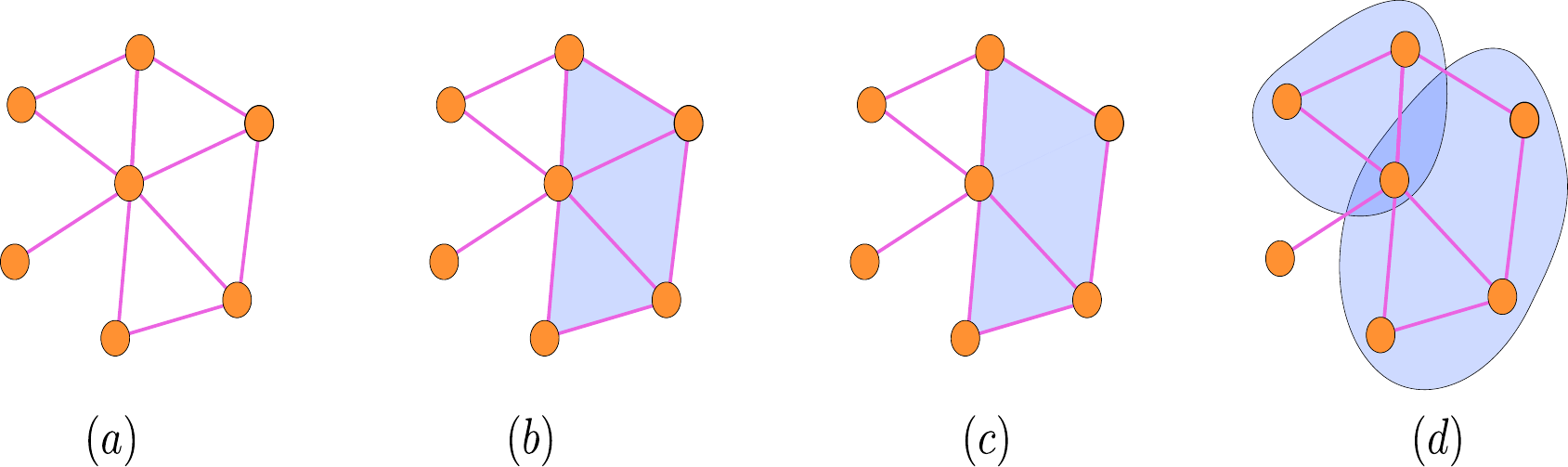}
\vspace{-0.5em}
\caption{
Progression of topological domains with increasing modeling flexibility.
\textbf{(a)} A \textit{graph} represents only vertices and pairwise edges.
\textbf{(b)} A \textit{simplicial complex} adds higher-order simplices, but each
simplex is determined by its lower-dimensional faces.
\textbf{(c)} A \textit{cell complex} allows more general cells, such as polygonal
regions, whose boundaries need not be simplices.
\textbf{(d)} A \textit{combinatorial complex} further relaxes the construction by
allowing independently specified, possibly overlapping higher-order
relations. Each step increases the flexibility of the domain on which
cochains and topological neural operators can be defined.
}
\label{fig:topological-domain-hierarchy}
\vspace{-0.8em}
\end{figure}

Combinatorial complexes generalize simplicial and cell complexes as well as hypergraphs, acting as a unifying topological structure:
\begin{definition}[Combinatorial complex~\citep{hajij2023topological,hajij2023combinatorial}]
\label{def:cc}
A \emph{combinatorial complex} (CC) is a triple
$(S, \mathcal{X}, \mathrm{rk})$ consisting of a set $S$ of entities,
a subset
$\mathcal{X} \subseteq \mathcal{P}(S) \setminus \{\emptyset\}$ of
cells, and a rank function
$\mathrm{rk}: \mathcal{X} \to \mathbb{Z}_{\geq 0}$ satisfying
$\mathrm{rk}(\{v\}) = 0$ for all $v \in S$ and
$\mathrm{rk}(x) \leq \mathrm{rk}(y)$ whenever $x \subseteq y$.
\end{definition}
\vspace{-1.5mm}
\noindent\textbf{Lifting} an RCC $K$ to a CC $\widetilde{K}$ retains the cochains of $K$ but may carry additional cells, e.g., higher-order groupings or augmented neighborhoods that enrich the information flow. 
The lift does not change the target operator $\mathcal{G}$ or the input/output spaces $\mathcal{X}(K), \mathcal{Y}(K)$; it only enlarges the domain on which the layer acts.  
See \cref{fig:topological-domain-hierarchy}.

\subsection{Discrete Differential Forms and Physical Cochains}
\label{app:derham-cochains}

We recall the discrete differential-form viewpoint used throughout the
paper~\citep{hirani2003discrete,desbrun2005discrete,desbrun2006discrete,crane2013digital,crane2018discrete}, together with the finite-element-exterior-calculus perspective of \citet{arnold2006finite}. 
Our goal is not to introduce differential forms for their own
sake, but to explain why physical fields should not all be represented
as vertex signals. 
Many physical quantities are not point samples. 
A potential is naturally measured at points, a circulation along curves, a flux through surfaces, and a density over volumes. 
Discrete exterior calculus (DEC) keeps this measurement type in the discrete model.

\begin{figure}[ht]
\centering
$\vcenter{\hbox{\includegraphics[scale = 0.1, keepaspectratio = 0.20]{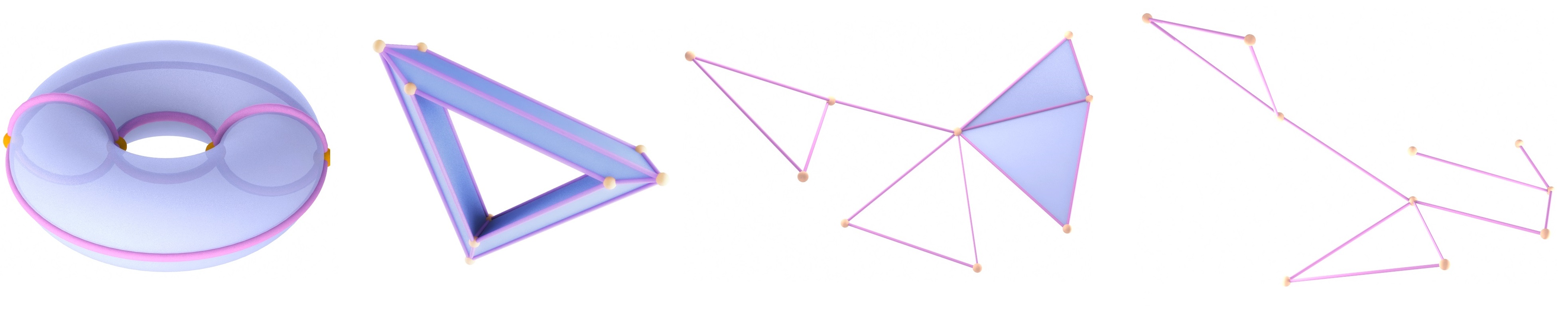}}}$
\caption{Four examples of regular cell complexes.}
\label{fig:CCC}
\end{figure}

Throughout, \(K\) denotes an oriented regular cell complex. 
We write \(K_k\) for the set of \(k\)-cells: vertices for \(k=0\), edges for \(k=1\), faces for \(k=2\), and volumes for \(k=3\). 
Examples of regular cell complexes are shown in \cref{fig:CCC}.

A \emph{discrete \(k\)-form}, or \(k\)-cochain, is a field supported on
oriented \(k\)-cells:
\[
C^k(K;\mathbb{R}^d)
=
\{u^k:K_k\to \mathbb{R}^d\}
\cong
\mathbb{R}^{|K_k|\times d}.
\]
Thus \(0\)-cochains live on vertices, \(1\)-cochains on edges,
\(2\)-cochains on faces, and \(3\)-cochains on volumes. The degree \(k\)
is therefore not merely an index. It records the geometric type of the
quantity.

The useful intuition is that cochains are \emph{cellular measurements}.
A \(0\)-cochain stores point measurements, A \(1\)-cochain oriented line measurements, \(2\)-cochains oriented surface measurements, \(3\)-cochain volume measurements, and so forth. 
This is the discrete analog of the integral viewpoint in exterior calculus: one does not approximate every physical field by point values; one stores each quantity on the cells over which they are naturally measured.

This distinction is important. 
A flux through a face is not the same kind of object as a scalar value at a vertex. 
A circulation along an edge cannot be represented faithfully as a node feature. 
If such fields are collapsed to vertices, their orientation, incidence, and conservation relations become hidden or must be relearned from data (see \cref{sec:rankin-aniso}). 
DEC avoids this collapse by making the support dimension part of the representation.

\paragraph{Chains, cochains, and discrete integration}
Let \(K\) be an oriented cell complex and let \(K_k\) denote its set of oriented \(k\)-cells~\citep{hirani2003discrete}. 
A \(k\)-chain is a finite formal linear combination of \(k\)-cells,
\[
    c=\sum_{\sigma\in K_k} c_\sigma\,\sigma,
    \qquad c_\sigma\in\mathbb{R}.
\]
The coefficients \(c_\sigma\) record how the cells are assembled. 
A positive coefficient uses the chosen orientation of \(\sigma\), while a negative coefficient uses the opposite orientation. 
Thus, a chain is a discrete geometric domain: a \(0\)-chain is a signed collection of vertices, a \(1\)-chain is an oriented path made of edges, a \(2\)-chain is an oriented surface patch made of faces, and so on.

In this paragraph, we take cochains to be scalar-valued, since this is
the setting in which the integration pairing and Stokes' theorem are
most directly stated. 
A \(k\)-cochain is a linear functional on
\(k\)-chains. 
Equivalently, it is specified by assigning one scalar to
each oriented \(k\)-cell,
\[
    u^k:K_k\to\mathbb{R}.
\]
Its value on a chain \(c=\sum_{\sigma\in K_k}c_\sigma\sigma\) is defined by linearity:
\[
    \langle u^k,c\rangle
    =
    \sum_{\sigma\in K_k} c_\sigma\,u^k(\sigma).
\]
This pairing is the discrete analog of integration. 
If \(\omega^k\)
is a smooth \(k\)-form and \(R\) is a \(k\)-dimensional oriented region, then the continuum measurement is
\[
    \int_R \omega^k .
\]
In the discrete setting, the region \(R\) is replaced by a \(k\)-chain
\(c\), the form \(\omega^k\) is replaced by a \(k\)-cochain \(u^k\), and
the integral is replaced by
\[
    \int_R \omega^k
    \quad\leadsto\quad
    \langle u^k,c\rangle .
\]

\begin{figure}[t]
\centering
\resizebox{\linewidth}{!}{%
\begin{tikzpicture}[
    every node/.style={font=\small},
    cell/.style={
        draw,
        rounded corners=2pt,
        fill=blue!7,
        minimum width=1.15cm,
        minimum height=0.52cm
    },
    boxA/.style={
        draw,
        rounded corners=3pt,
        fill=orange!12,
        minimum width=3.4cm,
        minimum height=0.85cm,
        align=center
    },
    boxB/.style={
        draw,
        rounded corners=3pt,
        fill=green!10,
        minimum width=3.4cm,
        minimum height=0.85cm,
        align=center
    },
    boxC/.style={
        draw,
        rounded corners=3pt,
        fill=gray!10,
        minimum width=4.3cm,
        minimum height=0.85cm,
        align=center
    },
    arrow/.style={->, thick}
]

\node[cell] (s1) at (0,1.0) {\(\sigma_1\)};
\node[cell] (s2) at (0,0.0) {\(\sigma_2\)};
\node[cell] (s3) at (0,-1.0) {\(\sigma_3\)};
\node[font=\small\bfseries] at (0,1.65) {oriented \(k\)-cells};

\node[boxA] (chain) at (4.1,0) {
\(k\)-chain\\[0.15em]
\(c=c_1\sigma_1+c_2\sigma_2+c_3\sigma_3\)
};

\node[boxB] (cochain) at (8.6,0) {
\(k\)-cochain\\[0.15em]
\(u^k(\sigma_i)\in\mathbb{R}\)
};

\node[boxC] (pairing) at (13.9,0) {
discrete integral\\[0.15em]
\(\langle u^k,c\rangle=\sum_i c_i\,u^k(\sigma_i)\)
};

\draw[arrow] (s1.east) -- (chain.west);
\draw[arrow] (s2.east) -- (chain.west);
\draw[arrow] (s3.east) -- (chain.west);

\draw[arrow] (chain.east) -- node[above] {domain} (cochain.west);
\draw[arrow] (cochain.east) -- node[above] {pairing} (pairing.west);

\end{tikzpicture}
}
\caption{
Chains and cochains as discrete integration. A \(k\)-chain is a signed
combination of oriented \(k\)-cells and represents a discrete
\(k\)-dimensional domain. A \(k\)-cochain assigns measurements to
oriented \(k\)-cells. Their pairing sums the cochain values over the
chain with the chain coefficients, giving the discrete analogue of
integrating a \(k\)-form over a \(k\)-dimensional region.
}
\label{fig:chain-cochain-pairing}
\end{figure}
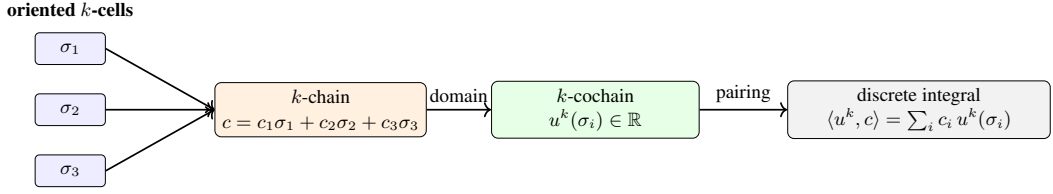

This is the sense in which cochains are discrete integrands. A
\(1\)-cochain can be summed along an oriented path. A \(2\)-cochain can
be summed over an oriented surface patch. A \(3\)-cochain can be summed
over a collection of volume cells. The degree \(k\) specifies both where
the quantity is stored and what kind of discrete region it can be
measured on. See Figure \ref{fig:chain-cochain-pairing}.

For neural-network features we often use vector-valued cochains
\(u^k:K_k\to\mathbb{R}^{d_k}\). These should be understood as
\(d_k\) scalar cochains stored in parallel, with the same pairing applied
componentwise. The scalar case is the one used to state the discrete
Stokes relation; the vector-valued case is its channel-wise extension.

This interpretation prepares the definition of the exterior derivative.
Once cochains are viewed as discrete integrands, Stokes' theorem becomes
a statement about how the measurement of a derivative on a cell is
related to the measurement of the original cochain on the oriented
boundary of that cell.

\paragraph{From cellular measurements to the discrete exterior derivative}
A cochain stores measurements on cells. The degree records the type of
measurement: values on vertices, line measurements on edges, surface
measurements on faces, and volume measurements on \(3\)-cells. This is
the de Rham convention used throughout the paper: a physical quantity is
placed on the cell type over which it is naturally measured.

\begin{table}[h]
\centering
\caption{
Physical quantities and their natural cochain degrees under the primal
de Rham convention~\citep{desbrun2006discrete,arnold2006finite}. A \(k\)-form is discretized as a \(k\)-cochain by
integration over oriented \(k\)-cells.
}
\label{tab:cochains}
\begin{tabular}{llll}
\toprule
Physical quantity & Geometric type & Degree \(k\) & Primal cell \\
\midrule
Temperature, pressure, potential
& \(0\)-form & \(0\) & Vertices \\
Electric field \(E\), circulation, line flow, Darcy flux
& \(1\)-form & \(1\) & Edges \\
Magnetic flux \(B\), vorticity flux
& \(2\)-form & \(2\) & Faces \\
Volumetric charge, mass, source density
& \(3\)-form & \(3\) & Volumes \\
\bottomrule
\end{tabular}
\end{table}

Table~\ref{tab:cochains} should be read as a statement about
measurement, not only about notation. 
A scalar potential is sampled at points. 
A circulation or line integral is measured along an oriented curve, and in the discrete complex, this means an edge. 
A flux is measured through an oriented surface, and in the discrete complex, this means a face. 
A density or source term is measured over a volume. This is why the cochain degree is part of the field's physical meaning.

The table uses the primal convention. In an \(n\)-dimensional domain, the Hodge star identifies primal \(k\)-cochains with dual \((n-k)\)-cochains. 
Consequently, the same physical vector quantity may be stored on different cells depending on whether one uses the primal or dual complex. Unless stated otherwise, this paper uses the primal convention in Table~\ref{tab:cochains}.

The question is then how to differentiate such quantities without forgetting where they are measured. The classical answer is that differentiation and boundary measurement are tied together. 
In one dimension, the fundamental theorem of calculus says that the integral of a derivative over an interval is determined by the values at the two boundary points:
\[
    \int_a^b f'(t)\,dt = f(b)-f(a).
\]
If the interval is oriented from \(a\) to \(b\), its boundary is the
formal signed sum
\[
    \partial[a,b]=[b]-[a].
\]
Thus the derivative measured on the interval is obtained from the original function measured on the oriented boundary. 
The endpoint at the head of the interval contributes with sign \(+1\), and the endpoint at the tail contributes with sign \(-1\).

The same boundary principle appears in higher dimensions. For a vector
field in the plane, Green's theorem relates circulation around the
boundary of a region to a curl-type quantity inside the region. In three
dimensions, Stokes' theorem relates circulation around the boundary
curve of a surface to curl through the surface. Gauss' theorem relates
flux through the boundary surface of a volume to divergence inside the
volume. Exterior calculus writes these boundary laws as one statement:
\[
    \int_R d\omega = \int_{\partial R} \omega .
\]
Here \(\omega\) is measured on \(k\)-dimensional objects, \(d\omega\) is
measured on \((k+1)\)-dimensional objects, \(R\) is a
\((k+1)\)-dimensional region, and \(\partial R\) is its oriented boundary. 
The identity states that the derivative of a quantity over a region is determined by the original quantity on the boundary of that region.

This is the viewpoint DEC keeps. 
Since a cochain already stores integrated measurements on cells, the discrete derivative is defined by the same boundary relation. 
A vertex quantity differentiates to an edge quantity by taking signed endpoint differences. 
An edge quantity differentiates to a face quantity by taking signed circulation around the face boundary. 
A face quantity differentiates to a volume quantity by taking signed flux through the volume boundary. 
The common operation is evaluation on an oriented boundary.

On a cell complex, the role of the region \(R\) is played by a
\((k+1)\)-cell \(\tau\), and the role of \(\partial R\) is played by the
oriented cellular boundary of \(\tau\). Let
\[
B_{k+1}: C_{k+1}(K)\to C_k(K)
\]
be the oriented boundary matrix. Its columns are indexed by
\((k+1)\)-cells and its rows are indexed by \(k\)-cells. For
\(\tau\in K_{k+1}\), the column indexed by \(\tau\) records
\[
\partial \tau
=
\sum_{\sigma\in K_k}
[\tau:\sigma]\,\sigma,
\qquad
[\tau:\sigma]\in\{-1,0,+1\}.
\]
The coefficient \([\tau:\sigma]\) is nonzero exactly when \(\sigma\) is
a boundary face of \(\tau\). The sign records orientation: \(+1\) if the
chosen orientation of \(\sigma\) agrees with the orientation induced from
\(\tau\), and \(-1\) if it disagrees.

Now let \(u^k\in C^k(K)\) be a \(k\)-cochain. Its discrete exterior
derivative \(d^k u^k\) is the \((k+1)\)-cochain obtained by measuring
\(u^k\) on the oriented boundary of each \((k+1)\)-cell:
\[
\langle d^k u^k,\tau\rangle
=
\langle u^k,\partial\tau\rangle,
\qquad
\tau\in K_{k+1}.
\]
In matrix form,
\[
d^k:C^k(K)\to C^{k+1}(K),
\qquad
d^k=B_{k+1}^{\top}.
\]
Equivalently,
\[
(d^k u^k)(\tau)
=
\sum_{\sigma\prec\tau}
[\tau:\sigma]\,u^k(\sigma).
\]
Thus \(d^k\) takes measurements on \(k\)-cells and produces measurements
on \((k+1)\)-cells by signed summation over oriented boundaries. Its
nonzero pattern comes from incidence, and its signs come from
orientation. This is why the discrete exterior derivative is a
topological operator~\citep{hirani2003discrete,desbrun2005discrete}.

For TNOs, this is the basic upward cross-rank route. Vertex cochains are
coupled to edge cochains, edge cochains to face cochains, and face
cochains to volume cochains through the same boundary rule that appears
in the underlying differential equations. The cell complex fixes the
route; the learnable part is the channel mixing applied to the routed
cochain features.

In low degrees, we obtain the sequence
\[
C^0(K)
\xrightarrow{\;d^0\;}
C^1(K)
\xrightarrow{\;d^1\;}
C^2(K)
\xrightarrow{\;d^2\;}
C^3(K).
\]
These maps are the cellular analogues of the familiar vector-calculus operators:
\[
d^0=\text{gradient-type map},
\qquad
d^1=\text{curl-type map},
\qquad
d^2=\text{divergence-type map}.
\]
Here ``type'' means that the maps act between cellular measurement
spaces. The map \(d^0\) takes vertex measurements to edge measurements.
The map \(d^1\) takes edge measurements to face measurements. The map
\(d^2\) takes face measurements to volume measurements.

For \(x\in C^0(K)\) and an oriented edge \(e=[i,j]\),
\[
(d^0x)([i,j])=x_j-x_i.
\]
This is the fundamental theorem of calculus on a single edge: the
measurement of the derivative along the edge is the signed difference of
the endpoint values. Hence \(d^0x\) is an edge cochain.

For \(y\in C^1(K)\) and an oriented face \(f\in K_2\),
\[
(d^1y)(f)
=
\sum_{e\prec f}
[f:e]\,y(e).
\]
The value on the face is the signed circulation of the edge measurements
around its boundary. This is the cellular version of Green's/Stokes'
theorem. For an oriented triangle \([i,j,k]\), the induced boundary is
\[
\partial[i,j,k]=[j,k]-[i,k]+[i,j].
\]
With stored edge orientations \([i,j]\), \([i,k]\), and \([j,k]\), this
gives
\[
(d^1y)([i,j,k])
=
y_{ij}-y_{ik}+y_{jk}.
\]
If the edge is stored as \([k,i]\) rather than \([i,k]\), the same
circulation is written as
\[
(d^1y)([i,j,k])
=
y_{ij}+y_{jk}+y_{ki}.
\]
The two expressions represent the same oriented boundary measurement.
The signs make the result independent of the bookkeeping choice for edge
orientations.

For \(z\in C^2(K)\) and an oriented volume cell \(c\in K_3\),
\[
(d^2z)(c)
=
\sum_{f\prec c}
[c:f]\,z(f).
\]
The value on the volume is the signed accumulation of the face measurements on its boundary. This is the cellular version of Gauss' theorem: it measures net boundary flux.

\begin{figure}[h]
\centering
\scriptsize
\setlength{\arraycolsep}{3pt}
\renewcommand{\arraystretch}{0.9}
\resizebox{0.96\linewidth}{!}{%
\begin{tabular}{@{}c@{\qquad}c@{}}
\includegraphics[width=0.18\linewidth]{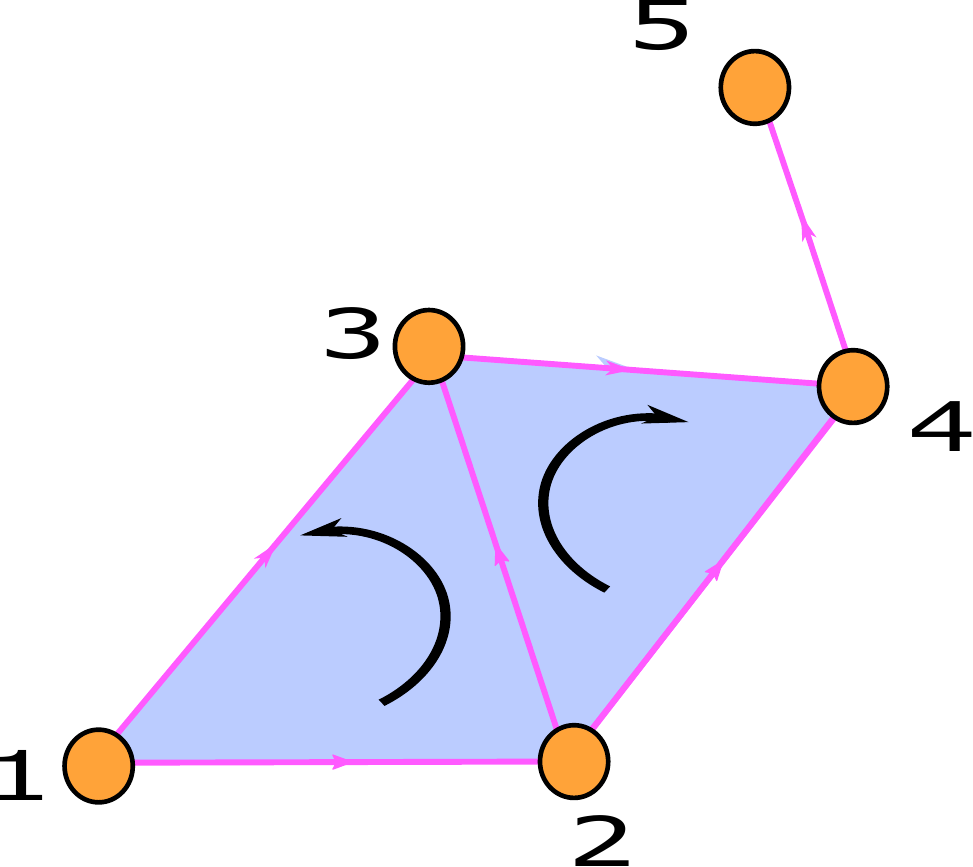}
&
\(
\begin{gathered}
d^0=B_1^{\top}:C^0(K)\to C^1(K),
\qquad
d^1=B_2^{\top}:C^1(K)\to C^2(K)
\\[0.5em]
d^0 =
\begin{bmatrix}
-1 &  1 &  0 &  0 &  0 \\
-1 &  0 &  1 &  0 &  0 \\
 0 & -1 &  1 &  0 &  0 \\
 0 & -1 &  0 &  1 &  0 \\
 0 &  0 & -1 &  1 &  0 \\
 0 &  0 &  0 & -1 &  1
\end{bmatrix},
\qquad
d^1 =
\begin{bmatrix}
 1 & -1 &  1 &  0 &  0 &  0 \\
 0 &  0 &  1 & -1 &  1 &  0
\end{bmatrix}.
\end{gathered}
\)
\end{tabular}
}
\caption{
Coboundary matrices for the oriented complex. The matrix \(d^0\) maps
vertex cochains to edge cochains: each row computes one oriented
difference, e.g. \([-1,1,0,0,0]\) gives \(x_2-x_1\) on \([1,2]\). 
The matrix \(d^1\) maps edge cochains to face cochains: each row computes one signed circulation. For the face \([1,2,3]\), the boundary relation \(\partial[1,2,3]=[2,3]-[1,3]+[1,2]\) gives \((d^1y)([1,2,3])=y_{12}-y_{13}+y_{23}\); similarly, \((d^1y)([2,3,4])=y_{23}-y_{24}+y_{34}\). 
Hence, these matrices encode oriented incidence, not just adjacency.
}
\label{fig:coboundary_matrices}
\end{figure}

The following example makes the construction explicit~\citep{bell2012pydec}. Consider the
oriented complex in \cref{fig:coboundary_matrices} with
\[
K_0=([1],[2],[3],[4],[5]),
\]
\[
K_1=([1,2],[1,3],[2,3],[2,4],[3,4],[4,5]),
\qquad
K_2=([1,2,3],[2,3,4]).
\]
The matrix \(d^0\) has one row per edge and one column per vertex. The
row indexed by \([i,j]\) has a \(-1\) in column \(i\), a \(+1\) in
column \(j\), and zeros elsewhere. Multiplying \(d^0\) by a vertex
cochain therefore computes all oriented edge differences at once.

The matrix \(d^1\) has one row per face and one column per edge. Each
row is the oriented boundary of one face written in the chosen edge
basis. For the first face,
\[
\partial[1,2,3]=[2,3]-[1,3]+[1,2],
\]
so the first row of \(d^1\), using the edge ordering above, is
\[
[\,1,\,-1,\,1,\,0,\,0,\,0\,].
\]
For the second face,
\[
\partial[2,3,4]=[3,4]-[2,4]+[2,3],
\]
so the second row is
\[
[\,0,\,0,\,1,\,-1,\,1,\,0\,].
\]
Multiplying \(d^1\) by an edge cochain \(y\) returns one circulation
value per filled face.

The example shows the general rule. The operator \(d^k\) evaluates a \(k\)-cochain on the oriented boundary of each \((k+1)\)-cell. This is the discrete Stokes map. It is also the basic upward route used by TNO layers for cross-rank coupling.

\paragraph{Boundaries cancel in pairs}
The boundary-based definition of \(d^k\) has an immediate consequence: applying two such maps in a row gives zero. Algebraically,
\[
B_kB_{k+1}=0,
\qquad\text{equivalently}\qquad
d^{k+1}d^k=0.
\]
The first identity says that the boundary of a cell has no remaining
boundary. The second identity is the corresponding statement on
cochains:
\[
C^k(K)
\xrightarrow{\;d^k\;}
C^{k+1}(K)
\xrightarrow{\;d^{k+1}\;}
C^{k+2}(K),
\qquad
d^{k+1}d^k=0.
\]

The reason is entirely local. 
Take a \((k+2)\)-cell and look at the
\(k\)-cells that appear after taking its boundary twice. 
Each such
\(k\)-cell appears exactly twice, once from each adjacent
\((k+1)\)-face, and the two induced orientations are opposite, meaning these two contributions cancel.  
Thus, the identity is not a limiting statement and does not depend on mesh refinement, training, or numerical accuracy. 
It holds exactly on the oriented cell complex.

In low degrees, this gives the familiar compatibility laws
\[
d^1d^0=0
\qquad\Longleftrightarrow\qquad
\operatorname{curl}(\operatorname{grad})=0,
\]
and
\[
d^2d^1=0
\qquad\Longleftrightarrow\qquad
\operatorname{div}(\operatorname{curl})=0.
\]
The first says that a field built from vertex differences has zero signed circulation around every filled face. Around a triangle, the edge differences telescope:
\[
(x_j-x_i)+(x_k-x_j)+(x_i-x_k)=0.
\]
The second says that a face field built from edge circulations has zero signed accumulation around every volume. Each interior edge contributes twice, with opposite signs, when the face circulations around the volume are summed.

\paragraph{Topology fixes incidence; geometry weights it}
The matrices \(B_k\) and \(d^k=B_{k+1}^{\top}\) know which cells touch
and how their orientations agree. They do not know lengths, areas,
volumes, angles, or material coefficients. Geometry enters through
Hodge-star, or mass, matrices
\[
M_k\in\mathbb{R}^{|K_k|\times |K_k|},
\qquad
M_k \succ 0,
\]
which define inner products on cochains:
\[
\langle u^k,v^k\rangle_{M_k}
=
(u^k)^{\top}M_kv^k .
\]
The matrix \(M_k\) stores metric and material information at degree
\(k\): edge lengths, face areas, cell volumes, anisotropic weights,
permittivity, permeability, conductivity, or other coefficient fields~\citep{hirani2003discrete,bell2012pydec}.
In primal-dual DEC, \(M_k\) can be interpreted as the map relating a
primal \(k\)-cochain to a dual \((n-k)\)-cochain.

\paragraph{The codifferential is the metric adjoint of \(d\)}
The codifferential maps downward in degree,
\[
\delta^k:C^k(K)\to C^{k-1}(K),
\]
and is defined as the adjoint of \(d^{k-1}\) with respect to the Hodge
inner products:
\[
\langle d^{k-1}u^{k-1},v^k\rangle_{M_k}
=
\langle u^{k-1},\delta^k v^k\rangle_{M_{k-1}} .
\]
In matrix form, up to the global sign convention chosen for the Hodge
star,
\[
\delta^k
=
M_{k-1}^{-1} B_k M_k .
\]
Thus \(d\) moves cochains upward by taking signed boundary sums, while
\(\delta\) moves cochains downward by the metric adjoint operation~\citep{hirani2003discrete,desbrun2005discrete}. This
is why \(\delta\) is divergence-like: it measures how a \(k\)-cell
quantity accumulates onto incident \((k-1)\)-cells after applying the
appropriate geometric weights.

\begin{figure}[h]
\centering
\includegraphics[width=0.7\textwidth]{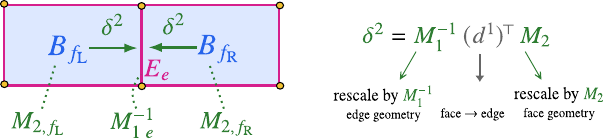}
\caption{
Codifferential as face-to-edge transport
(Eq.~\eqref{eq:codifferential-face-to-edge}): neighboring faces
\(f_{\mathrm L},f_{\mathrm R}\in K_2\) sharing edge \(e\in K_1\)
contribute via the incidence sign \([f:e]\), weighted by the Hodge
factors \(M_2\) and \(M_1^{-1}\).
}
\label{fig:codifferential-face-to-edge}
\vspace{-0.8em}
\end{figure}

Concretely, at \(k=2\) the codifferential
\(\delta^2=M_1^{-1}B_2 M_2=M_1^{-1}(d^1)^\top M_2\) sends a face cochain
\(B\) to the edge cochain
\begin{equation}
\label{eq:codifferential-face-to-edge}
(\delta^2 B)_e
=
(M_1^{-1})_{ee}
\sum_{f\succ e}
[f:e]\,(M_2)_{ff}\,B_f ,
\end{equation}
illustrated in Fig.~\ref{fig:codifferential-face-to-edge}.

This clarifies the role of the DEC maps inside a TNO. The incidence matrices \(B_k\) define the cellular routes along which information can be transported: they specify which cells are incident, how ranks are coupled, and how orientations enter the coupling. The exterior derivative \(d^k=B_{k+1}^{\top}\) moves a \(k\)-cochain upward to \((k+1)\)-cells by signed aggregation over boundaries. On the other hand, the codifferential
$
\delta^k=M_{k-1}^{-1}B_kM_k$

moves a \(k\)-cochain downward to \((k-1)\)-cells by the metric-weighted adjoint route. Thus \(d^k\) and \(\delta^k\) are not generic message-passing maps: they are structured operators for moving, coupling, and aggregating cochain features across adjacent ranks. The Hodge matrices \(M_k\) determine how geometry and material coefficients weight these transports.

\paragraph{The Hodge Laplacian has two same-rank channels}
The \(k\)-Hodge Laplacian decomposes as
\[
\Delta_k
=
\underbrace{\delta^{k+1}d^k}_{\Delta_k^\uparrow}
+
\underbrace{d^{k-1}\delta^k}_{\Delta_k^\downarrow}
:
C^k(K)\to C^k(K).
\]
Both terms start and end in \(C^k(K)\), but they route the signal through
different adjacent ranks. The upper term sends a \(k\)-cochain through
\((k+1)\)-cells and back, while the lower term sends it through
\((k-1)\)-cells and back. See Figure \ref{fig:hodge-laplacian-channels}. Thus \(\Delta_k\) is not only a same-rank
operator; it records which neighboring dimension mediates the interaction~\citep{hirani2003discrete,crane2018discrete,lim2020hodge}.

\begin{figure}[h]
\centering
\[
\begin{tikzcd}[column sep=5.8em, row sep=large]
C^{k-1}(K)
  \arrow[loop above, "{\Delta_{k-1}}"]
  \arrow[r, shift left=1.1ex, "{d^{k-1}}"]
&
C^{k}(K)
  \arrow[loop above, "{\Delta_{k}}"]
  \arrow[l, shift left=1.1ex, "{\delta^{k}}"]
  \arrow[r, shift left=1.1ex, "{d^{k}}"]
&
C^{k+1}(K)
  \arrow[loop above, "{\Delta_{k+1}}"]
  \arrow[l, shift left=1.1ex, "{\delta^{k+1}}"]
\end{tikzcd}
\]
\vspace{-0.3em}
\[
\Delta_k
=
\underbrace{\delta^{k+1}d^k}_{\Delta_k^\uparrow}
+
\underbrace{d^{k-1}\delta^k}_{\Delta_k^\downarrow}.
\]
\vspace{-0.8em}
\caption{
Upper and lower Hodge--Laplacian channels around \(C^k(K)\). The
differential \(d\) raises degree and moves rightward, while the
codifferential \(\delta\) lowers degree and moves leftward. The upper
channel \(\delta^{k+1}d^k\) propagates \(k\)-cochains through
\((k+1)\)-cells; the lower channel \(d^{k-1}\delta^k\) propagates them
through \((k-1)\)-cells. The self-loops indicate same-rank Hodge
Laplacian propagation on each cochain space.
}
\label{fig:hodge-laplacian-channels}
\vspace{-0.8em}
\end{figure}

\begin{figure}[h]
\centering
\includegraphics[width=0.72\textwidth]{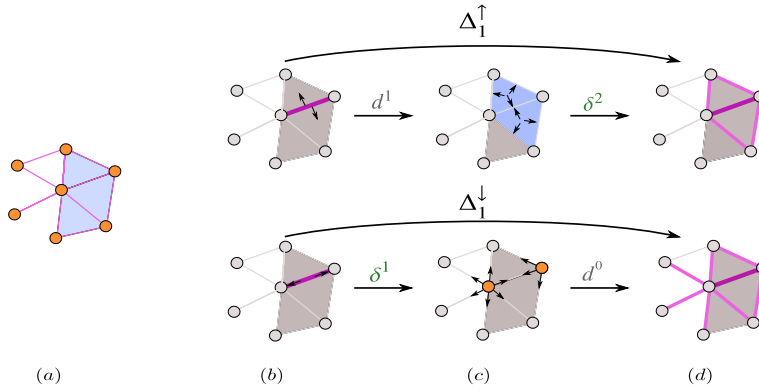}
\vspace{-0.5em}
\caption{
Upper and lower Hodge--Laplacian channels on edge cochains. Starting from
an edge-supported signal \(u\in C^1(K)\), the upper channel first applies
\(d^1:C^1(K)\to C^2(K)\), which computes signed circulation around each
face, and then applies \(\delta^2:C^2(K)\to C^1(K)\), which returns this
face quantity to incident edges through the metric-weighted adjoint route.
This gives $
\Delta_1^\uparrow u=\delta^2 d^1 u,$ a face-mediated edge-to-edge coupling. The lower channel first applies
\(\delta^1:C^1(K)\to C^0(K)\), which measures the metric-weighted
accumulation of the edge field at vertices, and then applies
\(d^0:C^0(K)\to C^1(K)\), which sends the resulting vertex potential back
to edges by oriented differences. This gives$
\Delta_1^\downarrow u=d^0\delta^1 u,$ a vertex-mediated edge-to-edge coupling. Thus \(\Delta_1^\uparrow\) and
\(\Delta_1^\downarrow\) are two distinct same-rank propagation mechanisms:
one couples edges through shared faces, the other through shared vertices.
This distinction is the degree-\(1\) DEC analogue of separating curl-type
and divergence-type effects, and is precisely what is collapsed when edge
fields are treated as ordinary graph features.}
\label{fig:edge-up-down-laplacians}
\vspace{-0.8em}
\end{figure}

For \(0\)-cochains, the lower channel is absent:
\[
\Delta_0=\delta^1 d^0,
\]
which is the usual graph or grid Laplacian on vertex fields. For
\(1\)-cochains, both channels are present:
\[
\Delta_1
=
\underbrace{\delta^2 d^1}_{\text{face-mediated}}
+
\underbrace{d^0\delta^1}_{\text{vertex-mediated}} .
\]
The first term sends edge values around faces and back to edges, coupling
edges that bound common \(2\)-cells. This is the circulation, or curl-type,
route. The second term sends edge values to vertices and back to edges,
coupling edges that meet at common \(0\)-cells. This is the accumulation,
or divergence-type, route. See Figure \ref{fig:edge-up-down-laplacians}.
Hence, an edge field has two distinct
edge-to-edge propagation mechanisms: one mediated by faces and one
mediated by vertices. Treating edge fields as ordinary graph features
collapses this distinction into a single adjacency-based route.

\paragraph{Hodge decomposition separates physical modes}
With the Hodge inner product, the \(k\)-cochain space decomposes as
\[
C^k(K;\mathbb{R})
=
\underbrace{\operatorname{im}(d^{k-1})}_{\text{exact}}
\;\oplus\;
\underbrace{\ker(\Delta_k)}_{\text{harmonic}}
\;\oplus\;
\underbrace{\operatorname{im}(\delta^{k+1})}_{\text{coexact}} .
\]
Exact cochains come from lower-degree potentials. Coexact cochains come
from higher-degree flux or circulation potentials. Harmonic cochains
satisfy
\[
d^k h=0,
\qquad
\delta^k h=0,
\qquad
\Delta_k h=0,
\]
and encode global topological degrees of freedom. In the scalar case,
\[
\dim \ker(\Delta_k)=\beta_k,
\]
the \(k\)-th Betti number~\citep{hirani2003discrete,arnold2006finite}. Thus harmonic components are not local noise:
they are the cochain modes supported by the topology of the domain.

This is one of the main reasons DEC is useful for operator learning.
Many physical fields naturally live in one of these components.
Conservative fields are exact. Divergence-free fields are coexact, up to
harmonic components. Topological modes, such as circulation around
non-contractible cycles, appear in the harmonic subspace. A model that
does not expose these channels must infer them indirectly from data.

\subsection{Continuum Physics and the mPDE Formulation}
\label{app:continuum-physics}

The governing equations of continuum physics can be phrased in terms of how different geometric dimensions interact.
An electric field $\mathbf{E}$, integrated along oriented edges, and a magnetic flux $\mathbf{B}$, integrated over oriented faces, are in this case naturally 1-cochain and 2-cochains, respectively. 
Faraday's law,
$
\partial_t B = -d^1 E,$
couples them through the exterior derivative $d^1$, which maps edge-valued data to face-valued data by accumulating signed values around each face.
In physical terms, the circulation of $\mathbf{E}$ around a face boundary equals the negative rate of change of the magnetic flux through that face~\citep{bossavit1998computational,hiptmair2002finite}.
Likewise, the incompressibility constraint $\delta^1 v = 0$ expresses divergence-freedom as a structural property rather than a penalty to enforce~\citep{elcott2007stable,mohamed2016discrete}.
More broadly, many physical laws are organized by the identity
$d^{k+1}\circ d^k = 0,$
which encodes, in a single algebraic statement, facts such as the absence of magnetic monopoles, the exactness of conservative fields, and the compatibility of flows~\citep{arnold2006finite}.
All are consequences of the same geometric principle: the boundary of a boundary is empty.
The natural discrete setting for this is a cell complex $K$ consisting of vertices, edges, faces, and higher-dimensional cells with orientations: the minimal structure on which Stokes' theorem holds exactly.
The theorem forces $B_k B_{k+1} = 0$, from which the exterior derivative $d^k = B_{k+1}^\top$, the codifferentials $\delta^k$, and the Hodge Laplacians $\Delta_k$ all follow, forming the discrete exterior calculus on $K$.
On $K$, the governing laws of a broad class of physical systems define
an operator
\begin{equation*}
    G : \bigoplus_i C^{k_i}(K;\mathbb{R}^{d_i}) \;\longrightarrow\;
    \bigoplus_j C^{\ell_j}(K;\mathbb{R}^{r_j}),
\end{equation*}
implicitly characterized by
\begin{equation}
    \mathcal{F}_k\!\left(\sigma,\; u^k,\;
    d^{k-1}u^{k-1},\;
    \delta^{k+1}u^{k+1},\;
    \Delta_k u^k,\;
    a,\; f\right)=0,
    \qquad \forall\,k,\;\forall\,\sigma\in K_k,
    \tag{mPDE}
\end{equation}
where $u^k$ are unknown cochains, $a$ encodes material parameters at
their natural degree, $f$ a source, and $\mathcal{F}_k$ a possibly
nonlinear local functional. The structure of \eqref{eq:mpde} is
what makes the problem hard for existing architectures: the equation
at degree $k$ couples $u^k$ to cochains one degree below via
$d^{k-1}u^{k-1}$ and one degree above via $\delta^{k+1}u^{k+1}$ —
cross-dimensional coupling that neither a grid nor a graph can express.
Maxwell, incompressible Navier--Stokes, and linear elasticity are all
instances of \eqref{eq:mpde}; single-degree equations are the
degenerate exception.

\subsubsection{Example PDEs}
\label{app:pde-examples}
We illustrate \eqref{eq:evol} with two standard systems whose fields live on different cochain degrees and couple through $d$ and $\delta$.
\begin{example}[Discrete Maxwell-type coupling]\label{ex:maxwell}
Maxwell's equations describe how electric and magnetic fields evolve
together: a changing magnetic field induces an electric field, and a
changing electric field, together with current, induces a magnetic
field. On an oriented $3$-complex $K$, the electric field is naturally
represented as an edge cochain $E\in C^1$, while the magnetic flux is
represented as a face cochain $B\in C^2$~\citep{bossavit1998computational,hiptmair2002finite}. The core discrete coupling has the form
\[
\partial_t B = -d^1 E,
\qquad
\partial_t E = \delta^2 B - J,
\]
with constraints
\[
d^2B=0,
\qquad
\delta^1E=\rho .
\]
Thus $E$ drives $B$ through $d^1$, which measures circulation around
faces, while $B$ drives $E$ through $\delta^2$, which aggregates adjacent
face fluxes onto edges. The essential point is not the specific physical
normalization, but the cross-degree structure: fields living on different
ranks of the complex interact through the exterior derivative and its
adjoint.
\end{example}

\begin{example}[Discrete wave-type coupling]\label{ex:wave}
The wave equation describes how a disturbance propagates through an
elastic medium: a displacement at one point pulls on its neighbors,
which pull on theirs, and the wave travels outward. Let $u\in C^0$ be a
scalar displacement on vertices, and let $p\in C^1$ be an edge variable
encoding local stretch or momentum-like flux. A first-order discrete
wave system can be written schematically as
\[
\partial_t u=\delta^1p,
\qquad
\partial_t p=-c^2d^0u ,
\]
where $c$ is the wave speed. Here $d^0$ sends vertex displacements to
edge differences, while $\delta^1$ aggregates edge quantities back to
vertices. Eliminating $p$ gives
\[
\partial_t^2u
=
-c^2\delta^1d^0u
=
-c^2\Delta_0u,
\]
showing that the wave operator itself is built from the same
cross-degree maps.
\end{example}

\section{PDE Templates as Typed Cochain Systems}
\label{sec:pde_templates}

The previous sections define the main objects of the framework: a physical
state is a tuple of cochains, the operators \(d,\delta,\Delta\) provide the
allowed transport between and within degrees, and a TNO layer learns how to mix
the transported features without learning the incidence support itself.  The
purpose of this section is to make this correspondence explicit for standard
PDE systems. 

\begin{table*}[!t]
\centering
\scriptsize
\setlength{\tabcolsep}{2.8pt}
\renewcommand{\arraystretch}{1.12}
\caption{PDE templates as typed cochain systems.  The DEC column shows the
structural skeleton of the equation, while the TNO column shows the corresponding
learned nonlinear cochain map.}
\label{tab:pde_tno_templates}

\begin{adjustbox}{max width=\textwidth,max totalheight=.78\textheight}
\begin{tabularx}{\textwidth}{@{}
>{\RaggedRight\arraybackslash}p{2.35cm}
>{\RaggedRight\arraybackslash}p{3.45cm}
>{\RaggedRight\arraybackslash}p{5.05cm}
>{\RaggedRight\arraybackslash}X
@{}}
\toprule
\textbf{PDE / system}
& \textbf{DEC core structure}
& \textbf{TNO generalization}
& \textbf{Practical challenge / insight}
\\
\midrule

Poisson / diffusion
&
\(\Delta_0u=f,\quad \Delta_0=\delta^1d^0\)
&
\(\displaystyle
H^0_{\rm out}
=
\phi_\theta(H^0,\Delta_0H^0W_\Delta,f)
\)
&
Learns a Green-type solution operator on irregular meshes.
\\

Heat equation
&
\(\partial_tu+\Delta_0u=f\)
&
\(\displaystyle
H^0_{t+\Delta t}
=
\phi_\theta(H^0_t,\Delta_0H^0_tW_\Delta,f_t,t)
\)
&
Time enters as a feature, recurrent state, or learned time-step map.
\\

Advection--diffusion \\ reaction
&
\(\begin{aligned}[t]
\partial_tc+\mathcal{L}_vc
&=\kappa\Delta_0c+R(c)
\end{aligned}\)
&
\(\begin{aligned}[t]
H^0_{\rm out}
=
\phi_\theta(&H^0,\mathcal{L}_vH^0,\\
&\Delta_0H^0W_\Delta,R_\theta(H^0))
\end{aligned}\)
&
Needs stable advection, upwinding, and control of stiff reactions.
\\

Darcy / mixed elliptic
&
\(\begin{aligned}[t]
q&=-a\,d^0p,\\
\delta^1q&=f
\end{aligned}\)
&
\(\displaystyle
H^{0:1}_{\rm out}
=
\phi_\theta(\mathcal{B}_{01}H^{0:1}W,a,f)
\)
&
Pressure and flux are kept as different cochain types; conservation is native.
\\

Conservation law
&
\(\partial_t\rho+\delta^1q=s\)
&
\(\displaystyle
H^{0:1}_{\rm out}
=
\phi_\theta(H^0,H^1,\delta^1H^1,s)
\)
&
Telescoping conservation follows from incidence algebra.
\\

Wave, first-order
&
\(\begin{aligned}[t]
\partial_tz&=\mathcal{A}_{01}z,\\
z&=(u,p)
\end{aligned}\)
&
\(\displaystyle
H^{0:1}_{t+\Delta t}
=
\phi_\theta(H^{0:1}_t,\mathcal{A}_{01}H^{0:1}_t)
\)
&
Same cross-rank structure as Darcy, but used for evolution; energy matters.
\\

Maxwell
&
\(\begin{aligned}[t]
\partial_tz&=\mathcal{A}_{12}z-(J,0),\\
z&=(E,B)
\end{aligned}\)
&
\(\displaystyle
H^{1:2}_{\rm out}
=
\phi_\theta(\mathcal{B}_{12}H^{1:2}W,J)
\)
&
Separates electric circulation from magnetic flux; gauge and Gauss constraints matter.
\\

Incompressible \\ Navier--Stokes
&
\(\begin{aligned}[t]
\partial_tu+\mathcal{L}_uu
&=-\nabla p+\nu\Delta_1u,\\
\delta^1u&=0
\end{aligned}\)
&
\(\begin{aligned}[t]
H^1_{\rm out}
=
\Pi_{\delta=0}\phi_\theta(&H^1,\mathcal{L}_{H^1}H^1,\\
&\Delta_1H^1W_\Delta,\delta^2H^2W_\delta)
\end{aligned}\)
&
Nonlinear advection must be paired with a hard divergence-free constraint.
\\

Compressible \\ Euler / NS
&
Continuity, momentum, and energy equations with fluxes
&
\(\displaystyle
H^{0:2}_{\rm out}
=
\phi_\theta(H^{0:2},dH,\delta H,\mathcal{F}_\theta(H^{0:2}))
\)
&
Requires shock-aware fluxes, positivity, and entropy consistency.
\\

Elasticity
&
\(\begin{aligned}[t]
\mathrm{div}\,\sigma&=f,\\
\sigma&=\mathbb{C}:\varepsilon(u),\\
\varepsilon(u)&=\tfrac12(d^0u+(d^0u)^\top)
\end{aligned}\)
&
\(\begin{aligned}[t]
H^1_{\rm out}
=
\phi_\theta(&H^1,\mathrm{sym}(d^0H^0)W_\varepsilon,\\
&\Delta_1H^1W_\Delta,\mathbb{C})
\end{aligned}\)
&
The symmetric gradient and constitutive law are the load-bearing structures.
\\

Shallow water
&
\(\begin{aligned}[t]
\partial_th+\delta^1(hu)&=0,\\
\text{momentum balance}
\end{aligned}\)
&
\(\begin{aligned}[t]
H^{0:1}_{\rm out}
=
\phi_\theta(&H^0,H^1,\delta^1(H^0H^1),\\
&d^0H^0,b)
\end{aligned}\)
&
Couples height, flux, bathymetry, wetting/drying, and positivity.
\\

Vorticity--stream
&
\(\begin{aligned}[t]
\partial_t\omega+\{\psi,\omega\}
&=\nu\Delta_2\omega,\\
\Delta_2\psi&=\omega
\end{aligned}\)
&
\(\begin{aligned}[t]
H^2_{\rm out}
=
\phi_\theta(&H^2,\mathrm{Poisson}_\theta(H^2),\\
&\mathcal{L}_{H^1}H^2,\Delta_2H^2W_\Delta)
\end{aligned}\)
&
The Poisson / Biot--Savart solve is the global component.
\\

Mean-field games \\
Optimal transport
&
Continuity coupled with Hamilton--Jacobi--Bellman
&
\(\begin{aligned}[t]
(\rho_{\rm out},\varphi_{\rm out})
=
\phi_\theta(&\rho,\varphi,d^0\varphi,\Delta_0\varphi,\\
&\delta^1(\rho\,d^0\varphi))
\end{aligned}\)
&
Requires monotonicity, Wasserstein geometry, and stable density--potential coupling.
\\

General \(k\)-form \\ Hodge flow
&
\(\begin{aligned}[t]
\partial_tu^k+\Delta_ku^k&=f^k,\\
\Delta_k&=\Delta_k^\uparrow+\Delta_k^\downarrow
\end{aligned}\)
&
\(\begin{aligned}[t]
H^k_{\rm out}
=
\phi_\theta(&H^k,\Delta_k^\uparrow H^kW_\uparrow,\\
&\Delta_k^\downarrow H^kW_\downarrow,f^k)
\end{aligned}\)
&
Universal form-valued PDE template; only the active operator block changes.
\\

\bottomrule
\end{tabularx}
\end{adjustbox}

\vspace{1mm}
{\scriptsize
\[
\mathcal{B}_{01}=
\begin{bmatrix}
\Delta_0 & \delta^1\\
d^0 & \Delta_1^\downarrow
\end{bmatrix},
\qquad
\mathcal{B}_{12}=
\begin{bmatrix}
\Delta_1^\uparrow & \delta^2\\
d^1 & \Delta_2^\downarrow
\end{bmatrix},
\qquad
\mathcal{A}_{01}=
\begin{bmatrix}
0 & \delta^1\\
d^0 & 0
\end{bmatrix},
\qquad
\mathcal{A}_{12}=
\begin{bmatrix}
0 & \delta^2\\
-d^1 & 0
\end{bmatrix}.
\]
}
\end{table*}

\begin{figure}[!h]
\centering
\includegraphics[width=0.7\textwidth]{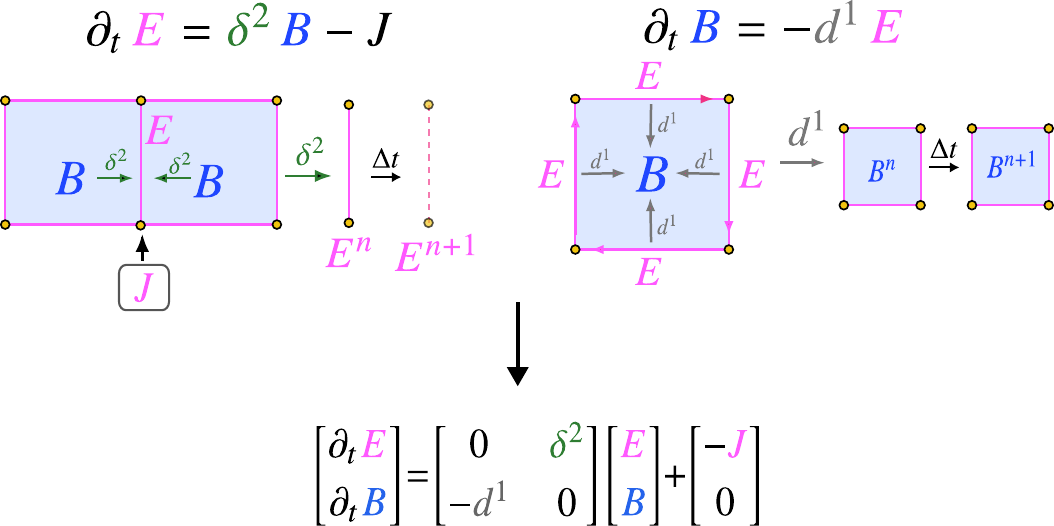}

\caption{
Maxwell coupling as cross-rank cochain transport. The electric field
$E$ lives on edges and the magnetic flux $B$ lives on faces. Faraday's
law uses $d^1$ to map edge circulation to face updates, while the
Amp\`ere--Maxwell law uses $\delta^2$ to map neighboring face fluxes
back to edge updates, with source $J$. Together, these two typed
updates form the block Maxwell system.
}
\label{fig:maxwell-block-appendix}
\end{figure}

A useful example to explain this correspondence is Maxwell's equations illustrated in Figure \ref{fig:maxwell-block-appendix}.  In a compatible discretization, the
electric field is not naturally a scalar value at vertices; it is a circulation
along oriented edges.  Likewise, the magnetic field is a flux through oriented
faces~\citep{bossavit1998computational,hiptmair2002finite}.  Thus
\[
E\in C^1(K),
\qquad
B\in C^2(K).
\]
The discrete curl is the coboundary
\[
d^1:C^1(K)\to C^2(K),
\]
which sends edge circulations to face fluxes.  Its metric adjoint,
\[
\delta^2:C^2(K)\to C^1(K),
\]
sends face fluxes back to edge-supported quantities.  Ignoring material weights
for notational clarity, the semi-discrete Maxwell system has the typed form
\[
\partial_t B=-d^1E,
\qquad
\partial_t E=\delta^2B-J,
\]
or, equivalently,
\[
\partial_t
\begin{bmatrix}
E\\
B
\end{bmatrix}
=
\begin{bmatrix}
0 & \delta^2\\
-d^1 & 0
\end{bmatrix}
\begin{bmatrix}
E\\
B
\end{bmatrix}
-
\begin{bmatrix}
J\\
0
\end{bmatrix}.
\]

\begin{figure}[t]
\centering

\begin{subfigure}[t]{0.46\textwidth}
\centering
\includegraphics[width=\linewidth]{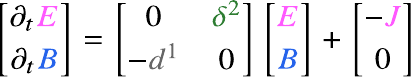}
\caption{Typed Maxwell coupling.}
\label{fig:maxwell-block}
\end{subfigure}
\hfill
\begin{subfigure}[t]{0.46\textwidth}
\centering
\includegraphics[width=\linewidth]{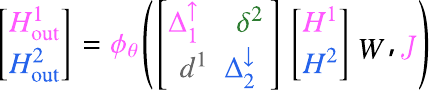}
\caption{Induced TNO layer.}
\label{fig:maxwell-tno-block}
\end{subfigure}

\vspace{-0.5em}
\caption{
From Maxwell coupling to a TNO template.
\textbf{(a)} The Maxwell system is a typed cochain system:
the electric field $E$ lives on edges, the magnetic flux $B$ lives on
faces, and the PDE couples them through the cross-rank DEC maps
$d^1:C^1\to C^2$ and $\delta^2:C^2\to C^1$, with source $J$ acting on
the edge equation.
\textbf{(b)} The corresponding TNO layer keeps this typed routing. The
off-diagonal blocks $d^1$ and $\delta^2$ transport information between
edge and face cochains, while the diagonal Hodge-Laplacian channels
$\Delta_1^\uparrow$ and $\Delta_2^\downarrow$ perform same-rank
propagation. Channel maps act on the feature dimension after DEC
transport and are suppressed in the visual notation for readability.
Learning enters through the channel weights $W^\bullet$ and the nonlinear
update $\phi_\theta$, which determine how the routed features are mixed
rather than where information can flow.
}
\label{fig:maxwell-block-to-tno}
\vspace{-0.8em}
\end{figure}

This is precisely the kind of coupling encoded by the TNO block: the equation
does not merely say that two feature arrays interact; it specifies that one
field lives on edges, the other on faces, and that the interaction must pass
through the incidence operators of the complex.  The corresponding learned
cochain update has the same structural support,
\[
\begin{bmatrix}
H^1_{\rm out}\\
H^2_{\rm out}
\end{bmatrix}
=
\phi_\theta\!\left(
\begin{bmatrix}
\Delta_1^\uparrow & \delta^2\\
d^1 & \Delta_2^\downarrow
\end{bmatrix}
\begin{bmatrix}
H^1\\
H^2
\end{bmatrix}
W,\,
J
\right).
\]
Here \(H^1\) and \(H^2\) are hidden edge and face cochains.  The block matrix
fixes where information can flow: edge-to-face through \(d^1\), face-to-edge
through \(\delta^2\), and same-rank propagation through the Hodge Laplacian
channels.  The source \(J\in C^1(K)\) is an external edge-supported input to
the update, not part of the DEC transport block. The trainable part, \(W\) and
\(\phi_\theta\), learns how these transported features are combined. The signs
and physical constants in the Maxwell operator may be fixed explicitly or
absorbed into the learned channel maps; the displayed TNO block is meant to
preserve the typed incidence support of the PDE. See Figure
\ref{fig:maxwell-block-to-tno}.

This example also shows the exact-sequence structure. Namely, the identity \(d^2d^1=0\) implies
\[
\partial_t(d^2B)=-d^2d^1E=0.
\]
Hence, the magnetic Gauss constraint is preserved by the algebra of the complex.
Similarly, \(\delta^1\delta^2=0\), together with the discrete charge-continuity law, gives the corresponding compatibility condition for the electric equation.
These identities are inherited from the incidence structure, not learned as soft penalties.

\paragraph{Darcy (following \cref{sec:experiments})}
The Darcy / mixed-elliptic system studied in
\cref{sec:experiments} constructs a typed routing across all three cochain ranks of a \(2\)D complex. 
The pressure is defined as a scalar at the vertices, and the Darcy flux is a line integral along oriented edges, and the diffusivity is face-supported (a scalar on faces in the isotropic case, a rank-\(2\) tensor on faces in \cref{sec:rankin-aniso}):
\[
p\in C^0(K),\qquad q\in C^1(K),\qquad a\in C^2(K),
\]
with constitutive and balance equations
\[
q=-\mathcal{K}_a d^0p,\qquad \delta^1q=f.
\]
Here \(\mathcal{K}_a:C^1(K)\to C^1(K)\) is the Darcy constitutive map induced by the face-supported coefficient \(a\), mapping edge gradients to edge fluxes~\citep{hyman1997numerical,lipnikov2014mimetic}.
A TNO block on this system carries hidden pressure and flux cochains
\(H^0,H^1\) and routes them through the \(\mathcal{B}_{01}\) block of
Table~\ref{tab:pde_tno_templates}; the material field \(a\) enters at its
native rank as a face-valued input channel that conditions the
constitutive map.  
This is the synthetic mesh family we examine in
\cref{sec:rankin-aniso}: when the per-face orientation is constructed
so that vertex projection is lossy by design, an architecture
that ingests \(a\) only at \(C^0\) cannot recover the operator, while the
rank-\(2\) ingestion path retains the information that makes the typed
routing load-bearing.  
The same incidence algebra makes the residual \(\delta^1q-f\) a
vertex-supported quantity computed exactly by the complex. Thus mass balance
can be enforced, penalized, or projected at the correct cochain degree rather
than approximated through an untyped graph message.

Together, Maxwell and Darcy cover two qualitatively distinct roles a
typed coupling can play: evolutionary cross-rank dynamics with gauge
structure (Maxwell), and an elliptic constraint with native discrete
conservation (Darcy).  
Poisson activates a single \(0\)-cochain and the scalar Hodge Laplacian
\(\Delta_0\).  
Fluids, elasticity, conservation laws, and transport systems introduce
problem-specific nonlinearities, such as advection, constitutive maps, learned
fluxes, projections, or positivity constraints.  
In all cases, the same principle remains: the PDE determines the typed
transport, while the TNO learns the nonlinear cochain map supported on that
transport.

Table \ref{tab:pde_tno_templates} summarizes this correspondence and should be
read as an unpacking of the multi-degree PDE form and the block TNO layer,
rather than as a collection of separate architectures.

\section{Proofs}

\subsection{Proof of \cref{prop:fno-recovery} [FNO-Recovery]}\label{app:proof-fno}

\begin{proof}
On a uniform periodic cubical complex, the rank-\(0\) Hodge Laplacian
\(\Delta_{0,N}=\delta^1 d^0\) is translation invariant and is diagonalized by
the discrete Fourier basis:
\[
\Delta_{0,N}
=
\mathcal F_N^{-1}\Lambda_N\mathcal F_N .
\]
The zero eigenspace is the space of constant \(0\)-cochains, hence
\[
P^{\rm harm}_0 u
=
\mathcal F_N^{-1}
\big(\mathbf 1_{\xi=0}\widehat u(\xi)\big),
\qquad
(I-P^{\rm harm}_0)u
=
\mathcal F_N^{-1}
\big(\mathbf 1_{\xi\neq 0}\widehat u(\xi)\big).
\]

Define the rank-\(0\) spectral TNO channel
\[
S_{\theta,N}u
=
\mathcal F_N^{-1}
\big(R_\theta(\xi)\widehat u(\xi)\big)_{\xi}
\]
with the usual real-valuedness condition
\[
R_\theta(-\xi)=\overline{R_\theta(\xi)} .
\]
Equivalently, separating the harmonic and non-harmonic parts,
\[
S_{\theta,N}u
=
\mathcal F_N^{-1}
\big(\mathbf 1_{\xi\neq 0}R_\theta(\xi)\widehat u(\xi)\big)
+
\mathcal F_N^{-1}
\big(\mathbf 1_{\xi=0}R_\theta(0)\widehat u(0)\big).
\]
Thus
\[
S_{\theta,N}u
=
S_{\theta,N}(I-P^{\rm harm}_0)u
+
R_\theta(0)P^{\rm harm}_0u .
\]

Now choose the rank-\(0\) TNO layer
\[
T_{\theta,N}(u)
=
W_{\rm self}u
+
S_{\theta,N}(I-P^{\rm harm}_0)u
+
W_{\rm harm}P^{\rm harm}_0u ,
\]
with
\[
W_{\rm self}=W,
\qquad
W_{\rm harm}=R_\theta(0).
\]
Then
\[
\begin{aligned}
T_{\theta,N}(u)
&=
Wu
+
\mathcal F_N^{-1}
\big(\mathbf 1_{\xi\neq 0}R_\theta(\xi)\widehat u(\xi)\big)
+
\mathcal F_N^{-1}
\big(\mathbf 1_{\xi=0}R_\theta(0)\widehat u(0)\big) \\
&=
Wu+\mathcal F_N^{-1}R_\theta\mathcal F_Nu .
\end{aligned}
\]
This is exactly the linear FNO layer.
\end{proof}

\subsection{Proof of \cref{prop:gno-recovery} [GNO-Recovery]}
\label{app:proof-gno}

\begin{proof}
A graph neural operator approximates an integral operator of the form
\[
    (\mathcal K_\theta u)(x)
    =
    \int_D \kappa_\theta(x,y)u(y)\,d\mu(y),
\]
where $\kappa_\theta(x,y)$ is a learned kernel acting on the feature value
at the source point $y$ and producing a contribution at the query point
$x$. On a finite point cloud $\{x_i\}_{i=1}^n$, this integral is replaced
by a quadrature rule over a neighborhood $\mathcal N(i)$ of each query
point:
\[
    (\mathcal K_\theta u)(x_i)
    \approx
    \sum_{j\in\mathcal N(i)}
    \kappa_\theta(x_i,x_j)u_j\omega_j,
\]
where $\omega_j>0$ is the quadrature weight associated with the source
point $x_j$. A GNO layer then adds a pointwise channel map:
\[
    u_i
    \longmapsto
    Wu_i+
    \sum_{j\in\mathcal N(i)}
    \kappa_\theta(x_i,x_j)u_j\omega_j .
\]

We now realize this update as a rank-$0$ graph-quadrature TNO. Construct a
directed graph complex $G=(V,E)$ with one vertex $v_i$ for each sample
point $x_i$. For every quadrature interaction $j\in\mathcal N(i)$, add a
directed rank-$1$ cell
\[
    e_{ij}: v_j \to v_i .
\]
Thus each edge represents one source-to-target quadrature contribution.

Define a rank-$1$ cochain $h^1\in C^1(G;\mathbb R^{d_h})$ by assigning to
each directed edge the corresponding learned kernel message:
\[
    h^1(e_{ij})
    =
    \kappa_\theta(x_i,x_j)u_j\omega_j .
\]
This is a cellular message: it lives on the rank-$1$ cell $e_{ij}$ and is
supported only where the graph complex contains an incidence from source
$v_j$ to target $v_i$.

Let
\[
    S_{\mathrm{tar}}:C^1(G;\mathbb R^{d_h})\to C^0(G;\mathbb R^{d_h})
\]
denote the target-incidence aggregation operator
\[
    (S_{\mathrm{tar}}h^1)_i
    =
    \sum_{e_{ij}: \operatorname{tar}(e_{ij})=v_i} h^1(e_{ij})
    =
    \sum_{j\in\mathcal N(i)} h^1(e_{ij}).
\]
Equivalently, $S_{\mathrm{tar}}$ is the unsigned incoming-incidence matrix
of the directed graph. It is important that this is the incoming-incidence
aggregation, not the signed boundary operator: the signed boundary would
also include source signs and would not equal the GNO quadrature sum in
general.

Substituting the definition of $h^1$ gives
\[
    (S_{\mathrm{tar}}h^1)_i
    =
    \sum_{j\in\mathcal N(i)}
    \kappa_\theta(x_i,x_j)u_j\omega_j .
\]
Therefore the rank-$0$ TNO update with self-channel $W_0^{\mathrm{self}}=W$
is
\[
    \mathcal T_{\theta,0}(u)_i
    =
    W u_i + (S_{\mathrm{tar}}h^1)_i
    =
    Wu_i+
    \sum_{j\in\mathcal N(i)}
    \kappa_\theta(x_i,x_j)u_j\omega_j .
\]
This is exactly the GNO graph-quadrature update. Applying the pointwise
nonlinearity after this affine update gives the usual nonlinear GNO layer.
\end{proof}

\subsection{Cell-wise Realization and Copresheaf Variant}
\label{app:cellwise-tno}

The block update in \cref{sec:tno-layers} can be written cell by cell.
This form makes explicit that a TNO layer is local, but with locality
defined by DEC rather than by an arbitrary graph adjacency. The relevant
neighborhoods are induced by the routes \(d\), \(\delta\), and the two
Hodge--Laplacian channels \(\Delta^\uparrow,\Delta^\downarrow\).

Let \(\widetilde K\) be the lifted complex used by the network, and let
\(\widetilde K_k\) denote its \(k\)-cells. For a hidden \(k\)-cochain
\(h^k\in C^k(\widetilde K;\mathbb{R}^{d_k})\), write
\[
    h^k_\sigma := h^k(\sigma),
    \qquad \sigma\in \widetilde K_k .
\]
For a target \(k\)-cell \(\sigma\), define the four DEC neighborhoods
\[
\mathcal N^k_{\mathrm{ex}}(\sigma)
=
\{\rho\in \widetilde K_{k-1}:\rho\prec\sigma\},
\qquad
\mathcal N^k_{\mathrm{coex}}(\sigma)
=
\{\eta\in \widetilde K_{k+1}:\sigma\prec\eta\},
\]
and
\[
\mathcal N^k_{\Delta^\uparrow}(\sigma)
=
\{\tau\in \widetilde K_k:(\Delta_k^\uparrow)_{\sigma\tau}\neq 0\},
\qquad
\mathcal N^k_{\Delta^\downarrow}(\sigma)
=
\{\tau\in \widetilde K_k:(\Delta_k^\downarrow)_{\sigma\tau}\neq 0\}.
\]
The first two neighborhoods are cross-rank. The exact route
\(\mathcal N^k_{\mathrm{ex}}\) receives from \((k-1)\)-cells through
\(d^{k-1}\), while the coexact route \(\mathcal N^k_{\mathrm{coex}}\)
receives from \((k+1)\)-cells through \(\delta^{k+1}\). The last two
neighborhoods are same-rank. The upper Laplacian route couples
\(k\)-cells through common \((k+1)\)-cells, while the lower Laplacian
route couples \(k\)-cells through common \((k-1)\)-cells. For \(k=1\),
these are edge-to-edge couplings through shared faces and shared
vertices, respectively.

We now attach a copresheaf to each route. For each route
\[
r\in\{\mathrm{ex},\mathrm{coex},\Delta^\uparrow,\Delta^\downarrow\},
\]
the neighborhood function \(\mathcal N^k_r\) induces a directed graph
whose edges are
\[
    y\to \sigma
    \qquad\text{whenever}\qquad
    y\in \mathcal N^k_r(\sigma).
\]
A route-dependent copresheaf assigns a feature space \(F_x\) to every
cell \(x\) appearing in this directed graph and a linear transport map
\[
    \rho^r_{y\to\sigma}:F_y\to F_\sigma
\]
to every directed neighborhood edge \(y\to\sigma\). In implementation,
one often takes \(F_\sigma\simeq\mathbb R^{d_k}\) for all \(k\)-cells
\(\sigma\), but the notation allows different ranks, cell types, or
incidences to carry different feature spaces.

The DEC scalar coefficient along each route is kept separate from the
copresheaf map. Define
\[
a^{k,\mathrm{ex}}_{\sigma\rho}=[\sigma:\rho],
\qquad
a^{k,\mathrm{coex}}_{\sigma\eta}=(\delta^{k+1})_{\sigma\eta},
\]
and
\[
a^{k,\Delta^\uparrow}_{\sigma\tau}
=
(\Delta_k^\uparrow)_{\sigma\tau},
\qquad
a^{k,\Delta^\downarrow}_{\sigma\tau}
=
(\Delta_k^\downarrow)_{\sigma\tau}.
\]
The coefficient \(a^{k,\mathrm{ex}}\) is purely incidence and
orientation. The coefficients involving \(\delta\) and \(\Delta\) carry
the metric or material weights induced by the Hodge stars. The learned
copresheaf map \(\rho^r_{y\to\sigma}\) then transports the source feature
into the target feature space. Equivalently, the scalar DEC coefficient
could be absorbed into \(\rho^r_{y\to\sigma}\), but separating them makes
clear what is fixed by DEC and what is learned.

The cell-wise update has four routed copresheaf channels:

\[
m^{k,\mathrm{ex}}(\sigma)
=
\bigoplus_{\rho\in\mathcal N^k_{\mathrm{ex}}(\sigma)}
\psi^{\mathrm{ex}}_\theta
\!\left(
h^k_\sigma,\;
a^{k,\mathrm{ex}}_{\sigma\rho}\,
\rho^{\mathrm{ex}}_{\rho\to\sigma}h^{k-1}_\rho
\right).
\]
\[
m^{k,\mathrm{coex}}(\sigma)
=
\bigoplus_{\eta\in\mathcal N^k_{\mathrm{coex}}(\sigma)}
\psi^{\mathrm{coex}}_\theta
\!\left(
h^k_\sigma,\;
a^{k,\mathrm{coex}}_{\sigma\eta}\,
\rho^{\mathrm{coex}}_{\eta\to\sigma}h^{k+1}_\eta
\right).
\]
\[
m^{k,\Delta^\uparrow}(\sigma)
=
\bigoplus_{\tau\in\mathcal N^k_{\Delta^\uparrow}(\sigma)}
\psi^{\Delta^\uparrow}_\theta
\!\left(
h^k_\sigma,\;
a^{k,\Delta^\uparrow}_{\sigma\tau}\,
\rho^{\Delta^\uparrow}_{\tau\to\sigma}h^k_\tau
\right).
\]
\[
m^{k,\Delta^\downarrow}(\sigma)
=
\bigoplus_{\tau\in\mathcal N^k_{\Delta^\downarrow}(\sigma)}
\psi^{\Delta^\downarrow}_\theta
\!\left(
h^k_\sigma,\;
a^{k,\Delta^\downarrow}_{\sigma\tau}\,
\rho^{\Delta^\downarrow}_{\tau\to\sigma}h^k_\tau
\right).
\]

Here \(\bigoplus\) is a permutation-invariant aggregation, such as a sum,
mean, or attention-weighted sum. Each message first transports the
source feature through the copresheaf map
\(\rho^r_{y\to\sigma}\), then weights it by the DEC coefficient of the
corresponding route, and finally applies the learnable message function
\(\psi_\theta^r\).

The output feature at \(\sigma\) is
\[
h^k_{\mathrm{out}}(\sigma)
=
\phi_\theta\!\left(
h^k_\sigma,\;
m^{k,\mathrm{ex}}(\sigma),\;
m^{k,\mathrm{coex}}(\sigma),\;
m^{k,\Delta^\uparrow}(\sigma),\;
m^{k,\Delta^\downarrow}(\sigma)
\right).
\]
The functions \(\psi_\theta^r\) form route-wise messages, while
\(\phi_\theta\) mixes the self feature with the four routed channels.
The DEC operators determine which cells can communicate and provide the
orientation/Hodge coefficients. The copresheaf maps determine how
features are transported between the local feature spaces attached to
the source and target cells.

In operator form, the route-wise copresheaf transports are
\[
(d^{k-1}_{\rho}h^{k-1})_\sigma
=
\sum_{\rho\prec\sigma}
[\sigma:\rho]\,
\rho^{\mathrm{ex}}_{\rho\to\sigma}h^{k-1}_\rho,
\]
\[
(\delta^{k+1}_{\rho}h^{k+1})_\sigma
=
\sum_{\eta\succ\sigma}
(\delta^{k+1})_{\sigma\eta}\,
\rho^{\mathrm{coex}}_{\eta\to\sigma}h^{k+1}_\eta,
\]
and
\[
(\Delta^{\uparrow}_{k,\rho}h^k)_\sigma
=
\sum_{\tau\in\mathcal N^k_{\Delta^\uparrow}(\sigma)}
(\Delta_k^\uparrow)_{\sigma\tau}\,
\rho^{\Delta^\uparrow}_{\tau\to\sigma}h^k_\tau,
\]
\[
(\Delta^{\downarrow}_{k,\rho}h^k)_\sigma
=
\sum_{\tau\in\mathcal N^k_{\Delta^\downarrow}(\sigma)}
(\Delta_k^\downarrow)_{\sigma\tau}\,
\rho^{\Delta^\downarrow}_{\tau\to\sigma}h^k_\tau.
\]
Thus the copresheaf variant replaces the fixed DEC transports by
route-wise transported DEC operators:
\[
d^{k-1}\rightsquigarrow d^{k-1}_{\rho},
\qquad
\delta^{k+1}\rightsquigarrow \delta^{k+1}_{\rho},
\qquad
\Delta_k^\uparrow,\Delta_k^\downarrow
\rightsquigarrow
\Delta^{\uparrow}_{k,\rho},\Delta^{\downarrow}_{k,\rho}.
\]

The fixed DEC layer is recovered by taking the copresheaf maps to be
identity maps, or shared channel maps independent of the incident pair.
Then the only route coefficients are the signed incidence coefficients
and the Hodge-weighted DEC coefficients. The copresheaf version keeps the
same DEC supports but allows the feature transport itself to vary by
route, rank, orientation, geometry, cell type, or learned attributes.
This gives a controlled extension of the DEC layer: the topology fixes
the admissible routes, while the copresheaf maps learn how features are
translated along those routes.

\section{TNOs, Hodge-Compatible Smoothing, and the Multigrid Correspondence}
\label{app:multigrid}

We now explain the numerical-analysis interpretation of TNO and HTNO.
The point is not that a trained TNO inherits the convergence theory of classical multigrid. 
It does not. 
Rather, the claim is structural: the four channels of the TNO layer, the use of residual depth, and the hierarchical construction of HTNO reproduce the algebraic ingredients that make multigrid work for de Rham complexes. 
These are precisely the ingredients missing from
ordinary point smoothers on vector-valued nodal fields.

\subsection{Why scalar multigrid does not directly extend}

\paragraph{How scalar multigrid works}
For the scalar Poisson problem
\[
    -\Delta_0 u = f, \qquad u \in C^0(K),
\]
geometric multigrid succeeds because of a clean separation of scales in
the error. The error $e = u - u_h$ between the true solution and the
current iterate can be thought of as a sum of components that oscillate
at different spatial frequencies on the mesh. After a few sweeps of a
local smoother---a simple iterative method such as Gauss--Seidel or
damped Jacobi that updates each unknown from a small patch of the
mesh---the high-frequency (rapidly oscillating) part of the error is
reduced efficiently. What remains varies slowly across the mesh and is
therefore well represented on a coarser version of the same grid. The
coarse-grid correction step solves a cheaper problem there and returns a
correction to the fine grid; cycling between levels yields a combined
iteration whose convergence rate is independent of the mesh size under
standard ellipticity and approximation assumptions
~\citep{trottenberg2000multigrid,bramble1991analysis}.

The critical point is that for $-\Delta_0$ \emph{every} error component
is handled by one of the two mechanisms: either the smoother kills it, or
the coarse grid represents it. No part of the error escapes both.

\paragraph{The curl--curl problem and its large kernel}
This clean picture breaks for problems involving differential forms, that
is, for unknowns that live on edges, faces, or cells of the mesh rather
than on nodes. Consider the curl--curl block
\[
    \delta^2 d^1 e + \tau e = f, \qquad e \in C^1(K),
\]
where $e$ is an edge-valued unknown (a discrete $1$-form), $d^1$ is the
discrete curl operator (mapping edges to faces), and $\delta^2$ is its
$L^2$-adjoint (mapping faces back to edges). This equation appears in
time-harmonic Maxwell equations, magnetostatic vector-potential
formulations, and the degree-one Hodge Laplacian---the natural
generalization of the scalar Laplacian to edge-valued fields
~\citep{monk1992finite,monk2003finite,bossavit1998computational,
hiptmair2002finite,arnold2006finite,arnold2010finite}.
When $\tau = 0$ the operator $\delta^2 d^1$ is only positive
semidefinite; even for small $\tau > 0$ the near-null space is so large
that the same practical difficulties arise.

The source of trouble is a fundamental identity in the de~Rham complex.
The discrete gradient $d^0$ maps node values to edge values (it assigns
to each edge the signed difference of the node values at its endpoints),
and the discrete curl $d^1$ maps edge values to face values. These two
operators satisfy
\[
    d^1 d^0 = 0,
\]
which is simply the fact that the curl of any gradient is zero.
Consequently, every edge cochain of the form $d^0\phi$---a discrete
gradient field---is in the kernel of the curl--curl operator:
\[
    \delta^2 d^1 d^0\phi = 0.
\]
This means
\[
    \operatorname{im} d^0 \subset \ker(\delta^2 d^1),
\]
and the dimension of this invisible subspace is
\[
    \dim \operatorname{im} d^0 = \dim C^0 - \beta_0,
\]
where $\beta_0$ is the number of connected components of the domain
(typically one). On topologically nontrivial domains---those with loops
or handles---there are further null modes corresponding to harmonic
representatives, fields that are curl-free and divergence-free but not
globally a gradient~\citep{hiptmair2002finite,arnold2006finite,
arnold2010finite}.

\paragraph{Why a smoother is blind to gradient error}
To see concretely why this causes trouble, suppose the current error $e$
contains a gradient component $e_{\mathrm{grad}} = d^0\phi$ for some
node cochain $\phi$. This component contributes nothing to the curl--curl
residual:
\[
    \delta^2 d^1 e_{\mathrm{grad}} = 0.
\]
A smoother that updates the edge unknowns using only the curl--curl
residual---regardless of how many sweeps are applied---never sees this
component and therefore never reduces it. The iteration is not merely
\emph{slow} on gradient error; it is \emph{completely blind} to it. This
is a qualitative failure, not a quantitative one.

Compare with the scalar Poisson case: there, the Laplacian $-\Delta_0$
has a null space of dimension one (constant functions on a closed domain),
handled by fixing a single degree of freedom. Here the null space has
dimension equal to the number of nodes, comparable to the total number of
unknowns. It cannot be removed by a simple constraint; it must be
explicitly visited by the solver at every level.

\paragraph{The Hodge decomposition and three-component obstruction.}
The full picture is given by the Hodge decomposition, which states that
any edge cochain can be written as the sum of three mutually orthogonal
pieces~\citep{arnold2000multigrid,arnold2006finite,arnold2010finite}:
\[
    C^k = \underbrace{\operatorname{im} d^{k-1}}_{\text{exact (gradient)}}
     \;\oplus\;
     \underbrace{\operatorname{im} \delta^{k+1}}_{\text{coexact}}
     \;\oplus\;
     \underbrace{\mathcal{H}^k}_{\text{harmonic}},
\]
where $\mathcal{H}^k = \ker d^k \cap \ker \delta^k$ is the space of
fields that are simultaneously curl-free and divergence-free. Each piece
behaves differently under the Hodge Laplacian $\Delta_k = \delta^{k+1}d^k
+ d^{k-1}\delta^k$:

\begin{itemize}
    \item The \textbf{exact part} $\operatorname{im} d^{k-1}$ lies in the
    kernel of the upper block $\delta^{k+1}d^k$; it is governed entirely
    by the lower block $d^{k-1}\delta^k$ and is invisible to any smoother
    that acts only through the upper block.

    \item The \textbf{coexact part} $\operatorname{im}\delta^{k+1}$ lies
    in the kernel of the lower block; it is governed by the upper block
    and is what a standard curl--curl smoother can address.

    \item The \textbf{harmonic part} $\mathcal{H}^k$ lies in the kernel
    of both blocks and is in the null space of $\Delta_k$; its dimension
    equals the $k$-th Betti number $\beta_k$, a topological invariant of
    the domain.
\end{itemize}

A scalar smoother acting through a single energy can handle at most one
of these three components. The other two are left entirely untouched.
This is the fundamental obstruction: the difficulty is not that the
unknown is edge-valued, but that the operator has a topologically
structured null and near-null space that a uniform local relaxation
cannot see~\citep{hiptmair2002finite,arnold2000multigrid,arnold2006finite,
arnold2010finite}.

\paragraph{Consequences for solver design.}
Robust solvers for $H(\mathrm{curl})$, $H(\mathrm{div})$, and
Hodge--Laplacian problems must therefore be designed to address all three
Hodge components explicitly:

\begin{itemize}
    \item \textbf{Coarse spaces} must carry a compatible discrete
    complex---a coarse-level version of the gradient, curl, and
    divergence operators that mirrors the fine-level structure, so that
    gradient error on the fine grid is represented as gradient error on
    the coarse grid.

    \item \textbf{Transfer operators} (prolongation and restriction) must
    commute with the discrete differential operators $d^k$, so that a
    gradient on the coarse grid prolongs to a gradient on the fine
    grid---not to a spurious mix of components.

    \item \textbf{Smoothers} must address all three Hodge components,
    either by augmenting the edge smoother with a nodal (scalar) solve
    that targets gradient error---Hiptmair's subspace decomposition
    smoother~\citep{hiptmair1998multigrid}---or by constructing block
    smoothers that simultaneously see both blocks of $\Delta_k$.
\end{itemize}

These requirements are the mechanism behind specialized Maxwell multigrid,
auxiliary-space preconditioners~\citep{hiptmair2007nodal}, and
FEEC-based multilevel methods~\citep{hiptmair1998multigrid,
arnold2000multigrid,hiptmair2002finite,arnold2006finite,arnold2010finite,
arnold2018feec}.

\subsection{Hiptmair smoothing}
Hiptmair's remedy is to smooth not only on the space of the unknown, but
also on the neighboring spaces in the de~Rham complex
~\citep{hiptmair1998multigrid}. For an edge variable $e \in C^1(K)$,
the smoother combines:
\begin{enumerate}[noitemsep,leftmargin=*,topsep=0em]
    \item a relaxation on $C^1$, which acts on the coexact component;
    \item an auxiliary relaxation on $C^0$, lifted by $d^0$, which acts
    on the exact component~\citep{hiptmair1998multigrid,hiptmair2007nodal};
    \item a harmonic correction when $\beta_1>0$~\citep{hiptmair1998multigrid}.
\end{enumerate}
This matches the discrete Hodge decomposition
~\citep{hirani2003discrete,tong2003multiscale,
arnold2006finite,arnold2010finite}
\[
    C^1(K)
    =
    \operatorname{im} d^0
    \oplus
    \mathcal H^1
    \oplus
    \operatorname{im}\delta^2 .
\]
At degree $k$, the same principle uses the neighboring spaces $C^{k-1}$
and $C^{k+1}$ together with the harmonic subspace $\mathcal H^k =
\ker\Delta_k$~\citep{hiptmair1998multigrid,arnold2000multigrid,
arnold2006finite}. The lesson is simple but important: a Hodge-compatible
smoother for $k$-cochains must communicate with adjacent cochain degrees
through the discrete differential and codifferential. This is not an
implementation detail; it is what removes the kernel blindness of ordinary
local relaxation~\citep{hiptmair1998multigrid,hiptmair2002finite,
arnold2000multigrid}.

\subsection{The TNO layer as a Hodge-structured update}
The linear TNO layer at degree $k$ has the form
\[
T_{\theta,k}(h)
=
d^{k-1} h^{k-1} W^{\downarrow}_k
+
\delta^{k+1} h^{k+1} W^{\uparrow}_k
+
P^{\mathrm{harm}}_k h^k W^{\mathrm{harm}}_k
+
h^k W^{\mathrm{self}}_k .
\]
The first three terms land in the three Hodge components of $C^k$
~\citep{hirani2003discrete,arnold2006finite,arnold2010finite}:
\[
\operatorname{im} d^{k-1},
\qquad
\operatorname{im}\delta^{k+1},
\qquad
\mathcal H^k=\ker \Delta_k .
\]
The final term is a local residual channel on the full space $C^k$.
The branch $d^{k-1}h^{k-1}$ moves lower-degree information into degree $k$ through the exact
component. This is analogous to the auxiliary-space correction in
Hiptmair-type smoothers, where a lower-degree potential is lifted into
the target cochain degree~\citep{hiptmair1998multigrid,hiptmair2007nodal}.
The branch $\delta^{k+1}h^{k+1}$
moves higher-degree information into degree $k$ through the coexact
component. The harmonic branch $P^{\mathrm{harm}}_k h^k$
extracts the component not seen by either $d^k$ or $\delta^k$, while the
self-channel provides a local update on $C^k$.
Thus the layer has the same typed correction paths used by
Hodge-decomposition-based smoothers~\citep{hiptmair1998multigrid,
arnold2000multigrid,arnold2006finite}, but with learned channel maps in
place of fixed scalar relaxation parameters. This is a structural analogy
rather than a multigrid convergence statement: a classical smoother
updates solver residuals, whereas a TNO layer updates learned cochain
features.

\subsection{Depth as preconditioned iteration}
The residual TNO update
\[
h^k_{\ell+1}
=
h^k_\ell
+
H^k_\theta
\bigl(
h^{k-1}_\ell,
h^k_\ell,
h^{k+1}_\ell,
d^{k-1}h^{k-1}_\ell,
\delta^{k+1}h^{k+1}_\ell,
\Delta_k h^k_\ell,
a,
f
\bigr)
\]
has the form of a stationary residual iteration~\citep{saad2003iterative}.
If $A$ is the discrete PDE operator and $B_\theta$ is the learned
Hodge-compatible preconditioner encoded by the layer, then the idealized
update is
\[
    h_{\ell+1}
    =
    h_\ell
    +
    B_\theta^{-1}(f-Ah_\ell),
\]
where $B_\theta^{-1}(\cdot)$ denotes the action of the preconditioner.
Stacking $L$ residual layers is therefore analogous to applying $L$ steps
of a preconditioned Richardson iteration~\citep{saad2003iterative,
trottenberg2000multigrid,briggs2000multigrid}. This gives a concrete
interpretation of depth: it is not merely additional expressivity, it is
additional correction time. The diminishing returns observed in our
experiments with increasing depth are consistent with this solver
interpretation: once the dominant residual components have been reduced,
further iterations contribute less.

\paragraph{Residual layers and physical structure}
When the target \(\mathcal{G}\) arises from a discrete PDE, the architecture should reflect the structure of that law.
A \emph{residual TNO layer} therefore keeps the DEC operators
\(d^k,\delta^k,\Delta_k\) as fixed transport scaffolding determined by the
cell complex and its Hodge stars, and learns only how the transported cochain
features are mixed and corrected. Writing
\(\mathbf{H}_\ell = (H^0_\ell, \ldots, H^N_\ell)\) for the hidden cochains at
layer \(\ell\) on the lifted complex \(\widetilde{K}\), the residual form of
\(\mathbf{H}_{out}\) (as in \cref{sec:tno-layers}) is
\begin{equation}
  \mathbf{H}_{\ell+1}
  =
  \mathbf{H}_\ell
  +
  \phi_\theta\!\bigl(\mathcal{D}\,\mathbf{H}_\ell W\bigr),
  \label{eq:rtnl}
\end{equation}
where \(\mathcal{D}\) is the block Dirac-type operator in
\cref{sec:tno-layers}~\citep{bianconi2021dirac,calmon2022dirac}, with
\(\Delta_k\) on the diagonal and \(d^{k-1},\delta^{k+1}\) on the off-diagonals.
Coefficient and source cochains \(a,f\) enter as additional input channels at
the appropriate ranks, consistent with (P1). Per rank, this expands to
\begin{equation}
  H^k_{\ell+1}
  =
  H^k_\ell
  +
  \phi_\theta\!\Bigl(
    H^k_\ell,\;
    d^{k-1}H^{k-1}_\ell\,W^\downarrow_k,\;
    \delta^{k+1}H^{k+1}_\ell\,W^\uparrow_k,\;
    \Delta_k^\uparrow H^k_\ell\,W^{\Delta,\uparrow}_k,\;
    \Delta_k^\downarrow H^k_\ell\,W^{\Delta,\downarrow}_k
  \Bigr),
  \label{eq:rtnl-perrank}
\end{equation}
with out-of-range terms omitted. 
The residual connection stabilizes training and makes the identity easy to
represent, which is useful for small-\(\Delta t\) time-stepping operators.

\paragraph{Per-layer pseudocode}
\Cref{alg:tno-layer} unrolls one residual TNO layer in the form actually computed at runtime: each rank-$k$ block is updated from its own state and the transported contributions of its neighboring ranks, with the supports prescribed by the DEC operators of $\widetilde{K}$. 
The block operator $\mathcal{D}$ of~\cref{sec:tno-layers} already encodes the cross-rank routing, so no separate message-passing scaffolding is required: applying $\mathcal{D}$ to $\mathbf{H}_\ell$ is a single sparse block-matrix multiply that simultaneously realizes all three channels (exact, coexact, Laplacian) at every rank.

\begin{algorithm}[h]
\caption{Residual TNO layer (per layer $\ell$, on lifted complex $\widetilde{K}$)}
\label{alg:tno-layer}
\KwIn{%
  Cochain features $\mathbf{H}_\ell = (H^0_\ell, \ldots, H^N_\ell)$;
  DEC operators $\{d^{k-1}\}, \{\delta^{k+1}\}, \{\Delta_k^\uparrow\}, \{\Delta_k^\downarrow\}$ of $\widetilde{K}$;
  channel-mixing matrices $\{W^\downarrow_k, W^\uparrow_k, W^{\Delta,\uparrow}_k, W^{\Delta,\downarrow}_k\}$;
  feature transformation $\phi_\theta$%
}
\KwOut{Updated features $\mathbf{H}_{\ell+1} = (H^0_{\ell+1}, \ldots, H^N_{\ell+1})$}
\For{$k = 0, 1, \ldots, N$}{
  $m_k^{\downarrow} \gets d^{k-1}H^{k-1}_\ell\,W^\downarrow_k$
    \tcp*{exact channel}
  $m_k^{\uparrow} \gets \delta^{k+1}H^{k+1}_\ell\,W^\uparrow_k$
    \tcp*{coexact channel}
  $m_k^{\Delta} \gets \Delta_k^\uparrow H^k_\ell\,W^{\Delta,\uparrow}_k + \Delta_k^\downarrow H^k_\ell\,W^{\Delta,\downarrow}_k$
    \tcp*{Hodge--Laplacian channel}
  $H^k_{\ell+1} \gets H^k_\ell + \phi_\theta\!\bigl(H^k_\ell,\, m_k^{\downarrow},\, m_k^{\uparrow},\, m_k^{\Delta}\bigr)$
    \tcp*{cf.~\cref{eq:tl-single}}
}
\Return $\mathbf{H}_{\ell+1}$
\end{algorithm}

\subsection{HTNO as a learned two-grid method}
\label{app:htno-twogrid}

HTNO adds a coarse complex $K^c$ and degree-wise transfer maps
\[
    R_k : C^k(K) \to C^k(K^c),
    \qquad
    \Pi_k : C^k(K^c) \to C^k(K).
\]
For de~Rham systems, the ideal structure-preserving condition is that
transfer commute with the coboundary:
\[
    d^k_{K^c} R_k = R_{k+1} d^k_K,
    \qquad
    d^k_K \Pi_k = \Pi_{k+1} d^k_{K^c}.
\]
This is the standard commuting-diagram condition in FEEC multigrid
~\citep{arnold2000multigrid,arnold2006finite,arnold2018feec}.
When these identities hold, taking a coboundary and transferring levels
agrees with transferring first and then taking the coboundary. In
particular, exact cochains remain exact across levels, and the discrete
differential structure is preserved by restriction and prolongation.
Without such compatibility, a coarse correction can turn a gradient-like
error into a non-gradient error after prolongation, weakening the
decomposition on which Hodge-type smoothers rely
~\citep{hiptmair1998multigrid,arnold2000multigrid}.

In the present HTNO implementation, these commuting identities are not
imposed as hard algebraic constraints. Rather, they serve as the FEEC
ideal motivating the hierarchy. The TNO blocks on each level use the DEC
operators of that level, so typed cochain transport is preserved
level-wise. The inter-level maps used in our experiments are induced by a
partition of the fine vertices and should therefore be viewed as
cochain-compatible coarse correction maps, not as exact commuting-diagram
transfers.

\begin{figure}[t!]
  \centering
  \input{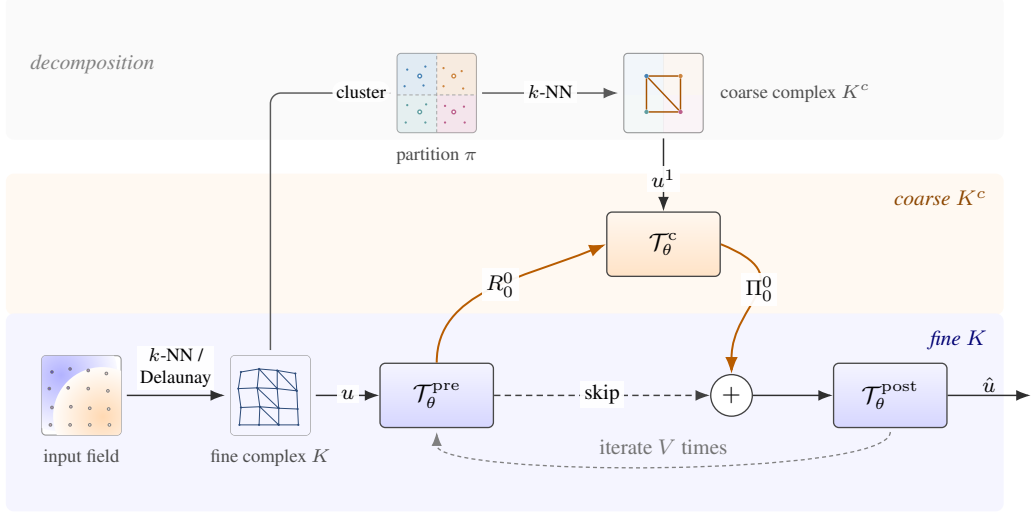}
    \caption{HTNO as a learned two-grid V-cycle. \textbf{Bottom:} the input
    field on a point cloud is lifted to a fine cell complex $K$ ($k$-NN or
    Delaunay), giving an input cochain $u \in C^{\bullet}(K)$ that drives
    the V-cycle. The pre-smoother $\mathcal{T}^{\mathrm{pre}}_\theta$ runs
    on $K$; its output is restricted by $R^{0}_{0}$ to the coarse complex,
    where $\mathcal{T}^{\mathrm{c}}_\theta$ computes a correction; the
    correction is prolonged back by $\Pi^{0}_{0}$ and added to the
    pre-smoothed state via the skip; $\mathcal{T}^{\mathrm{post}}_\theta$
    closes the cycle, producing $\hat{u}$.
    The whole block can be iterated $V$ times with shared weights.
    \textbf{Top:} the coarse complex is built once from $K$. Fine vertices
    are clustered (fixed $k$-means or learned soft Voronoi) into a partition
    $\pi$, and $k$-NN on the resulting centroids yields the coarse complex
    $K^{c}$, on which the coarse cochain $u^{c}$ lives.\vspace{-3mm}}
    \label{fig:htno-vcycle}
\end{figure}

With this interpretation, an HTNO block is a learned two-grid V-cycle
(\cref{fig:htno-vcycle})~\citep{trottenberg2000multigrid,
briggs2000multigrid} that pre-smooths on $K$ with Hodge-structured TNO
layers~\citep{hiptmair1998multigrid,hiptmair2002finite,arnold2000multigrid},
restricts to $K^c$, computes an approximate correction on $K^c$ with
coarse-level TNO layers, prolongs the correction back to $K$, additively
adds it to the pre-smoothed state, and post-smooths on $K$. With $L=2$
this reads
\begin{equation}
\mathrm{HTNO}_\theta(h)
\;=\;
\mathcal T^{\mathrm{post}}_\theta\!\Bigl(\,
\mathcal T^{\mathrm{pre}}_\theta(h)
\;+\;
\Pi^{0}_{0}\,\mathcal T^{\mathrm{c}}_\theta\!\bigl(
  R^{0}_{0}\,\mathcal T^{\mathrm{pre}}_\theta(h)\bigr)
\,\Bigr).
\label{eq:htno-block}
\end{equation}
Our implementation realizes $R^0_0,\Pi^0_0$ as mean-pool and broadcast over a partition $\pi:V(K)\!\to\!\{1,\dots,K_c\}$ of the fine vertices.
The partition is either \emph{fixed} ($k$-means on input coordinates, SLIC superpixels, or an MS-complex segmentation), or \emph{learned end-to-end} as a soft Voronoi diagram: $K_c$ trainable query vectors attend over the warm-up features to predict centroid positions, soft assignments are obtained from $W=\mathrm{softmax}(-\|p-c\|^2/T)$ (or a Gaussian RBF kernel), and auxiliary spread and entropy losses on $W$ keep the centroids from collapsing. 
The coarse complex $K^c$ is built by $k$-NN over the centroids, inheriting its own incidence matrices and Hodge stars.

Higher-rank transfers ($k\!\ge\!1$) would require structured prolongation and restriction maps, for example Whitney/N\'ed\'elec-style constructions, to satisfy the commuting identities exactly ~\citep{nedelec1980mixed,bossavit1998computational,arnold2006finite}.
We do not impose such maps in the present implementation. 
Instead, we re-derive rank-$1$ and rank-$2$ features on $K^c$ from the pooled vertex field using the coarse incidence structure. 
Thus $\mathcal T^{\mathrm{c}}_\theta$ runs the multi-rank TNO update of \cref{eq:tno-linear} on $K^c$ when higher-rank features are present, and a rank-$0$ message-passing block otherwise. 
The fine--coarse--fine block used in our experiments is the $V\!=\!1$ instance of \cref{eq:htno-block}; iterating with shared weights for $V\!>\!1$, or recursing on deeper hierarchies, gives learned multilevel V-cycles.

\subsection{Scope of the analogy}
The correspondence with Hodge-compatible smoothing is structural, not a multigrid convergence theorem. Classical multigrid convergence depends on a specified residual equation, smoother, transfer operators, and regularity assumptions~\citep{trottenberg2000multigrid,briggs2000multigrid,
bramble1991analysis}. 
A TNO layer does not inherit these rates
automatically. 
What is shared is the algebraic form: cross-degree transport is fixed by $d^{k-1}$ and $\delta^{k+1}$, harmonic components are accessed through $P_k^{\mathrm{harm}}$, and inter-level maps can be chosen to commute with $d$~\citep{arnold2000multigrid,arnold2006finite}.
Training then learns channel mixing, nonlinear correction, and coarse representations inside this fixed de~Rham structure.

This distinction is also what separates the construction from point-only neural operators~\citep{kovachki2023neural,li2021fourier}. 
A point-only model may use pooling, attention, U-Net skips, or learned coarse variables, but it does not by default carry typed cochain spaces $C^k$, fixed maps $d^k:C^k\to C^{k+1}$, codifferentials $\delta^{k+1}$, or
harmonic projectors. 
TNOs make these objects primitive operators of the architecture rather than behaviors to be inferred from data.

\section{Implementation Details}
\label{app:implementation_details}

\subsection{MPNN ablation baseline}
\label{app:mpnn-baseline}

The MPNN used in the ablations of~\cref{sec:ablations} is a standard residual message-passing network on rank-$0$ (vertex) features only, sharing the Delaunay graph and the node/edge inputs of \name{} but with no sheaf, harmonic, or higher-rank channels. Inputs $x_v\!\in\!\mathbb{R}^{F_{\mathrm{in}}}$ and edge features $e_{ij}\!\in\!\mathbb{R}^{F_e}$ are lifted to width $w$ by a $\mathrm{Dense}\!\to\!\mathrm{swish}\!\to\!\mathrm{LayerNorm}$ encoder applied independently on vertices and edges. Each of the $L{=}12$ residual layers, with $\bar h = \mathrm{LN}(h)$, computes
\begin{equation}
m_{ij} = \mathrm{swish}\!\bigl(W_m\,[\,\bar h_i \,\Vert\, \bar h_j \,\Vert\, e_{ij}\,]\bigr),\qquad
a_i = \!\!\sum_{j:\,j\to i}\!\! m_{ij},
\label{eq:mpnn-msg}
\end{equation}
\begin{equation}
h_i \;\leftarrow\; h_i \;+\; \mathrm{LN}\!\Bigl(\mathrm{swish}\!\bigl(W_2\,\mathrm{swish}(W_1\,[\,\bar h_i \,\Vert\, a_i\,])\bigr)\Bigr),
\label{eq:mpnn-update}
\end{equation}
with sum aggregation on the destination index. A final $\mathrm{LN}\!\to\!\mathrm{Dense}$ decoder produces the vertex output. The matched-width baseline uses $w{=}192$, the parameter-matched baseline uses $w{=}272$ (cf.~\cref{app:ablation-details}); all other optimizer/training settings are inherited from the ablation backbone.

\subsection{Complexity of a TNO layer}
\label{app:tno-layer-complexity}

Let $n_k=|\widetilde K_k|$ be the number of rank-$k$ cells,
$n_\bullet=\sum_k n_k$, and $h=d_h$. Let
\[
  s_\bullet=\sum_{k=1}^{N}\operatorname{nnz}(B_k),
  \qquad
  \lambda_\bullet
  =
  \sum_{k=0}^{N}
  \Bigl(
  \operatorname{nnz}(\Delta_k^\uparrow)
  +
  \operatorname{nnz}(\Delta_k^\downarrow)
  \Bigr),
\]
with ''nnz`` the non-zero counts for the sparse operators at runtime. 
A residual TNO layer has two costs. First, the DEC transports $d^{k-1}$, $\delta^{k+1}$, and $\Delta_k^{\uparrow,\downarrow}$ are sparse matrix--dense matrix multiplies, costing $O(h(s_\bullet+\lambda_\bullet))$.
Second, the learned maps are shared channel mixes applied cellwise, costing $O(n_\bullet h^2)$. Thus
\begin{equation}
  T_{\mathrm{TNO}}^{(1)}
  =
  O\!\left(h(s_\bullet+\lambda_\bullet)+n_\bullet h^2\right).
  \label{eq:tno-layer-complexity}
\end{equation}
For bounded-degree cell complexes, $s_\bullet+\lambda_\bullet=O(n_\bullet)=O(n_0)$, so a \textit{rigid} DEC TNO layer is $O(n_0 h^2)$.

For the MPNN in \cref{eq:mpnn-msg,eq:mpnn-update}, let $n_0$ be the number of vertices and $m$ the number of directed edges; edge features are lifted to width $w$ by the encoder, so they enter the layer at width $w$.
The edge message $W_m$ costs $O(mw^2)$, aggregation costs $O(mw)$, and the two-layer node update costs $O(n_0 w^2)$. Hence
\begin{equation}
  T_{\mathrm{MPNN}}^{(1)}
  =
  O\!\left((m+n_0)w^2\right),
  \label{eq:mpnn-layer-complexity}
\end{equation}
which is $O(n_0 w^2)$ under bounded degree.

Both layers are quadratic in their channel width: $O(n_0 w^2)$ for MPNN and $O(n_0 h^2)$ for rigid DEC TNO, but the cost is accrued in different places.
An MPNN spends $O(w^2)$ \emph{per edge} in the message MLP; a rigid DEC TNO spends only $O(h)$ per incidence in transport and $O(h^2)$ \emph{per cell} across all ranks in channel mixing.
The extra factor $n_\bullet$ instead of $n_0$ is the cost of carrying edge, face, and higher-rank cochains as native state variables, which is what prescribes typed cross-rank routing.
If instead the copresheaf realization is used with incidence-specific dense fiber maps $\rho_{y\to x}\!\in\!\mathbb{R}^{h\times h}$ (as in the CopresheafGIN variant of~\cite{hajij2025copresheaf}; their GCN/Sage variants restrict $\rho_{y\to x}$ to be diagonal, retaining the $O(s_\bullet h)$ cost), transport raises to $O(s_\bullet h^2)$, matching the per-edge $O(w^2)$ scaling of MPNN message MLPs at common width.

\subsection{Official Baseline Implementations: RIGNO, GAOT, HOGNN}
\label{app:author-impls}

For RIGNO~\cite{mousavi2025rigno}, GAOT~\cite{hao2024gaot}, and HOGNN~\cite{liao2025boundaryvalue}, we use the authors' official implementations. 
These are adopted from published optimizer settings; we tune only the epoch budget and per-dataset width on the validation split; test numbers come from the best-validation checkpoint consistent with our models' evaluation protocol.

\textbf{GAOT~\cite{hao2024gaot}} is evaluated on both the RIGNO~\cref{tab:rigno-steady-state} and GAOT~\cref{tab:gaot-suite} benchmarks as it is the strongest published unstructured-mesh operator outside the methods evaluated by RIGNO \cref{tab:rigno-steady-state}, and it shares the same problem setting.
We follow the upstream optimizer (AdamW with cosine-decay, peak $10^{-3}$, weight decay $10^{-5}$, gradient clipping $1.0$) and its effective batch of $64$ (data-parallel over four GH200).
We extend the epoch budget to $2{,}000$ with patience $300$ (the upstream configurations use $100$ and $1{,}000$ for PG and Elasticity, respectively); due to best-validation checkpoint selection, this remains a safety margin to assure convergence.

We note that the public GAOT release does not match the paper's Table~1 numbers on the GAOT suite itself (e.g.\ NACA0012 $12.07\%$ vs.\ paper $6.81\%$); this has been observed in issue \#1 of the public repository that not all results are reproducible: for instance, they use 2048 train samples instead of 1024 (RIGNO / our protocol) on Poisson Gauss.
GAOT entries are therefore a head-to-head comparison under identical experimental conditions, not as a reproduction of the paper's reported results.

\begin{figure}[t]
  \centering
  \includegraphics[width=\linewidth]{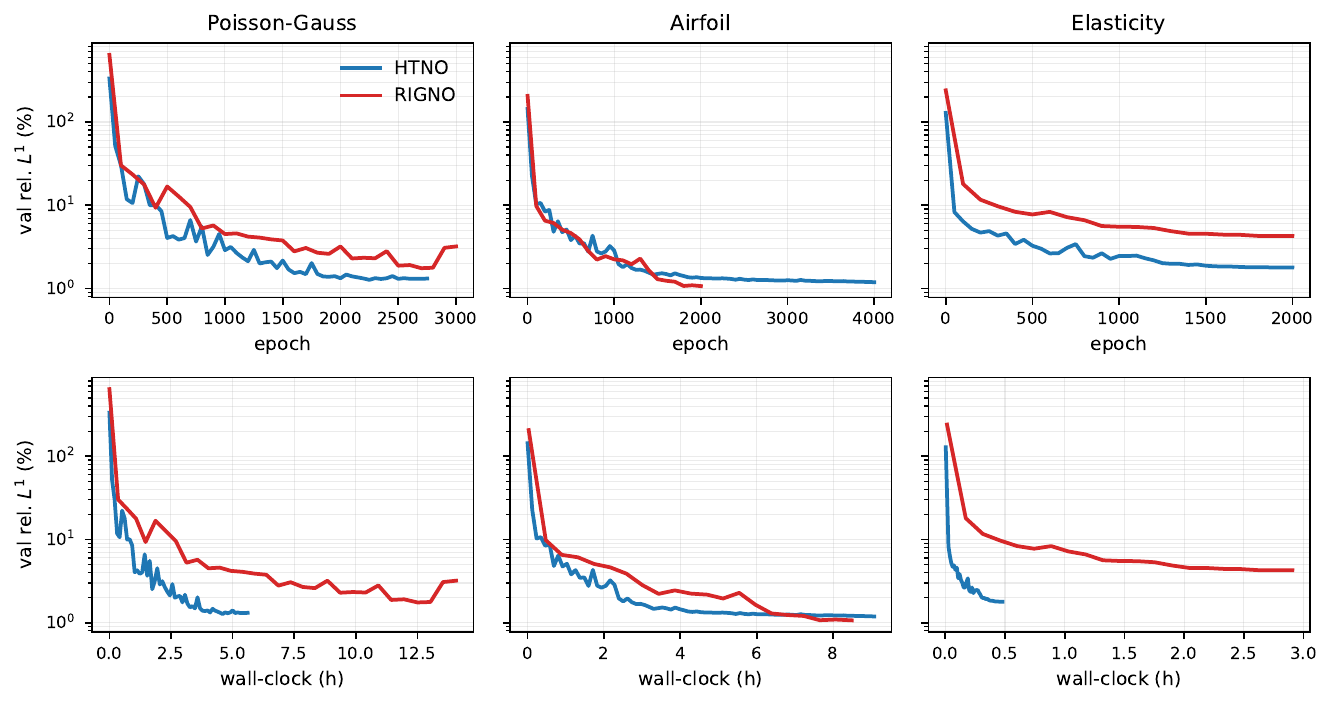}
  \caption{Validation relative $L^1$ vs.\ training epochs (left column) and cumulative wall-clock hours (right column) for HTNO and RIGNO on the three RIGNO steady-state benchmarks. Both models were trained under their respective official optimizer/batch settings; final test numbers correspond to the entries in~\cref{tab:rigno-steady-state}.}
  \label{fig:htnn-vs-rigno-convergence}
  \vspace{-4mm}
\end{figure}

\paragraph{RIGNO~\cite{mousavi2025rigno} convergence observations}
In contrast to the RIGNO multi-level physical/regional graph, HTNO arranges TNO blocks as a $V$-cycle over a sequence of coarsened cell complexes $K_0 \leftarrow \cdots \leftarrow K_L$ connected by degree-preserving restriction and prolongation maps $R^k_\ell,\Pi^k_\ell$ (\cref{sec:tno_main}): fine-level blocks capture short-range cross-degree physics, while coarse-level blocks, acting on the same cochain types after transfer, propagate long-range information.
In addition to a superior performance across numerous benchmarks (see \cref{sec:experiments}), we observe notably faster convergence.
\Cref{fig:htnn-vs-rigno-convergence} depicts two qualitative observations from training: on these three benchmarks, HTNO reaches lower validation error in notably fewer epochs and in less wall-clock time than RIGNO, except Airfoil, where the two are comparable.
Field-level prediction comparisons for these benchmarks are given in \cref{app:qualitative-steady-state}.

\textbf{HOGNN~\cite{liao2025boundaryvalue}} is a closely related baseline, in a spirit that shares motivations with \name{}. 
Both lift signals beyond the vertices using DEC operators, providing an informative head-to-head comparison.
We run it on every 2D benchmark with a matching \name{} or HTNO entry: PG, AF, Elasticity, NACA0012/2412, RAE2822, PCS, and the synthetic-topology family (\cref{sec:ablations}).
We use the upstream optimizer (AdamW with cosine-decay, peak $5{\times}10^{-4}$, weight decay $10^{-4}$, gradient clipping $1.0$), $300$ epochs, patience $200$, and the identical Delaunay graph construction as for TNO, HTNO, and RIGNO; per-dataset tuning is restricted to hidden width.
EmmiWing is excluded because the public release is 2D only, and porting its DEC operator stack to 3D surfaces is outside our scope.

\begin{table}[h]
\vspace{-6mm}
  \centering
  \caption{HOGNN reproduction across all 2D benchmarks: test relative $L^1$~(\%); lower is better, \textbf{bold}: best per column. ``Best of ours'' is the smaller of \name{} and HTNO from the cited table.}
  \label{tab:hognn-comparison}
  \small
  \setlength{\tabcolsep}{4pt}
  \resizebox{\linewidth}{!}{%
  \begin{tabular}{@{}l cccccccccccc@{}}
    \toprule
    & \multicolumn{3}{c}{\cref{tab:rigno-steady-state}}
    & \multicolumn{4}{c}{\cref{tab:gaot-suite}}
    & \multicolumn{2}{c}{\cref{tab:rankin-aniso}}
    & \multicolumn{2}{c}{\cref{tab:ablation-v19}} \\
    \cmidrule(lr){2-4}\cmidrule(lr){5-8}\cmidrule(lr){9-10}\cmidrule(lr){11-12}
    Method
      & \rotatebox{60}{Poisson-Gauss}
      & \rotatebox{60}{Airfoil}
      & \rotatebox{60}{Elasticity}
      & \rotatebox{60}{NACA0012}
      & \rotatebox{60}{NACA2412}
      & \rotatebox{60}{RAE2822}
      & \rotatebox{60}{PCS}
      & \rotatebox{60}{projected}
      & \rotatebox{60}{native rank-$2$}
      & \rotatebox{60}{Darcy}
      & \rotatebox{60}{Adv.-Diff.} \\
    \midrule
    Best of ours
      & \textbf{1.03} & \textbf{1.11} & \textbf{1.70}
      & \textbf{3.77} & \textbf{3.92} & \textbf{3.92} & \textbf{0.52}
      & \textbf{5.42} & \textbf{4.88}
      & \textbf{4.87} & \textbf{5.20} \\
    HOGNN
      & 33.78 & 15.64 & 21.24
      &  9.36 & 10.33 &  9.28 & 50.44
      & 14.84 & 15.97
      & 12.04 & 13.23 \\
    \bottomrule
  \end{tabular}%
  }
\end{table}

\textbf{Results.}
\Cref{tab:hognn-comparison} lists test relative $L^1$ error for HOGNN on every benchmark with a matching \name{} or HTNO entry. 
HOGNN does not match the leading method in any of the cells evaluated; we therefore consider it not well-suited for these domains, as the authors primarily study applications to electrostatics.

\section{Experimental Details}
\label{app:experimental-details}

We provide additional implementation details for our models, and baselines. 
\name{} and HTNO are implemented in JAX~\cite{jax2018github}. 
We use the author's official implementations (see \cref{app:implementation_details}),
which are often run in PyTorch~\cite{paszke2019pytorch}; we configure these with the same dataloaders our JAX models use to guarantee consistency.
Implementations will be published upon acceptance.
All models are run on a cluster of nodes consisting of 4 NVIDIA GH200s (96GB VRAM).
The maximum compute budget on larger datasets (Emmiwing, PCS) is 4 GPUs simultaneously. 
Other steady-state benchmarks are run with 1 GPU per model.

\subsection{Poisson-Gauss, Airfoil, Elasticity}
\label{app:steady-state-rigno}

\paragraph{Datasets}
We use three datasets from the RIGNO benchmark~\cite{mousavi2025rigno}:
\begin{enumerate}[nosep,leftmargin=*]
    \item \textbf{Poisson-Gauss}~(PG, Poisson with Gaussian sources on a random point cloud, $9{,}216$ nodes, $1024/128/256$ train/val/test, fixed positions);
    \item \textbf{Airfoil Flow}~(AF, compressible Euler flow with varying airfoil geometry at fixed far-field $M_\infty{=}0.8$, $\alpha{=}0^\circ$~\cite[\S4.4]{li2023fourier}, $200{\times}50$ C-grid of quadrilateral elements (${\sim}11$K nodes) redistributed by~\cite{mousavi2025rigno} with $2048/128/256$ train/val/test, variable positions);
    \item \textbf{Elasticity}~(hyper-elastic deformation of varying domains, ${\sim}1$K nodes, $1024/256/512$, variable).
We note that AF is, on closer inspection, an effectively single-condition transonic benchmark; we audit its regime spread against the GAOT airfoils in \cref{app:af-vs-gaot-regime}.
\end{enumerate}

\paragraph{Baselines}
This section contains additional details regarding baselines and training protocols.
We evaluate RIGNO~\cite{mousavi2025rigno}, GAOT~\cite{hao2024gaot} and HOGNN~\cite{liao2025boundaryvalue} on the same splits using the author's official implementations (\cref{app:author-impls}).

\paragraph{Training protocol}
For \name{}, we report the best result from per-dataset tuning around a harmonic baseline (12 layers, hidden $128$, $r{=}8$ harmonic modes), varying depth, width, learning rate, neighborhood size, and Delaunay graph construction.
HTNO uses dynamic $k$-means clustering ($K{=}32$--$128$ regions), 18 layers ($6{+}6{+}6$ fine/coarse/fine), hidden $128$--$256$ depending on dataset scale.
Training uses AdamW with cosine-decay (peak $5{\times}10^{-4}$), gradient clipping at $1.0$, and early stopping with patience $200$--$500$.
Models train for $2{,}000$--$4{,}000$ depending on convergence.
The same protocol applies to the other benchmarks below.

\subsection{Compressible Airfoils and Poisson-with-sines}
\label{app:steady-state-gaot}

\paragraph{Datasets}
From the GAOT release~\cite{hao2024gaot} we use NACA0012/2412 and RAE2822 (${\sim}8$K-node airfoils, spanning subsonic, transonic, and supersonic regimes) and Poisson-C-Sines~(PCS, fixed $16{,}431$-node mesh).

\paragraph{Baselines}
HTNO and \name{} are trained as in the \cref{app:steady-state-rigno} protocol; RIGNO and GAOT are evaluated on identical splits using the authors' official implementations (\cref{app:author-impls}).

\paragraph{Splits}
\label{app:gaot-splits}
The GAOT public release contains per-dataset NetCDF files but does not publish the train/val/test indices, making it challenging to reproduce Table~1. 
We thus reconstruct splits based on the sample counts declared in the paper.
We follow the upstream data pipeline's default contiguous-block convention,
\[
\mathrm{train} = [0,\,N_\mathrm{tr}),\quad
\mathrm{val}   = [N_\mathrm{tr},\,N_\mathrm{tr}+N_\mathrm{val}),\quad
\mathrm{test}  = [N_\mathrm{tot}-N_\mathrm{te},\,N_\mathrm{tot}),
\]
i.e.\ the test split is the last $N_\mathrm{te}$ samples of the source file, with a gap between val and test on the larger airfoil files. Counts per dataset are listed in \cref{tab:gaot-splits}.

\begin{table}[h]
\centering
\caption{Reconstructed GAOT-suite splits. $N_\mathrm{tot}$ is the sample count of the public release; the $5000/128/256$ airfoil train/val/test split matches~\cite[\S D.2.1, p.36]{hao2024gaot}, drawn from a pool of $5384$ samples per airfoil; PCS uses the maximal contiguous train slice the file admits.}
\label{tab:gaot-splits}
\small
\begin{tabular}{@{}l rrrr@{}}
\toprule
Dataset & $N_\mathrm{tot}$ & $N_\mathrm{tr}$ & $N_\mathrm{val}$ & $N_\mathrm{te}$ \\
\midrule
NACA0012        & 43{,}490 & 5000 & 128 & 256 \\
NACA2412        & 47{,}925 & 5000 & 128 & 256 \\
RAE2822         & 48{,}375 & 5000 & 128 & 256 \\
Poisson-C-Sines &  5{,}000 & 4616 & 128 & 256 \\
\bottomrule
\end{tabular}
\end{table}

\paragraph{Bluff-Body}
The Bluff-Body benchmark from the same release is omitted: the public collection contains only $7$ of the $10$ paper pretraining shapes (Square, Cone, and Rectangle-L are absent, and the paper's $5$-Ellipse pretraining set is replaced in the release by Ellipse-1/2 plus other shapes designated in the paper itself as fine-tuning geometries); we thus omit this evaluation due to incompleteness.

\subsection{Large-Scale Surface PDEs: EmmiWing}
\label{app:emmiwing-details}

\paragraph{Dataset and resolution protocol}
EmmiWing~\cite{paischer2025going} provides $29{,}727$ steady-state compressible-RANS wing surfaces with raw meshes of $244$K--$426$K nodes (variable across geometries). We use the official convex-hull-peeling split (ID-random, OOD-extrapolation, OOD-interpolation; $25{,}674/999/2{,}992$ train/val/test, with $62$ erroneous cases excluded). 
Following AB-UPT~\cite{alkin2025abupt}, we adopt a $65{,}536$/$16{,}384$-point sampling protocol: each raw mesh is once FPS-resampled to $65$K shared query points, and a fresh $16{,}384$-node subset is drawn each training step. 
This follows established conventions in the literature: in ~\cite{paischer2025going}, (PointNet, Transolver, ViT) also train at $16{,}384$ surface points; AB-UPT~\cite{alkin2025abupt} uses the $65{,}536{\to}16{,}384$ supernode/anchor pattern that we follow most closely.

\paragraph{Deviations from the published protocol}
We evaluate on the $65$K FPS subset, where~\cite{paischer2025going,alkin2025abupt} evaluate on the full $244$K--$426$K raw mesh via chunked inference. 
Aggregate-$L^1$ numbers are therefore not directly comparable to the published Table~2 of~\cite{paischer2025going}; the comparisons in \cref{tab:emmiwing} are between our four models trained and evaluated under the same regime. 
Closing this gap would require equipping the topological backbone with a neural-field-style decoder that decouples prediction resolution from the fixed shared mesh, which we leave as future work.

\paragraph{Training protocol}
In addition to HTNO and RIGNO-18, we report Transolver~\cite{wu2024transolver} (the slice-attention transformer designed for unstructured meshes), a pre-norm self-attention Transformer~\cite{vaswani2017attention} with sinusoidal coordinate-based positional encoding, and a PointNet-style baseline~\cite{qi2017pointnet} (per-point MLPs with global max-pool, implemented as pre-norm residual blocks).
All models are trained under matched compute budgets with AdamW, peak learning rate $5{\times}10^{-4}$ on a cosine-decay schedule, gradient clipping at $1.0$, and early stopping on validation $L^1$.

\paragraph{Per-channel results}
Pressure is recovered to high fidelity by every model ($\leq 0.30\%$); the wall-shear-stress channels dominate the spread. HTNO leads on every channel; Transolver is the consistent runner-up, $0.2$--$0.3$~pp behind on each $\tau$ component (see \cref{tab:emmiwing}).

\subsection{Anisotropic Darcy with Per-Face Random Tensor Orientation}
\label{app:rankin-aniso-details}

This section details the simulation protocol, dataset construction, and full result matrix for the rank-input study summarized in~\cref{sec:rankin-aniso}.

\paragraph{Synthetic-topology benchmark}
The dataset family used in~\cref{sec:ablations} is a 2D planar mesh distribution with a square outer boundary and zero, one, or two interior holes (${\sim}1{,}000$ nodes per mesh) under standard boundary jitter and adaptive mesh-density variation.
The data layout is $100$ unique meshes $\times$ $50$ samples per mesh (each dataset has ($4{,}000/500/500$) train/val/test splits).

\paragraph{Anisotropic Darcy with face-valued $\kappa$-tensor}
We solve
\[
  -\nabla\!\cdot\!\big(\kappa(x)\,\nabla u\big) = f,\qquad u|_{\partial\Omega_{\mathrm{outer}}}=0,\quad u|_{\partial\Omega_{\mathrm{hole},k}}=g_k\!\sim\!\mathcal{U}[0.5,1.5],
\]
with $\kappa(x)$ a DG0 (piecewise-constant per triangle) symmetric $2{\times}2$ tensor field reconstructed from a per-face orientation angle drawn iid per triangle, $\phi_t \!\sim\! \mathcal{U}[0,\pi)$:
\[
  \kappa_{\mathrm{face}}[t] \;=\; R(\phi_t)\,\mathrm{diag}(\kappa_\parallel,\kappa_\perp)\,R(\phi_t)^{\!\top},\qquad \kappa_\parallel{=}4,\ \kappa_\perp{=}1\ \text{(anisotropy ratio $4$).}
\]
Concretely $\kappa_{xx}[t]{=}\kappa_\perp+(\kappa_\parallel{-}\kappa_\perp)\cos^2\phi_t$, $\kappa_{yy}[t]{=}\kappa_\perp+(\kappa_\parallel{-}\kappa_\perp)\sin^2\phi_t$, $\kappa_{xy}[t]{=}(\kappa_\parallel{-}\kappa_\perp)\cos\phi_t\sin\phi_t$.
The forcing $f$ is a superposition of $1$--$3$ signed Gaussian bumps with amplitudes $a_i\!\sim\!\mathcal{U}[1.0,2.5]$, $\sigma{=}0.25$, and the same mesh-adaptive minimum-width constraint used in \cref{sec:ablations}.
The variational form is assembled in FEniCSx \cite{baratta2024dolfinx} with $P_1$ Lagrange for $u$ and DG0 for the three $\kappa$-tensor channels; the bilinear form expands the inner product directly to avoid UFL rank-mismatch on the asymmetric off-diagonal term.

\paragraph{Significance}
The face channel $\cos(2\phi_t)$ has $\mathrm{Var}[\cos(2\phi_t)]{=}\tfrac12$ per face under $\phi_t\!\sim\!\mathcal{U}[0,\pi)$, but population mean $\mathbb{E}[\cos(2\,\mathcal{U}[0,\pi))]{=}0$. 
The vertex-mean projection (incident-face average over $5$--$6$ neighbors) therefore concentrates further toward $0$ at every vertex: the projected channel is near-constant noise. 
Empirically the face channel has $\mathrm{mean}{=}{-}0.003$, $\mathrm{std}{=}0.696$, matching the analytic prediction. 
\textbf{Vertex projection is therefore lossy by construction}; the face orientation field $\cos(2\phi_t)$ cannot be reconstructed from any vertex-supported representation.

\paragraph{Model input}
Models receive $c{=}5$ vertex channels $[f, \mathbb{1}_{\partial}, g, \kappa_\parallel, \kappa_\perp]$ (the last two are global scalars carried to every vertex), an edge channel $c_e\!\in\!\mathbb{R}^{|E|}$ (face-incident mean of $\cos(2\phi_t)$), and a face channel $c_f\!\in\!\mathbb{R}^{|F|}$ ($\cos(2\phi_t)$ per triangle). 
For \name{}s, $c_e$ is routed through the rank-$1$ cochain channel and $c_f$ through the rank-$2$ channel. 
The \emph{projected} experiments instead replace the rank-aux channels with their vertex-mean projection appended as channels $6$--$7$ of the vertex-$c$ feature, with native rank-aux ingestion disabled.

\paragraph{Training protocol}
All cells share one backbone: standard \name{} with sheaf transport, harmonic basis, hidden $192$, dropout $0.20$, AdamW with cosine-decay (peak $5{\times}10^{-4}\!\to\!10^{-5}$, weight decay $10^{-4}$), gradient clipping $1.0$, batch size $16$, $300$ epochs. 
The two MPNN baselines do not have sheaf nor harmonic components; for parameter-matched experiments, MPNN uses a width $w{=}272$ to reach ${\sim}5.35$M parameters (vs.\ ${\sim}5.0$M for \name{}).

\subsection{Ablations}
\label{app:ablation-details}

We instantiate two scalar PDEs whose flux carries non-trivial curl on this distribution: Darcy ($-\nabla\!\cdot\!(\kappa\nabla u) + \sigma u = f$ with anisotropic $\kappa$), and conservative advection--diffusion, both assembled in FEniCSx~\cite{baratta2024dolfinx} with the same variational forms used in the rank-input study.
Within each mesh block, only the per-hole Dirichlet BC values vary across the $50$ samples (drawn $\sim \mathcal{U}[0.5, 1.5]$); the source field, diffusivity tensor, and advection scale are held fixed, so topology is controlled for.

\paragraph{Backbone}
All ablation variants share one backbone: hidden dim $192$, MLP messages, $r_{\max}{=}1$, dropout $0.20$, AdamW with cosine-decay (peak $5{\times}10^{-4}$, weight decay $10^{-4}$), gradient clipping at $1.0$, batch size $16$, $300$ epochs, rigid 2D augmentation enabled, single GH200.
Variants differ only in the sheaf-transport and the harmonic-basis input.
The two MPNN baselines use the same backbone with sheaf and harmonic disabled; the param-matched MPNN uses width $w{=}272$ to reach ${\sim}5.35$M parameters (vs.\ $5.0$M for \name{})

\section{Extended Experimental Results}
\label{app:extended-results}
Next, we collect qualitative field-level comparisons for the steady-state benchmarks of \cref{sec:experiments} and a per-dataset regime audit that explains why the \name{}/RIGNO ranking inverts between airfoil families.

\begin{figure}[t]
  \centering
  \includegraphics[width=\linewidth]{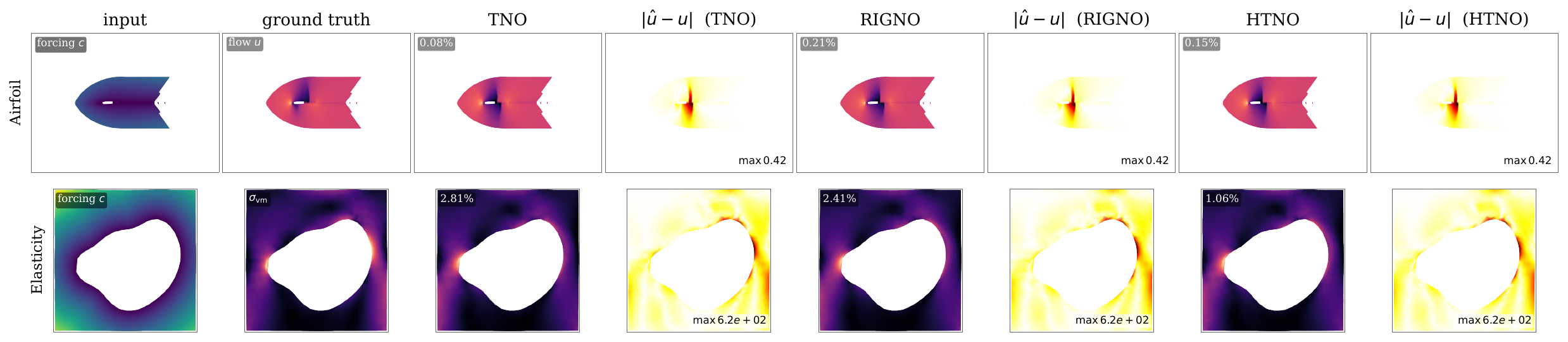}
  \caption{Qualitative comparison on the RIGNO-suite steady-state benchmarks (Airfoil, Elasticity). Each row shows one median-test-$L^1$ sample for the dataset; columns: input forcing, ground-truth field, and per-architecture prediction $\hat u$ followed by absolute error $|\hat u - u|$ for \name{}, RIGNO, and HTNO. The three error panels in a row share a common color scale, with the shared $\max$ stamped on each panel so residual magnitudes are directly comparable across architectures.\vspace{-4mm}}
  \label{fig:rigno-suite-qualitative}
\end{figure}

\subsection{Qualitative comparisons on the steady-state suites}
\label{app:qualitative-steady-state}

\paragraph{RIGNO-suite (Airfoil, Elasticity)}
\Cref{fig:rigno-suite-qualitative} shows field-level predictions on a median-test-$L^1$ sample of each dataset for \name{}, RIGNO, and HTNO.
On AF, all three architectures recover the ground-truth density field to within ${\sim}1$--$3\%$ relative $L^1$, with residuals concentrated near the airfoil tip and along the wake; on Elasticity, the residuals localize on the loaded notch boundary.
The visual closeness of the three error maps on AF mirrors the table-level finding (\cref{tab:rigno-steady-state}) that \name{} and RIGNO-18 are largely indistinguishable on this benchmark. 
This can be attributed to the dataset's simple regime (see \cref{app:af-vs-gaot-regime}).

\begin{figure}[t]
  \centering
  \includegraphics[width=\linewidth]{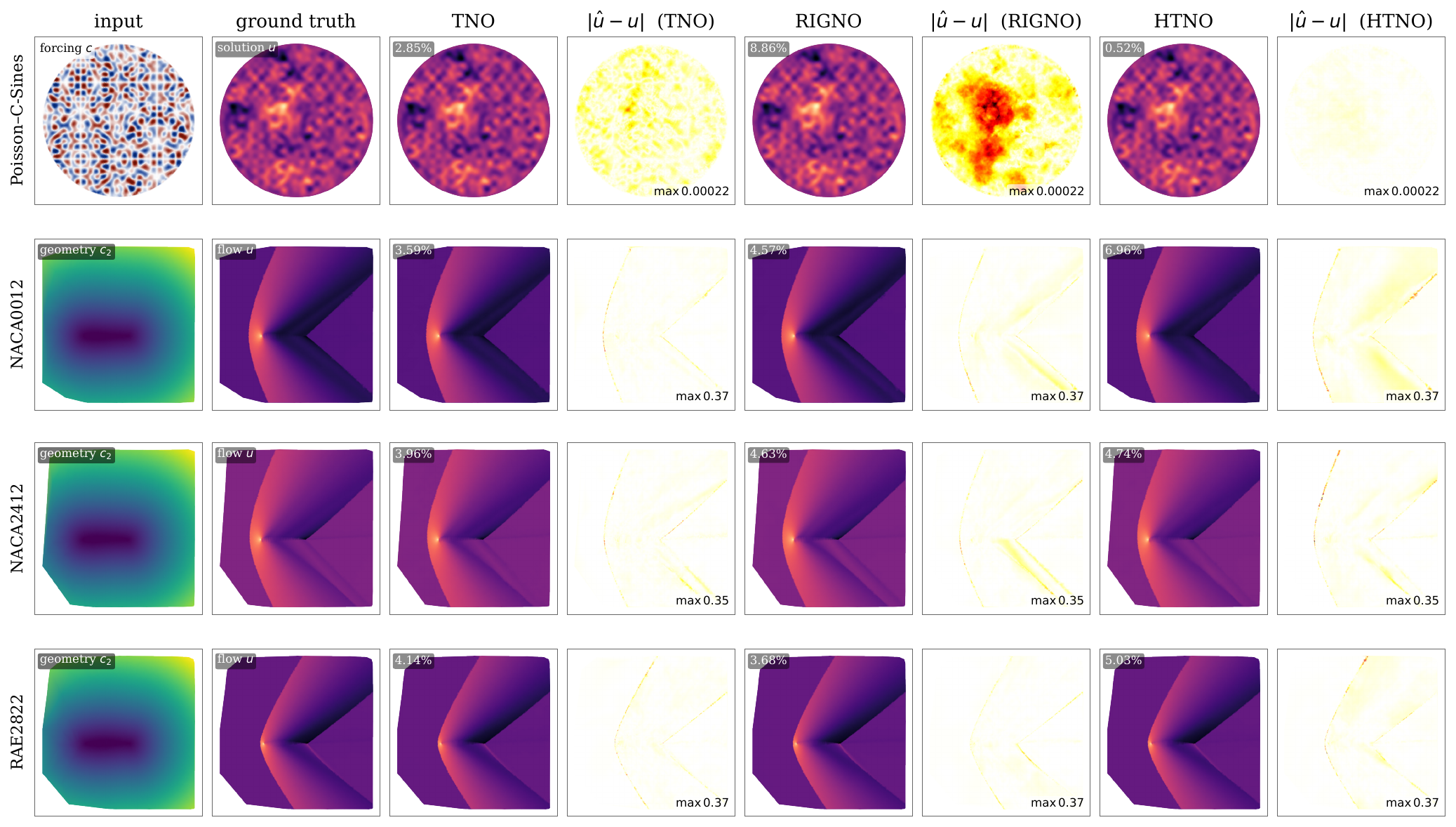}
  \caption{Qualitative comparison on the GAOT-suite benchmarks (Poisson-C-Sines, NACA0012, NACA2412, RAE2822). Each row shows one median-test-$L^1$ sample for the dataset; columns: model input, ground-truth field, and per-architecture prediction $\hat u$ followed by absolute error $|\hat u - u|$ for \name{}, RIGNO, and HTNO. Per-row error panels share a common color scale, allowing residual magnitudes to be directly comparable across architectures. Triangulation on the airfoil rows uses a vertex-local long-edge filter to drop spurious convex-hull bridges across the wing void.\vspace{-4mm}}
  \label{fig:gaot-suite-qualitative}
\end{figure}

\paragraph{GAOT-suite (PCS, NACA0012/2412, RAE2822)}
\Cref{fig:gaot-suite-qualitative} shows the analogous comparison on the four single-file steady-state datasets of~\cite{hao2024gaot}. PCS (top row) is dominated by a high-frequency multiscale sinusoidal forcing pattern, which HTNO and GAOT can represent well. 
The three airfoil rows expose the regime-sweep difficulty of this benchmark: the operators concentrate residual along the boundary layer and---for the supersonic RAE2822 sample---along the captured shock, where flatter message-passing baselines visibly under-resolve the discontinuity while \name{}'s harmonic + DEC structure appears to track it more tightly. 
Per-row error panels share a common color scale, so RIGNO and HTNO residuals are directly comparable.

\subsection{Differences in airfoils across datasets }
\label{app:af-vs-gaot-regime}

\name{} marginally underperforms RIGNO-18 on AF and outperforms it by $19$--$32\%$ relative test-$L^1$ on the GAOT. 
This is due to how the benchmarks differ in what their generators sweep across samples (\cref{tab:af-vs-gaot-regime}): \cite{li2023fourier} fixes the AF far-field at $M_\infty{=}0.8$, $\alpha{=}0^\circ$ and varies only the airfoil geometry, exposing a single distance-to-boundary field as conditioning (RIGNO~\cite{mousavi2025rigno} adopts this dataset directly); \cite{hao2024gaot} draws each NACA0012/2412/RAE2822 sample at a different $(M, \alpha)$ with $M\in[0.5,1.4]$ and $\alpha\in[0.5^\circ,5.0^\circ]$ on per-sample meshes, exposing $M$ and $\alpha$ as conditioning scalars.

\begin{table}[h]
  \centering
  \caption{What varies across samples for the two airfoil benchmark families (published quantities).}
  \label{tab:af-vs-gaot-regime}
  \small
  \begin{tabular}{@{}l c c c l@{}}
    \toprule
    Dataset family & $M_\infty$ & $\alpha$ & conditioning channels & varies per sample \\
    \midrule
    AF \cite{li2023fourier}                 & $0.8$ (fixed)  & $0^\circ$ (fixed)        & distance field             & airfoil geometry only \\
    GAOT airfoils \cite{hao2024gaot}        & $[0.5, 1.4]$   & $[0.5^\circ, 5.0^\circ]$ & $M$, $\alpha$ (scalars)    & $(M, \alpha,$ mesh$)$ \\
    \bottomrule
  \end{tabular}
\end{table}

\section{Related Work}
\label{app:related-work}
\vspace{-2mm}
\paragraph{Neural operators}
Operator learning (OL) approximates maps between infinite-dimensional function spaces, with the aim of producing models that generalize across initial conditions, coefficients, forcings, and discretizations~\citep{kovachki2023neural,boulle2024mathematical,kovachki2024operator,azizzadenesheli2024neural}.
\emph{DeepONet}~\citep{lu2021learning} and its physics-informed extensions~\citep{goswami2023physics,jiao2025one} learn a branch--trunk factorization of the operator with a universal approximation guarantee in function spaces.
The \emph{neural operator} family of \citet{li2020gno} instead realizes the operator through learned integral kernels: the Graph Neural Operator (GNO) approximates these kernels by message passing on sampled point clouds, while the Fourier Neural Operator (FNO)~\citep{li2021fourier} replaces the kernel by a parameterized spectral multiplier on a regular grid, with subsequent factorized~\citep{tran2023factorized}, wavelet~\citep{tripura2022wavelet}, and convolutional~\citep{raonic2023cno} variants.
Physics-informed neural networks complement these by deriving losses from the governing PDEs~\citep{li2024physics,viswanath2023neural,zhong2025physics}, and a separate line studies OL via model reduction~\citep{bhattacharya2021model}.

\paragraph{Neural operators on irregular geometries}
PDEs on irregular domains have motivated a body of operator-learning extensions beyond regular Cartesian grids.
Geo-FNO~\citep{li2023fourier} learns a deformation from the physical mesh to a uniform latent grid on which the FNO is applied; DAFNO~\citep{liu2023domain} embeds geometry through smoothed indicator masks; and Beyond-Regular-Grids~\citep{lingsch2024beyond} evaluates truncated spectral transforms directly on arbitrary point sets.
GINO~\citep{li2023geometry} couples a graph encoder/decoder with an FNO on a uniform latent grid, scaling operator learning to large 3D meshes.
Mesh-based simulators include MeshGraphNet~\citep{pfaff2021learning}, which propagates information on the simulation mesh together with a separate world-edge graph, and RIGNO~\citep{mousavi2025rigno}, which uses message passing on multiscale graphs; GAOT~\citep{hao2024gaot} pairs a multiscale attentional graph encoder with a transformer processor on a latent grid.
Transformer-based operators~\citep{li2023transformer,hao2023gnot,wu2024transolver,bryutkin2024hamlet,calvello2025continuum} treat unstructured nodes as tokens and attend either across all points or across learned cluster representatives, with universal physics transformers~\citep{alkin2024upt,alkin2025abupt} and PDE foundation models~\citep{herde2024poseidon} extending this idea to large-scale 3D aerodynamics~\citep{paischer2025going}.

A complementary line treats geometry intrinsically: DIMON~\citep{yin2024dimon} learns the operator across a diffeomorphic family of domains, geometric NOs on Riemannian manifolds~\citep{quackenbush2024geometric,chen2024learning} and gauge-equivariant intrinsic operators~\citep{cheng2026gauge,cheng2023equivariant} build manifold structure into the architecture, and Neural Green's Functions~\citep{yoo2025neuralgreens} parameterize geometry-dependent fundamental solutions.
What separates these methods from \name{} is that physical quantities are still represented as functions on \emph{points} (graph nodes, deformed grid samples, or tokens); cross-dimensional differential structure is recovered indirectly through learned kernels.
\name{}s explicitly address this gap: by keeping cochains of all degrees as first-class objects and routing information through fixed boundary, coboundary, and Hodge operators, geometric type and the conservation identities $d\circ d=0$ become structural rather than learned.

\paragraph{Discrete exterior calculus and structure-preserving discretizations}
The mathematical foundation behind TNOs is the discretization of differential forms.
Discrete exterior calculus~\citep{hirani2003discrete,desbrun2005discrete,crane2013digital,crane2018discrete,ptavckova2021simple,bell2012pydec} models physical quantities as cochains on oriented cells and replaces gradient, curl, and divergence by signed boundary operators that satisfy a discrete Stokes' theorem.
The same viewpoint underlies Whitney forms~\citep{bossavit1988whitney,bossavit1998computational}, mixed and N\'ed\'elec finite elements~\citep{nedelec1980mixed,hiptmair2002finite}, mimetic finite differences~\citep{hyman1997numerical,lipnikov2014mimetic}, and the finite element exterior calculus of \citet{arnold2006finite,arnold2010finite,arnold2018feec}, all of which preserve the de~Rham complex and the algebraic identity $B_k B_{k+1}=0$ at the discrete level.
TNOs inherit this structure: the cellular operators $d^k=B_{k+1}^\top$, $\delta^k$, and the Hodge Laplacian $\Delta_k$ used in our layers are exactly the DEC operators on the underlying complex.

\paragraph{Topological deep learning (TDL) \& architectures for PDE learning}
TDL~\citep{papamarkou2024position,hajij2023topological} extends graph neural networks by allowing message passing across cells of multiple ranks (vertices, edges, faces, and higher).
Simplicial complex networks generalize the Weisfeiler--Leman test to higher-order incidences~\citep{bodnar2021topological}, CW networks operate on regular cell complexes~\citep{bodnar2021cellular}. Combinatorial complexes~\citep{hajij2023combinatorial} unify these under a single neighborhood scheme. Sheaf neural networks~\citep{hansen2020sheaf,bodnar2022neural} and copresheaf topological networks~\citep{hajij2025copresheaf} attach learnable fiber maps to incidence relations, while Hodge Laplacian random walks~\citep{schaub2020random,vigano2026root} and topology-aware GNNs~\citep{leventhal2023modeling}, learning the hodge-star~\cite{smirnov2021hodgenet}, provide foundations.
Despite the appeal of the cellular viewpoint for physics, topological architectures remain almost absent from the OL literature, and the closest prior works are graph-based and message-passing in flavor: SNN-PDE~\citep{choi2024snnpde} uses simplicial convolutions for \textbf{time-dependent} PDEs.
Trask et al. \citet{trask2022ddec} learn Whitney-form-style metric data on a fixed graph to obtain a data-driven exterior calculus that enforces conservation.
Finally, DEC-HOGNN~\citep{liao2025boundaryvalue} builds a higher-order GNN on DEC and FEEC primitives for boundary-value electromagnetic problems.
While these methods exploit cell-level features, \textbf{they remain GNN-style architectures}.
None define true function-space operators, expose the Hodge decomposition as an architectural component or address multi-degree mPDE systems where unknowns at different ranks couple simultaneously through $d$ and $\delta$.
\name{}s address these by fixing information flow to the DEC operators and by exposing the Hodge decomposition in the layer (\cref{eq:tno-linear}), so that exact, coexact, and harmonic channels carry physically meaningful components.

\end{document}